**Original research article: Detecting the patient's need for help with machine learning.**
Corresponding author: Lauri Lahti (email: lauri.lahti@aalto.fi).

# Detecting the patient's need for help with machine learning

Lauri Lahti
*Department of Computer Science, Aalto University, Finland*

**Abstract:**

**Background:** Developing machine learning models to support health analytics requires increased understanding about statistical properties of self-rated expression statements used in health-related communication and decision making. To address this, our current research analyzes self-rated expression statements concerning the coronavirus COVID-19 epidemic and with a new methodology identifies how statistically significant differences between groups of respondents can be linked to machine learning results.

**Methods:** A quantitative study gathering the "need for help" ratings for twenty health-related expression statements concerning the coronavirus epidemic on an 11-point Likert scale, and nine answers about the person's health and wellbeing, sex and age. The study involved online respondents between 30 May and 3 August 2020 recruited from Finnish patient and disabled people's organizations, other health-related organizations and professionals, and educational institutions (n=673). We propose and experimentally motivate a new methodology of machine learning influence analysis for evaluating how machine learning results depend on and are influenced by various properties of the data which are identified with traditional statistical methods. We use a basic convolutional neural network model as a baseline architecture to enable comparable measurements between our parallel data subsets as well as for future experiments in a well-documented way.

**Results:** We found statistically significant Kendall rank-correlations and high cosine similarity values between various health-related expression statement pairs concerning the "need for help" ratings and a background question pair. With tests of Wilcoxon rank-sum, Kruskal-Wallis and one-way analysis of variance (ANOVA) between groups we identified statistically significant rating differences for several health-related expression statements in respect to groupings based on the answer values of background questions, such as the ratings of suspecting to have the coronavirus infection and having it depending on the estimated health condition, quality of life and sex. Our new methodology enabled us to identify how statistically significant rating differences were linked to machine learning results thus helping to develop better human-understandable machine learning models.

**Conclusions:** The self-rated "need for help" concerning health-related expression statements differs statistically significantly depending on the person's background information, such as his/her estimated health condition, quality of life and sex. With our new methodology statistically significant rating differences can be linked to machine learning results thus enabling to develop better machine learning to identify, interpret and address the patient's needs for well-personalized care.

**Ethics approval and consent to participate and for publication:** Aalto University Research Ethics Committee has carried out an ethical evaluation concerning the personal data acquisition of the current research project "Development of method for interpretation of health expressions based on machine learning to support various care events and persons" (DIHEML, in Finnish "Koneoppimista hyödyntävän menetelmän kehittäminen terveyttä koskevien ilmaisujen tulkitsemiseksi tukemaan erilaisia hoitotilanteita ja henkilöitä", see Lahti, 2017; Lahti, 2018) and has given a supporting ethical statement for it on 18 June 2019. DIHEML research project addresses the General Data Protection Regulation of the European Union in handling the research data. An informed consent was obtained from all individual persons participating in the data acquisition and regarding publishing their anonymized data sets.

**Availability of data and materials:** The data set supporting the conclusions of this article is included within the article and its supplementing Appendix A. While taking appropriate and sufficient anonymization actions in respect to addressing the General Data Protection Regulation of the European Union in handling the research data, DIHEML research project publishes an anonymized version of the current research data (an open access data set "Need for help related to the coronavirus COVID-19 epidemic") in the supplementing Appendix A, to be used by anyone for non-commercial purposes (while citing this publication).

**Competing interests:** The author declares that he has no competing interests.

**Funding:** The author did not receive any specific funding for this research work.

**Author contributions:** The contributions of the author Lauri Lahti include: conceived and developed the research plan; created the research design; designed the experiment; recruited the population sample and collected data; developed and implemented the online questionnaire system and its models, algorithms, data and materials; performed traditional statistical analyses; performed machine learning analyses; wrote the first and subsequent drafts and the final version of the manuscript; prepared tables, figures and the appendix; reviewed and commented the manuscript; performed editing of the manuscript.

**Acknowledgement:** The author wants to express gratitude to all the people who have kindly participated in answering to the online questionnaire of the research. Special thanks to the people associated with various Finnish patient and disabled people's organizations, other health and wellness organizations, and educational institutions as well as organizations of healthcare professionals, including also the representatives of Finnish Association for Emergency Medicine.

## 1. Background

A self-rated health condition answered to a single question has shown a strong validity and reliability for measuring and predicting multiple dimensions of the person's health (Gallagher et al., 2016; Wu et al., 2013). However, the self-rated health is affected by the phrasing, scales and ordering used in questions and answer options (Cullati et al., 2020; Garbarski et al., 2016; Joffer et al 2016; Borraccino et al., 2019). On the other hand, comprehensive modular questionnaire systems have been proposed and implemented, for example relying on International Classification of Functioning, Health and Disability, and Patient-Reported Outcomes Measurement Information System (PROMIS) (Tucker et al., 2014; Anttila et al., 2017). Despite the possibility to offer increasingly specifically tailored question sets and to create links between them (Jacobson et al., 2020; Schalet et al., 2015), a general challenge is to interpret the gained specific answers in greater agglomerated entities to make analytic conclusions and predictions in a broader context of the person's health and wellbeing, such as in a long-term care planning and clinical decision making (Deo, 2015).

Furthermore, besides using predefined questionnaire structures there is a great interest for developing adaptive methods that can identify the patient's needs from any kind of free text passages, such as from healthcare chatbots, patient diaries, online guidance and screening for care, or their derivatives, for example emergency phone calls that are immediately annotated with a speech recognition (resembling the previous proposals of Zhao et al., 2017; Gehrmann et al., 2018; Rojas-Barahona et al., 2018; Shickel et al., 2019). However, according to two reviews there is still a lack of systematic development for reliable evaluation metrics for healthcare chatbots (Abd-Alrazaq et al., 2020) and their algorithms have challenges in semantic understanding (Laranjo et al., 2018).

Think-aloud studies about self-rated health have identified sex- and age-dependent variations in the diversity and complexity of conceptualizations in interpretations and reasoning (Joffer et al 2016) and core categories that people use to describe and perceive health (Borraccino et al., 2019). Age-related differences in self-reported opinions, attitudes or behaviors about health can also be influenced by age-induced changes in cognitive and communicative functioning (Knäuper et al., 2016). There is a need to advance understandable and accurate communication between the patient and healthcare personnel and the patient's appropriate and sufficient involvement in decision making that addresses his/her needs (Sinclair, Jaggi, Hack, Russell et al., 2020; Sinclair, Jaggi, Hack, McClement et al., 2020).

These current challenges motivate us now to propose, develop and define a new methodology that we refer to as *machine learning influence analysis*. The methodology can be used to measure the patient's "need for help" ratings of expression statements in respect to groupings based on the answer values of background questions. Furthermore, the methodology enables to evaluate the applicability of training and validation of a machine learning model to learn the groupings concerning the ratings. The methodology enables to compare the validation accuracies of the machine learning model with the probabilities of pure chance of classifying the rating profiles correctly. In addition, the methodology enables to contrast the validation accuracies of the machine learning model with the occurrence of statistically significant rating differences for expression statements in respect to groupings based on the answer values of background questions. Table 1 summarizes the main steps of our proposed new methodology of machine learning influence analysis.

**Table 1.** A description of the methodology of machine learning influence analysis that can be used to measure the patient's "need for help" ratings of expression statements in respect to groupings based on the answer values of background questions, and further to evaluate the applicability of training and validation of a machine learning model to learn the groupings concerning the ratings.

| *Main steps of the methodology of machine learning influence analysis* |
|---|
| *Step 1.* Gathering questionnaire answers from persons representing various health and demographic backgrounds.<br>- Each person gives the *"need for help" ratings* for a set of common expression statements that describe imagined scenarios.<br>- The rating answers given by the person form his/her "need for help" rating profile. |
| *Step 2.* Identifying statistically significant and non-significant rating differences for expression statements in respect to groupings based on the *answer values of background questions* (for example groupings relying on the person's answer about his/her estimated health condition). |
| *Step 3.* Training and validation of a *machine learning model* (with a supervised learning approach) to learn the groupings concerning the "need for help" ratings. This step uses the same groupings of respondents that have been used in the step 2. |
| *Step 4.* Comparing the validation accuracies of the machine learning model with the probabilities of pure chance of classifying the rating profiles correctly (averaged from at least 100 separate training and validation sequences). |
| *Step 5.* Contrasting the validation accuracies of the machine learning model with the occurrence of statistically significant and non-significant rating differences for expression statements in respect to groupings based on the answer values of background questions (averaged from at least 100 separate training and validation sequences). |
| *Step 6.* Drawing conclusions about the applicability of the current machine learning model in this knowledge context. Based on the conclusions further finetuning of the model and iteratively repeating the steps 2-6. |

Within the limited space constraints of this research article, we now focus on introducing general principles of the new methodology and describe an illustrative empirical application of the methodology with our gathered experimental data.

In accordance with the methodology presented in Table 1, the above-mentioned previous research and current challenges motivate us now to address *two main research questions (RQ)*:

RQ1) How different people rate the "need for help" for a set of health-related expression statements and how this rating depends on the background information about the person (such as his/her demographic information and evaluation about own health and wellbeing)? This main research question RQ1 emphasizes especially the steps 1-2 of Table 1.

RQ2) What kinds of results can be gained when training a convolutional neural network model based on the "need for help" ratings to classify persons into groups based on their background information? This main research question RQ2 emphasizes especially the steps 3-6 of Table 1.

Relying on the methods and results developed in our previous research (Lahti 2017; Lahti 2018), we now analyze experimental measurements (n=673) including the "need for help" ratings for twenty health-related expression statements concerning COVID-19 epidemic, and nine answers about the person's health and wellbeing, sex and age. Our measuring methodology is adapted from the dimensional affective models which suggest that dimensions of pleasure, arousal, dominance and approach-avoidance have a fundamental role in human experience and response systems (Bradley & Lang, 1999a; Warriner et al., 2013; Mauss & Robinson, 2009). Our approach is also motivated by the previous research that has experimentally gathered a list of self-identified most significant mental imagery describing the patient's pain combined with associated triggers, affects, meanings and avoidance patterns (Berna et al., 2011).

Resembling the previous research in the context of artificial intelligence (Zhao et al., 2017; Gehrmann et al., 2018; Rojas-Barahona et al., 2018; Shickel et al., 2019), we wanted to evaluate the applicability of machine learning to support interpretation of the need for help in the patient's expressions. Machine learning is a methodology that aims at learning to recognize statistical patterns in data, typically relying on either an unsupervised or supervised approach. Unsupervised learning aims at identifying naturally occurring patterns or groupings which are present in the input data and it is often challenging for humans to judge the actual appropriateness and meaningfulness of the generated groupings (Deo 2015). On the other hand, supervised learning is often carried out with an aim to predict an outcome that is based on approximating an appropriate human-made classification. Supervised learning usually tries to perform classification by choosing among subgroups such a subgroup that can best describe a new instance of data and also to produce a prediction that consists of estimating an unknown parameter (Deo 2015). Supervised learning is also actively used to estimate risk and this can be considered to extend further than just approximating the human performance and to aim at identifying hidden characteristics of data (Deo 2015).

Since we aimed at identifying how the "need for help" ratings of expression statements can be used to classify persons into groups based on their background information, it was natural for us to focus now on experimenting with the supervised learning approach. To implement supervised learning, various alternative types of functions can be chosen to relate predicted values to the features that are present in the data, and these functions typically offer more flexibility for modeling than for example logistic regression models of traditional statistics (Deo 2015). These functions can be based on various alternative machine learning models, and among them artificial neural networks have achieved a high accuracy in classification tasks (Gehrmann et al., 2018). We decided to use a convolutional neural network model in our machine learning experiments since it has been successfully applied in classification of medical literature, patient records, clinical narratives and patient phenotypes (Hughes et al., 2017; Zhao et al., 2017; Gehrmann et al., 2018; Rojas-Barahona et al., 2018; Yao et al., 2019; Qing et al., 2019; Shickel et al., 2019), and it achieves good results with both image and textual input data (Bhandare et al., 2016).

In respect to the coronavirus COVID-19 epidemic, artificial neural networks have been applied to classify coronavirus-related online discussions and then to supply them with an emotional labeling based on a pre-existing emotion vocabulary and rules (Jelodar et al., 2020).

## 2. Materials and Methods

In accordance with Table 1, we now begin to describe our proposed new methodology in respect to the step 1 that consists of gathering questionnaire answers from persons representing various health and demographic backgrounds. In the time period ranging from 30 May to 3 August 2020 we gathered with an online questionnaire from 673 unique persons twenty rating answers that measured the degree of the "need for help" that the person associated with the imagined care situations related to the coronavirus COVID-19 epidemic. In addition, we gathered nine answers about the person's background information. All these answers were gathered as a part of a greater data acquisition entity (Lahti, 2020) with some supplementing questionnaire items that will be reported in a more detail in another future publication. The respondents were recruited from various Finnish patient and disabled people's organizations, other health and wellness organizations, and educational institutions as well as organizations of healthcare professionals. When accessing the online questionnaire, the person was informed that only persons who are at least 16 years old are allowed to participate. Furthermore, to address the General Data Protection Regulation of the European Union a privacy notice about the research was shown to the person and he/she was asked to give an approval for handling his/her data.

Before the online questionnaire started to collect actual answers, the person was provided with the following guidance texts about how he/she should perform the interpretation tasks: *"We ask you to evaluate different expressions, for example the expression 'I am happy'. Interpret how much each expression tells about the need for help. Give your interpretation about the expression on a numeric scale 0-10. 0 indicates the smallest possible need for help and 10 indicates the greatest possible need for help."* Then a small training phase allowed the person to get accustomed to give the "need for help" ratings by rating three expression statements: "I have a good health condition.", "I have a bad health condition." and "I have an ordinary health condition." The answers that the person gave during the training phase were excluded from the data set that we use in the following analysis.

After the training phase, the person was provided with the following guidance texts to still further clarify how he/she should perform the interpretation tasks: *"Do not interpret how much the expression tells about just your own situation. Instead, interpret what kind of impression this expression induces in you. Thus give your interpretation about the expression's meaning in respect to the mentioned property."* After showing those guidance texts, the person was allowed to start giving actual answers, i.e. to perform the actual interpretation tasks.

In the actual interpretation tasks, our online questionnaire asked the person to give a rating of the "need for help" for twenty expression statements (ES) that we had extracted with the method we developed and reported in our previous research (Lahti et al., 2018) from the official national guidelines of Finnish Institute for Health and Welfare (THL) (2020) and international guidelines of World Health Organization (WHO) (2020) concerning the coronavirus COVID-19 epidemic. These twenty expression statements ES1-ES20 included among others descriptions of possible symptoms of the coronavirus, how to deal mild cases of the coronavirus with just self-care, when one should seek admission for professional care and what kinds of practicalities are suggested as a prevention (see Table 2). The expression statements were shown, one at a time, in a speech bubble above a simple briefly animating face figure that remained the same for all the expression statements (see Figure 1 and further details in Appendix A).

Furthermore, the person was asked to answer to nine background questions (BQ, see Table 3). These gathered four answers concerning his/her evaluation about own health, quality of life, and satisfaction about health and ability, responded on a 9-point Likert scale (BQ1 and BQ5-BQ7, adapted from de Bruin et al., 1996; Nosikov & Gudex, 2003; Koskinen et al., 2012; Aalto et al., 2013). In addition, binary no/yes answers were gathered to questions asking if a health problem reduces the person's ability (BQ2) and if he/she has a continuous or repeated need for a doctor's care (BQ4) (adapted from Koskinen et al., 2012). The person was also asked to tell his/her sex (BQ8) and age (BQ9) and to indicate if a doctor had identified one or more diseases in him/her and to describe them (BQ3) (in a form adapted from Koskinen et al., 2012).

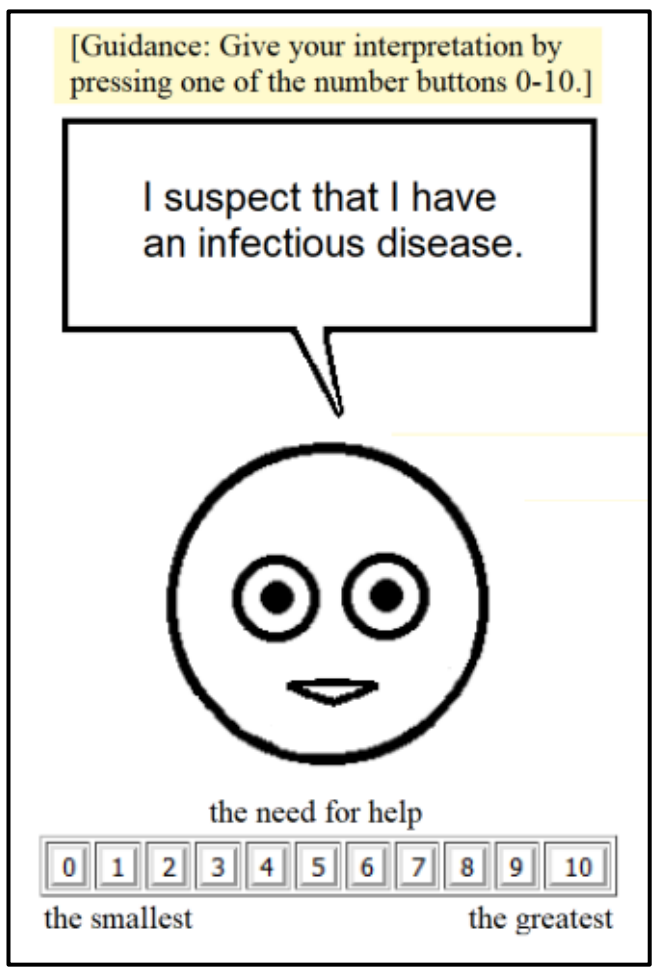

**Figure 1.** Gathering the "need for help" rating for an expression statement on an 11-point Likert scale with an online questionnaire.

We have gathered questionnaire answers in Finnish language but we now report our results in English (see original Finnish texts in Appendix A). Due to inherent linguistic and cultural differences we assume that the semantic meanings in the translated English versions of expression statements cannot fully match with the original Finnish meanings. On the other hand, we have aimed to follow carefully also those adapted Finnish translations that have been used already earlier in Finnish national health surveys (Koskinen et al., 2012; Aalto et al., 2013).

**Table 2.** Expression statements (ES) concerning the coronavirus COVID-19 epidemic that were rated by the person in respect to the impression about the "need for help".

| *Compact notation* | *Expression statement* | *Range of values for the person's answer (indicating the "need for help" rating)* |
|---|---|---|
| ES1 | "I have a flu." | 0-10 |
| ES2 | "I have a cough." | 0-10 |
| ES3 | "I have a shortness of breath." | 0-10 |
| ES4 | "My health condition is weakening." | 0-10 |
| ES5 | "I have a sore throat." | 0-10 |
| ES6 | "I have muscular ache." | 0-10 |
| ES7 | "I have a fever." | 0-10 |
| ES8 | "A sudden fever rises for me with 38 degrees of Celsius or more." | 0-10 |
| ES9 | "I suspect that I have now become infected by the coronavirus." | 0-10 |
| ES10 | "I have now become infected by the coronavirus." | 0-10 |
| ES11 | "I am quarantined from meeting other people ordinarily so that the spreading of an infectious disease could be prevented." | 0-10 |
| ES12 | "I must be inside a house without getting out." | 0-10 |
| ES13 | "I must be without a human companion." | 0-10 |
| ES14 | "I do not cope in everyday life independently without getting help from other persons." | 0-10 |
| ES15 | "I do not cope at home independently without getting help from persons who originate outside of my home." | 0-10 |
| ES16 | "I have an infectious disease." | 0-10 |
| ES17 | "I have an infectious disease that has been verified by a doctor." | 0-10 |
| ES18 | "I suspect that I have an infectious disease." | 0-10 |
| ES19 | "I have a bad health condition." | 0-10 |
| ES20 | "I have an ordinary health condition." | 0-10 |

**Table 3.** Background questions (BQ) presented to the person.

| *Compact notation* | *Question about the person's background information* | *Range of values for the person's answer* |
|---|---|---|
| BQ1: an estimated health condition | "What kind of health condition you have currently according to your opinion?" (de Bruin et al., 1996; Koskinen et al., 2012) | A 9-point Likert scale supplied with the following partial labeling: "9 Good", "8 –", "7 Rather good", "6 –", "5 Medium", "4 –", "3 Rather bad", "2 –", "1 Bad". |
| BQ2: a health problem reduces ability | "Do you have a permanent or long-lasting disease or such deficit, ailment or disability that reduces your ability to work or to perform your daily living activities? Here the question refers to all long-lasting diseases identified by a doctor, and also to such ailments not identified by a doctor which have lasted at least three months but which affect your ability to perform your daily living activities." (Koskinen et al., 2012) | No or yes. |
| BQ3: one or more diseases identified by a doctor | "Has there been a situation that a doctor has identified in you one or several of the following diseases?" (Koskinen et al., 2012) | The person answers by selecting one or more options from a list of diseases (Koskinen et al., 2012), see details in Appendix A. For some options there is a question "other, what?" and an adjacent text input box so that the person can write some additional information concerning that option. |
| BQ4: a continuous or repeated need for a doctor's care | "Do you need continuously or repeatedly care given by a doctor for a long-lasting disease, deficit or disability that you have just mentioned?" (Koskinen et al., 2012) | No or yes. |
| BQ5: the quality of life | "How would you rate your quality of life? Give your estimate based on the latest two weeks." (Nosikov & Gudex, 2003; Aalto et al., 2013) | A 9-point Likert scale supplied with the following partial labeling: "9 Very good", "8 –", "7 Good", "6 –", "5 Neither good nor bad", "4 –", "3 Bad", "2 –", "1 Very bad". |
| BQ6: the satisfaction about health | "How satisfied are you with your health? Give your estimate based on the latest two weeks." (Nosikov & Gudex, 2003; Aalto et al., 2013) | A 9-point Likert scale supplied with the following partial labeling: "9 Very satisfied", "8 –", "7 Satisfied", "6 –", "5 Neither satisfied nor unsatisfied", "4 –", "3 Unsatisfied", "2 –", "1 Very unsatisfied". |
| BQ7: the satisfaction about ability | "How satisfied are you with your ability to perform your daily living activities? Give your estimate based on the latest two weeks." (Nosikov & Gudex, 2003; Aalto et al., 2013) | A 9-point Likert scale supplied with the following partial labeling: "9 Very satisfied", "8 –", "7 Satisfied", "6 –", "5 Neither satisfied nor unsatisfied", "4 –", "3 Unsatisfied", "2 –", "1 Very unsatisfied". |
| BQ8: the sex | "Tell what is your sex. The answer alternatives are similar as in the earlier health surveys of Finnish Institute for Health and Welfare (THL) to maintain comparability with the earlier results." (Koskinen et al., 2012) | Man or woman. |
| BQ9: the age | "Tell what is your age." (Koskinen et al., 2012) | 16 years, 17 years, ..., 99 years, 100 years or more. |

To address our main research question RQ1, we use traditional statistical tests to evaluate overall answer distributions. We compute Kendall rank-correlation and cosine similarity measures for each comparable pair of parameter values of the "need for help" ratings of expression statements and the answers of the background questions. Then we compute Wilcoxon rank-sum test (i.e., Mann–Whitney U test) between two groups and Kruskal-Wallis test between three groups to identify statistically significant rating differences for each expression statement in respect to groupings based on the answer values of each background question. We supplement this with tests of one-way analysis of variance (ANOVA) between two groups and between three groups.

To address our main research question RQ2, we carry out machine learning experiments with a basic implementation of a convolutional neural network algorithm that we run in a TensorFlow programming environment (TensorFlow image classification tutorial, 2020). This enables to evaluate the general applicability of machine learning approach in this knowledge context. We have chosen this specific implementation of a convolutional neural network for our experiments since this model is openly and easily available for testing purposes in a currently popular programming environment and the model's internal computational logic is clearly documented. We use this model as a baseline architecture to gain measures of the performance of machine learning that enable comparison between our parallel data subsets as well as offer our current results to be compared later with future experiments in a well-documented way. We train the machine learning model with the same groups that we use to identify statistically significant rating differences, and this offers insight about how the dependencies between ratings and background information can influence the results of machine learning. Based on the gained findings we then make some conclusions motivated by the previous research and discuss about implications for developing the methodology for interpretation of the patient's expressions to support his/her personalized care.

It needs to be emphasized that we evaluate the general applicability of machine learning approach for interpretation of the patient's expressions now in such a way that our current highest developmental priority is *not* to reach a model that manages to learn to detect given groupings very well. Instead, our current highest developmental priority is to propose and experimentally motivate a new methodology that we have developed for evaluating how machine learning results depend on various properties of the data which can be inspected and identified with traditional statistical methods. Thus due to the overall complexity of modeling semantics of a natural language and the limited size of the current data set our gained results are *not* meant to introduce a model that can actually learn the groupings very well. Instead, we aim to introduce now a new methodology that can be used for analyzing how the machine learning models are influenced by the properties of the data so that these notions can be exploited to develop better human-understandable machine learning and furthermore to help to address the traditional challenges of interpreting reliably and intuitively machine learning results (Deo 2015).

## 3. Results

In accordance with Table 1, we now continue to describe our proposed new methodology in respect to the step 2 that consists of identifying statistically significant and non-significant differences for expression statements in respect to groupings based on the answer values of background questions (for example groupings relying on the person's answer about his/her estimated health condition). To offer some background how the statistically significant and non-significant patterns may emerge from the full set of questionnaire answers, we first describe in detail some general distributional properties of the gathered answers.

We gained a diverse distribution of answer values for the background questions (n=673). Table 4 shows the frequencies of persons giving the answer values 1-9 for the background questions BQ1 and BQ5-BQ7, and Table 5 describes the distribution of answer values for the background questions BQ2-BQ4 and BQ8-BQ9.

**Table 4.** Frequencies of persons giving the answer values 1-9 for the background questions BQ1 and BQ5-BQ7 (n=673). M=mean, Mdn=median, SD=standard deviation.

| *Background question (BQ)* | *Answer value* | | | | | | | | | | | |
|---|---|---|---|---|---|---|---|---|---|---|---|---|
| | *1* | *2* | *3* | *4* | *5* | *6* | *7* | *8* | *9* | *M* | *Mdn* | *SD* |
| BQ1: an estimated health condition | 11 (2%) | 3 (~0%) | 40 (6%) | 66 (10%) | 98 (15%) | 45 (7%) | 162 (24%) | 129 (19%) | 119 (18%) | 6.53 | 7 | 1.97 |
| BQ5: the quality of life | 7 (1%) | 6 (1%) | 29 (4%) | 47 (7%) | 101 (15%) | 84 (12%) | 187 (28%) | 123 (18%) | 89 (13%) | 6.53 | 7 | 1.77 |
| BQ6: the satisfaction about health | 17 (3%) | 10 (1%) | 68 (10 %) | 61 (9%) | 84 (12%) | 78 (12%) | 151 (22%) | 138 (21%) | 66 (10%) | 6.13 | 7 | 2.04 |
| BQ7: the satisfaction about ability | 8 (1%) | 9 (1%) | 44 (7%) | 28 (4%) | 54 (8%) | 58 (9%) | 156 (23%) | 128 (19%) | 188 (28%) | 6.98 | 7 | 1.98 |

**Table 5.** The distribution of answer values for the background questions BQ2-BQ4 and BQ8-BQ9. M=mean, Mdn=median, SD=standard deviation.

| *Background question (BQ)* | *Answer value* |
|---|---|
| BQ2: a health problem reduces ability | No (coded as 1): 219 (33%); Yes (coded as 2): 454 (67%) (M=1.67; Mdn=2; SD=0.47) |
| BQ3: one or more diseases identified by a doctor | Disease category (the number of unique persons who selected the category): Lung diseases: 126; Heart and circulatory diseases: 177; Joint and back diseases: 301; Injuries:103; Mental health problems: 188; Vision and hearing deficits: 191; Other diseases: 345 |
| BQ4: a continuous or repeated need for a doctor's care | No (coded as 1): 364 (54%); Yes (coded as 2): 309 (46%) (M=1.46; Mdn=1; SD=0.50) |
| BQ8: the sex | Man (coded as 1): 123 (18%); Woman (coded as 2): 550 (82%) (M=1.82; Mdn=2; SD=0.39) |
| BQ9: the age | Belonging to an age range category (the lower bound is included in the range but not the upper bound): 16-20 years: 143 (21%); 20-30 years: 21 (3%); 30-40 years: 61 (9%); 40-50 years: 96 (14%); 50-60 years: 135 (20%); 60-70 years: 141 (21%); 70-80 years: 64 (10%); 80-90 years: 12 (2%); 90 years or more: 0 (0%) (M=46.93; Mdn=51; SD=19.57) |

To address our main research question RQ1, we evaluated how the "need for help" ratings depend on the background information about the person. For the further analysis, the original "need for help" rating values in the range 0-10 were transformed linearly to a new range 0.0-1.0. Figures 2 and 3 show for five expression statements ES4, ES9-ES10 and ES19-ES20 how the "need for help" ratings depend on the person's answer value to the background question BQ1 that is the person's estimation about his/her health condition. Figure 2a shows rating mean values for the nine separate groups of respondents corresponding to each possible answer alternative about the estimated health condition (in the range 1-9). Figures 2b and 2c show the increase of the "need for help" rating mean values from the baseline rating mean value of ES20. On the other hand, Figure 3 illustrates in a more detail the distribution of the relative frequency of respondents for each alternative rating value in the range 0.0-1.0, in respect to the background questions BQ1 and BQ9.

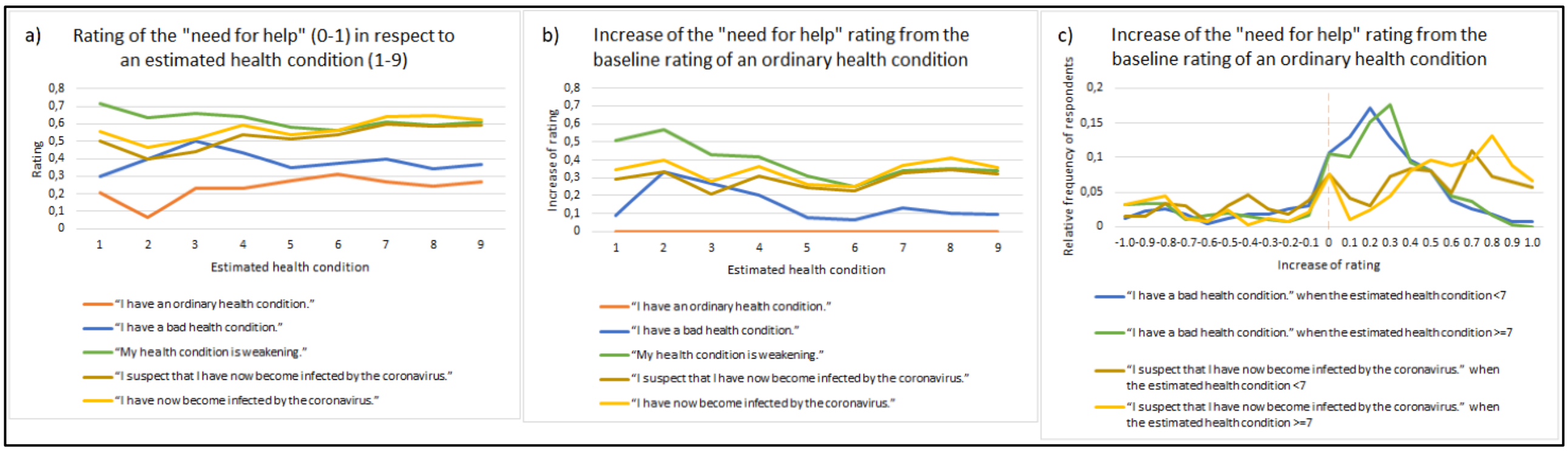

**Figure 2.** a) The "need for help" rating mean values (transformed into the range 0.0-1.0) for expression statements ES4, ES9-ES10 and ES19-ES20 in respect to the person's answer value to the background question BQ1 (an estimated health condition, 1-9), n=673. b)-c) Increase of the "need for help" rating mean values from the baseline rating mean value that the person gives for the expression statement ES20 ("I have an ordinary health condition."), n=673.

We computed Kendall rank-correlation and cosine similarity measures for each comparable pair of parameter values of the "need for help" ratings of expression statements ES1-ES20 and the answers of the background questions BQ1 and BQ5-BQ7. Motivated by a recommendation of Akoglu (2018) we considered a Kendall rank-correlation measure greater than or equal to 0.70 to indicate a significant correlation and the statistical significance levels were defined as $p<0.05$, $p<0.01$ and $p<0.001$. Before computing cosine similarity measures the answer values of each parameter were normalized by the formula $(x - min(x))/(max(x)-min(x))$ and then these new values were shifted so that the mean value was positioned to the zero by the formula $(x - mean(x))$.

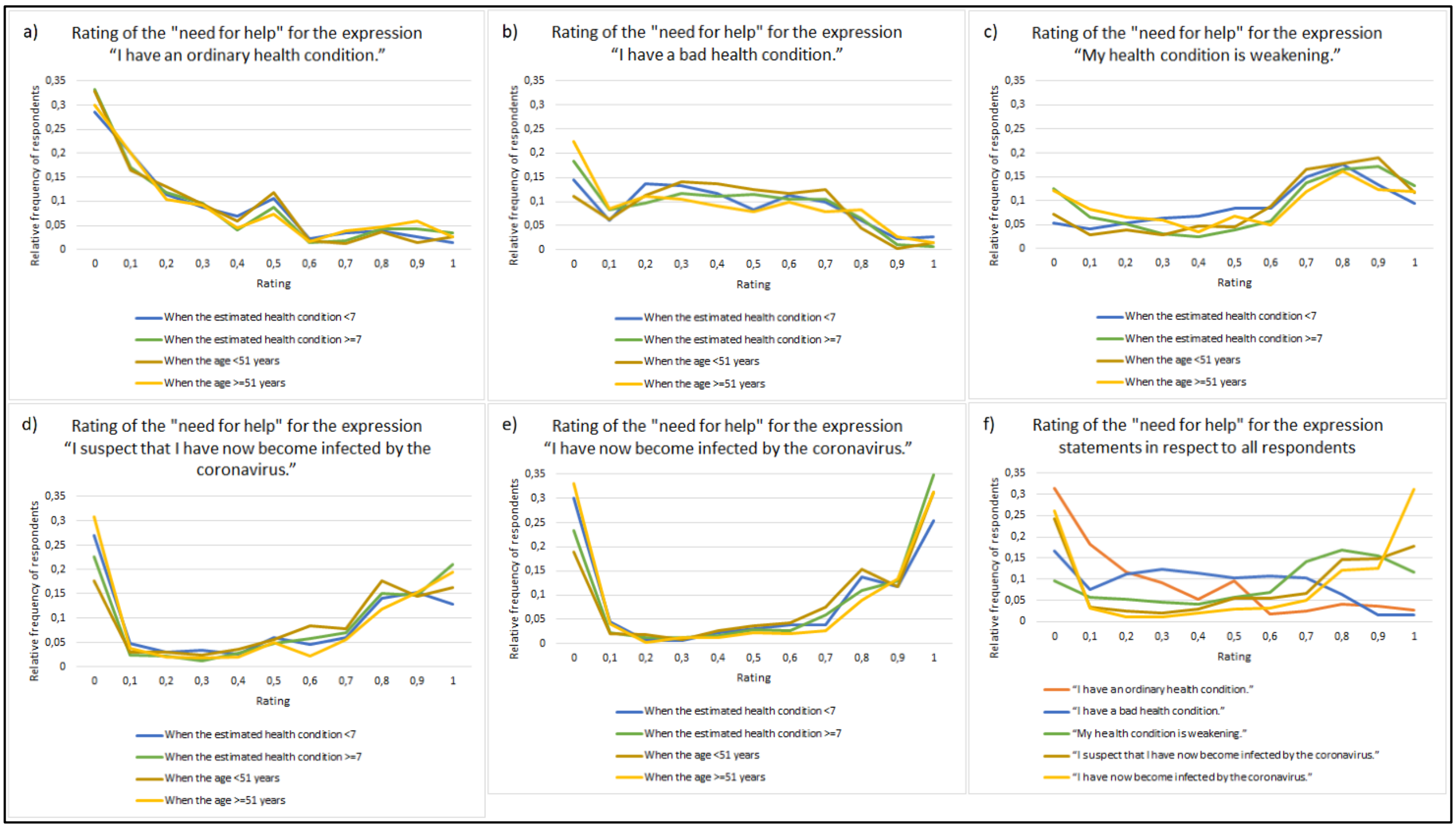

**Figure 3.** a)-e) The relative frequency of respondents for each alternative "need for help" rating value (transformed into the range 0.0-1.0) concerning expression statements ES4, ES9-ES10 and ES19-ES20 in respect to the person's answer value to the background questions BQ1 (an estimated health condition) and BQ9 (the age), n=673. f) Rating value distributions for the expression statements ES4, ES9-ES10 and ES19-ES20 in respect to all respondents together, n=673.

A significant correlation (>=0.70 with the level p<0.001; see Akoglu, 2018) was found between the expression statement pairs ES16&ES17 (0.91), ES14&ES15 (0.86), ES9&ES10 (0.79), ES16&ES18 (0.78), ES17&ES18 (0.77), ES7&ES8 (0.75) and ES1&ES2 (0.73), and between the background question pair BQ1&BQ6 (0.71), all these were statistically significant with the level p<0.001. The highest cosine similarity values included the same pairs of expression statements and the pair of background questions, thus reaching the following values: ES16&ES17 (0.97), ES14&ES15 (0.95), ES9&ES10 (0.92), ES16&ES18 (0.90), ES17&ES18 (0.89), ES7&ES8 (0.87), ES1&ES2 (0.80), and BQ1&BQ6 (0.82). This same highest cosine similarity value range was reached also by the pairs ES8&ES9 (0.87), ES3&ES4 (0.83), ES10&ES17 (0.82), ES9&ES17 (0.81), ES10&ES16 (0.80) and BQ5&BQ6 (0.80).

We computed Wilcoxon rank-sum test (i.e., Mann–Whitney U test) between two groups and Kruskal-Wallis test between three groups to identify statistically significant rating differences for expression statements ES1-ES20 in respect to groupings based on the answer values of each background question (BQ), as shown in Table 6. We created groupings of two groups so that the "group 1" contained those respondents who gave an answer value that was lower than the mean value of all the answer values to the background question, and the "group 2" contained all the other respondents. We created groupings of three groups so that the respondents could be divided the most evenly into three ranges of answer values of the background question. The statistical significance levels were defined as p<0.05, p<0.01 and p<0.001. Table 6 shows also the differences of mean ratings for the groupings. For example, for ES4 (a weakening health condition) the younger respondents gave a mean rating value 0.66 that was 0.10 greater than the mean rating value 0.56 given by the older respondents (BQ9, for two groups).

**Table 6.** Groupings based on the answer values of each background question (BQ) that enable to compute the differences of mean ratings for expression statements ES1-ES20 concerning the "need for help" ratings. For groupings of two groups the difference of mean ratings (each mean rating in the range 0.0-1.0) is computed by the formula $(M_1-M_2)$, and for groupings of three groups by the formula $max(\{(M_1-M_3),(M_1-M_2),(M_2-M_3)\})$. Wilcoxon rank-sum test (for two groups) and Kruskal-Wallis test (for three groups) indicate the statistically significant rating differences with levels $p<0.05$, $p<0.01$ and $p<0.001$, denoted by the symbols *, ** and ***, respectively. Training and validation metrics of the convolutional neural network model are averaged from 100 separate training and validation sequences to learn a labeling that matches the grouping (n=673). For groupings of two groups the probability of pure chance of classifying the rating profiles correctly is computed by the formula $(max(\{n_1,n_2\}))/(n_1+n_2)$, and for groupings of three groups by the formula $(max(\{n_1,n_2,n_3\}))/(n_1+n_2+n_3)$. M=mean, Mdn=median, SD=standard deviation.

| | *Statistically significant rating differences* | *Training and validation metrics of the convolutional neural network model to learn a labeling that matches the grouping* | | | | | *Comparison of the validation accuracy with the probability of pure chance* | |
|---|---|---|---|---|---|---|---|---|
| *Grouping based on the answer value (x) of the background question (BQ)* | *Expression statements (ES) having statistically significant rating differences in the grouping (the difference of mean ratings about the need for help)* | *Epoch step* | *Training loss* | *Training accuracy* | *Validation loss* | *Validation accuracy* | *Probability of pure chance of classifying the rating profiles correctly (based on the size of the greatest group)* | *Difference of the mean validation accuracy and the probability of pure chance of classifying the rating profiles correctly* |
| BQ1, two groups: x<7 ($n_1$=263), x>=7 ($n_2$=410) | ES6 (0.07)***, ES8 (-0.08)**, ES9 (-0.08)**, ES10 (-0.09)**, ES7 (-0.05)*, ES16 (-0.06)*, ES17 (-0.08)*, ES18 (-0.05)* | M=11.26 Mdn=11 SD=2.39 | M=0.55 Mdn=0.55 SD=0.03 | M=0.73 Mdn=0.72 SD=0.02 | M=0.59 Mdn=0.59 SD=0.01 | M=0.69 Mdn=0.69 SD=0.02 | 0.61 | 0.08 |
| BQ1, three groups: x<6 ($n_1$= 218), 6<=x<8 ($n_2$=207), x>=8 ($n_3$=248) | ES6 (0.08)**, ES5 (-0.05)*, ES8 (-0.08)*, ES9 (-0.09)*, ES11 (0.08)* | M=4.85 Mdn=4 SD=1.8 | M=1.02 Mdn=1.03 SD=0.03 | M=0.48 Mdn=0.47 SD=0.03 | M=1.06 Mdn=1.06 SD=0.01 | M=0.40 Mdn=0.40 SD=0.02 | 0.37 | 0.03 |
| BQ2, two groups: x<2 ($n_1$=219), x>=2 ($n_2$=454) | ES11 (-0.08)***, ES6 (-0.06)**, ES3 (0.06)*, ES14 (0.06)*, ES15 (0.08)* | M=5.55 Mdn=6 SD=2.91 | M=0.57 Mdn=0.56 SD=0.04 | M=0.69 Mdn=0.69 SD=0.02 | M=0.63 Mdn=0.63 SD=0.01 | M=0.66 Mdn=0.66 SD=0.02 | 0.67 | -0.01 |
| BQ4, two groups: x<2 ($n_1$=364), x>=2 ($n_2$=309) | ES6 (-0.06)**, ES11 (-0.06)* | M=3.44 Mdn=3 SD=1.28 | M=0.67 Mdn=0.67 SD=0.01 | M=0.59 Mdn=0.60 SD=0.02 | M=0.68 Mdn=0.67 SD=0 | M=0.57 Mdn=0.57 SD=0.02 | 0.54 | 0.03 |
| BQ5, two groups: x<7 ($n_1$=274), x>=7 ($n_2$=399) | ES6 (0.06)**, ES9 (-0.08)**, ES10 (-0.09)**, ES11 (0.06)*, ES16 (-0.06)*, ES17 (-0.07)* | M=3.27 Mdn=3 SD=1.65 | M=0.64 Mdn=0.63 SD=0.02 | M=0.65 Mdn=0.65 SD=0.03 | M=0.66 Mdn=0.67 SD=0 | M=0.60 Mdn=0.60 SD=0.02 | 0.59 | 0,01 |
| BQ5, three groups: x<6 ($n_1$=190), 6<=x<8 ($n_2$=271), x>=8 ($n_3$=212) | ES9 (-0.15)***, ES10 (-0.15)***, ES6 (0.07)**, ES8 (-0.11)*, ES16 (-0.09)*, ES17 (-0.10)*, ES20 (0.05)* | M=3.63 Mdn=4 SD=1.33 | M=1.05 Mdn=1.05 SD=0.02 | M=0.44 Mdn=0.44 SD=0.03 | M=1.07 Mdn=1.07 SD=0.01 | M=0.42 Mdn=0.43 SD=0.03 | 0.40 | 0.02 |
| BQ6, two groups: x<7 ($n_1$=318), x>=7 ($n_2$=355) | ES11 (0.08)***, ES6 (0.06)** | M=5.35 Mdn=5 SD=1.61 | M=0.62 Mdn=0.63 SD=0.02 | M=0.63 Mdn=0.63 SD=0.03 | M=0.65 Mdn=0.65 SD=0 | M=0.60 Mdn=0.60 SD=0,02 | 0.53 | 0.07 |
| BQ6, three groups: x<6 ($n_1$=240), 6<=x<8 ($n_2$=229), x>=8 ($n_3$=204) | ES11 (0.09)**, ES6 (0.07)* | M=3.89 Mdn=4 SD=1.8 | M=1.05 Mdn=1.06 SD=0.03 | M=0.41 Mdn=0.41 SD=0.03 | M=1.08 Mdn=1.08 SD=0 | M=0.39 Mdn=0.39 SD=0.03 | 0.36 | 0.03 |
| BQ7, two groups: x<7 ($n_1$=201), x>=7 ($n_2$=472) | ES6 (0.08)***, ES11 (0.07)**, ES19 (0.07)** | M=7.26 Mdn=7 SD=1.63 | M=0.53 Mdn=0.54 SD=0.02 | M=0.75 Mdn=0.74 SD=0.01 | M=0.59 Mdn=0.59 SD=0 | M=0.72 Mdn=0.72 SD=0.01 | 0.70 | 0.02 |
| BQ7, three groups: x<6 ($n_1$=143), 6<=x<8 ($n_2$=214), x>=8 ($n_3$=316) | ES6 (0.09)**, ES11 (0.10)** | M=1.31 Mdn=1 SD=0.61 | M=1.05 Mdn=1.06 SD=0.02 | M=0.45 Mdn=0.45 SD=0.02 | M=1.07 Mdn=1.07 SD=0.01 | M=0.47 Mdn=0.48 SD=0.02 | 0.47 | 0.00 |
| BQ8, two groups: x<2 ($n_1$=123), x>=2 ($n_2$=550) | ES4 (-0.11)***, ES12 (-0.13)***, ES14 (-0.20)***, ES15 (-0.20)***, ES3 (-0.10)**, ES10 (-0.12)**, ES11 (-0.08)**, ES8 (-0.09)*, ES9 (-0.10)*, ES13 (-0.07)*, ES16 (-0.10)*, ES17 (-0.09)*, ES18 (-0.08)* | M=6.14 Mdn=6 SD=1.69 | M=0.42 Mdn=0.42 SD=0.02 | M=0.83 Mdn=0.83 SD=0.01 | M=0.48 Mdn=0.48 SD=0.01 | M=0.79 Mdn=0.78 SD=0.01 | 0.82 | -0.03 |
| BQ9, two groups: x<51 ($n_1$=333), x>=51 ($n_2$=340) | ES1 (0.09)***, ES2 (0.10)***, ES3 (0.13)***, ES4 (0.10)***, ES5 (0.08)***, ES14 (0.14)***, ES15 (0.14)***, ES7 (0.06)*, ES8 (0.09)*, ES11 (-0.05)*, ES13 (0.06)*, ES19 (0.04)* | M=5.79 Mdn=6 SD=1.44 | M=0.58 Mdn=0.58 SD=0.02 | M=0.69 Mdn=0.69 SD=0.02 | M=0.61 Mdn=0.61 SD=0.01 | M=0.68 Mdn=0.68 SD=0.02 | 0.51 | 0.17 |
| BQ9, three groups: x<40 ($n_1$=225), 40<=x<60 ($n_2$=231), x>=60 ($n_3$=217) | ES1 (0.10)***, ES2 (0.12)***, ES3 (0.17)***, ES4 (0.13)***, ES5 (0.09)***, ES14 (0.17)***, ES15 (0.18)***, ES7 (0.09)**, ES11 (-0.10)**, ES8 (0.13)*, ES10 (0.14)*, ES19 (0.06)*, ES20 (-0.08)* | M=7.13 Mdn=7 SD=1.45 | M=0.93 Mdn=0.93 SD=0.03 | M=0.54 Mdn=0.54 SD=0.03 | M=0.98 Mdn=0.98 SD=0.01 | M=0.50 Mdn=0.50 SD=0.03 | 0.34 | 0.16 |

Supplementing tests of one-way analysis of variance (ANOVA) between two groups and between three groups indicated statistically significant rating differences largely for the same expression statements as Wilcoxon rank-sum test and Kruskal-Wallis test. However, this statistical significance did not reappear with ANOVA tests between groups for ES5 in respect to BQ1 for three groups, ES14 in respect to BQ2 for two groups, ES19 in respect to BQ9 for three groups, and ES20 in respect to BQ5 for three groups. ANOVA tests between groups indicated also some additional statistically significant rating differences, such as for ES9-ES10 and E17 in respect to BQ2 for two groups, ES9-ES10 in respect to BQ9 for two groups, and ES4 in respect to BQ7 for two groups.

A complete listing of means, medians and standard deviations of the "need for help" ratings for the groupings is provided in Appendix A which includes also a comprehensive listing of Kendall rank-correlation and cosine similarity measures, and tests of Wilcoxon rank-sum, Kruskal-Wallis and one-way analysis of variance (ANOVA) between groups.

Figure 4 illustrates for all the twenty expression statements ES1-ES20 how the "need for help" rating mean values differ between the respondents who indicate a lower estimated health condition and the respondents who indicate a higher estimated health condition (BQ1, for two groups). Besides comparing just single expression statements between groups, we can now also identify the emergence of two different ranking orders for all the twenty expression statements ES1-ES20 in respect to the grouping based on the answer values of the background question BQ1.

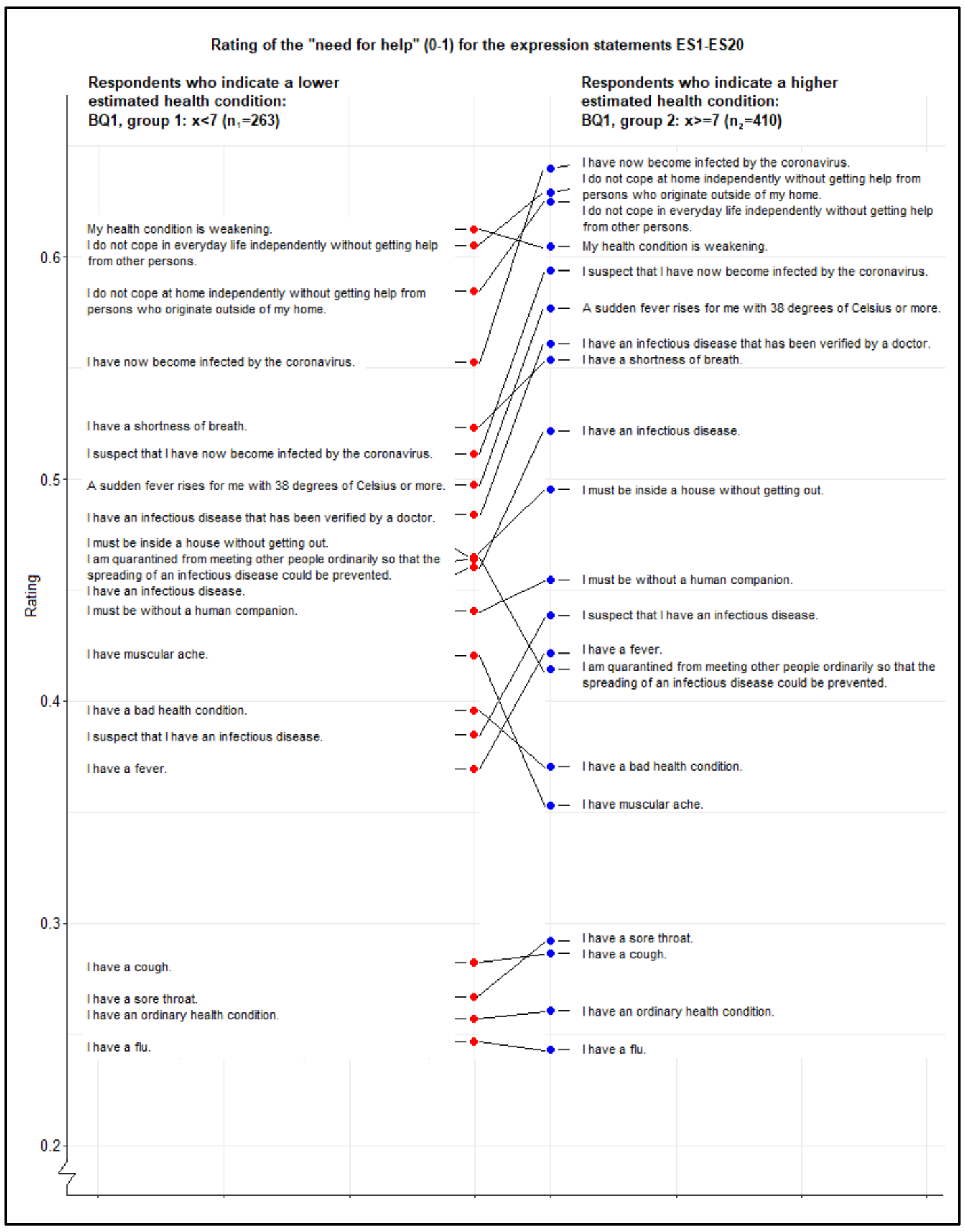

**Figure 4.** The "need for help" rating mean values of expression statements ES1-ES20 (transformed into the range 0.0-1.0) in respect to two groups based on the answer values of the background question BQ1 (in the range 1-9). The "group 1" contains those respondents who gave an answer value that was lower than 7 ($n_1$=263), and the "group 2" contains all the other respondents ($n_2$=410).

In accordance with Table 1, we now continue to describe our proposed new methodology in respect to the step 3 that consists of training and validation of a machine learning model (with a supervised learning approach) to learn the groupings concerning the "need for help" ratings. This step uses the same groupings of respondents that have been used in the step 2.

To address our main research question RQ2, we carried out machine learning experiments with a basic implementation of a convolutional neural network algorithm that we run in a TensorFlow programming environment (adapted from TensorFlow image classification tutorial, 2020). Our approach consisted of creating an image classifier using a *keras.Sequential* model with *layers.Conv2D* layers and then providing input data to the model in the form of images. We used a model consisting of three convolution blocks with a max pool layer in each of them and having on the top a fully connected layer that is activated by a *relu* activation function. We compiled our model with the *optimizers.Adam* optimizer and the *losses.SparseCategoricalCrossentropy* loss function. Table 7 describes layers of the convolutional neural network model used in the machine learning experiments.

Since the convolutional neural network model required labeled input data in the form of images, we transformed with a R language script our originally character-encoded questionnaire data into a set of grayscale raster images before feeding it to the model.

First the original rating answer values in the range 0-10 were transformed linearly into the range 0.0-1.0. Each entity of twenty rating answers (in the range 0.0-1.0) of expression statements ES1-ES20 given by a certain person were transformed into an individual raster image so that each single rating answer value was represented by a region of 25 pixels (width 5 pixels and height 5 pixels) having a brightness value in the range 0-255 directly proportional to the greatness of the transformed answer value in the range 0.0-1.0. All the twenty separate 25-pixel-sized regions were then joined as a 5×4 matrix to form a combined grayscale raster image (width 25 pixels and height 20 pixels).

**Table 7.** Layers of the convolutional neural network model used in the machine learning experiments.

| Model: "sequential"<br>*Parameters:* total 73112; trainable: 73112; non-trainable: 0 | | |
|---|---|---|
| *Layer (type)* | *Output shape* | *Number of parameters* |
| rescaling_1 (Rescaling) | (None, 20, 25, 3) | 0 |
| conv2d (Conv2D) | (None, 20, 25, 16) | 448 |
| max_pooling2d (MaxPooling2D) | (None, 10, 12, 16) | 0 |
| conv2d_1 (Conv2D) | (None, 10, 12, 32) | 4640 |
| max_pooling2d_1 (MaxPooling2D) | (None, 5, 6, 32) | 0 |
| conv2d_2 (Conv2D) | (None, 5, 6, 64) | 18496 |
| max_pooling2d_2 (MaxPooling2D) | (None, 2, 3, 64) | 0 |
| flatten (Flatten) | (None, 384) | 0 |
| dense (Dense) | (None, 128) | 49280 |
| dense_1 (Dense) | (None, 2) | 258 |

We performed machine learning experiments with labeled images so that their labeling matched the groupings that we have just previously analyzed with Wilcoxon rank-sum test (i.e., Mann–Whitney U test) between two groups and Kruskal-Wallis test between three groups to identify statistically significant rating differences (see Table 6). We allocated for the training and validation of the machine learning model 80 percent and 20 percent of the data, respectively. Table 6 shows our results about training and validation of the convolutional neural network model to learn a labeling that matches the grouping based on the answer values of each background question, among questions BQ1-BQ2 and BQ4-BQ9 (n=673). For each grouping we report training and validation metrics gained at such an epoch step when we reached the lowest value for the validation loss (ensured by further 50 evaluation steps with a patience procedure), averaged from 100 separate training and validation sequences.

Figure 5 illustrates the loss and accuracy for training and validation of the convolutional neural network model for one sequence to learn a labeling that matches the grouping of two groups based on the answer values of the background question BQ1 (n=673). In this illustrated single sequence the lowest value for the validation loss was reached at the epoch step 11 and at that step the following metrics were gained: training loss 0.53, training accuracy 0.73, validation loss 0.60 and validation accuracy 0.67.

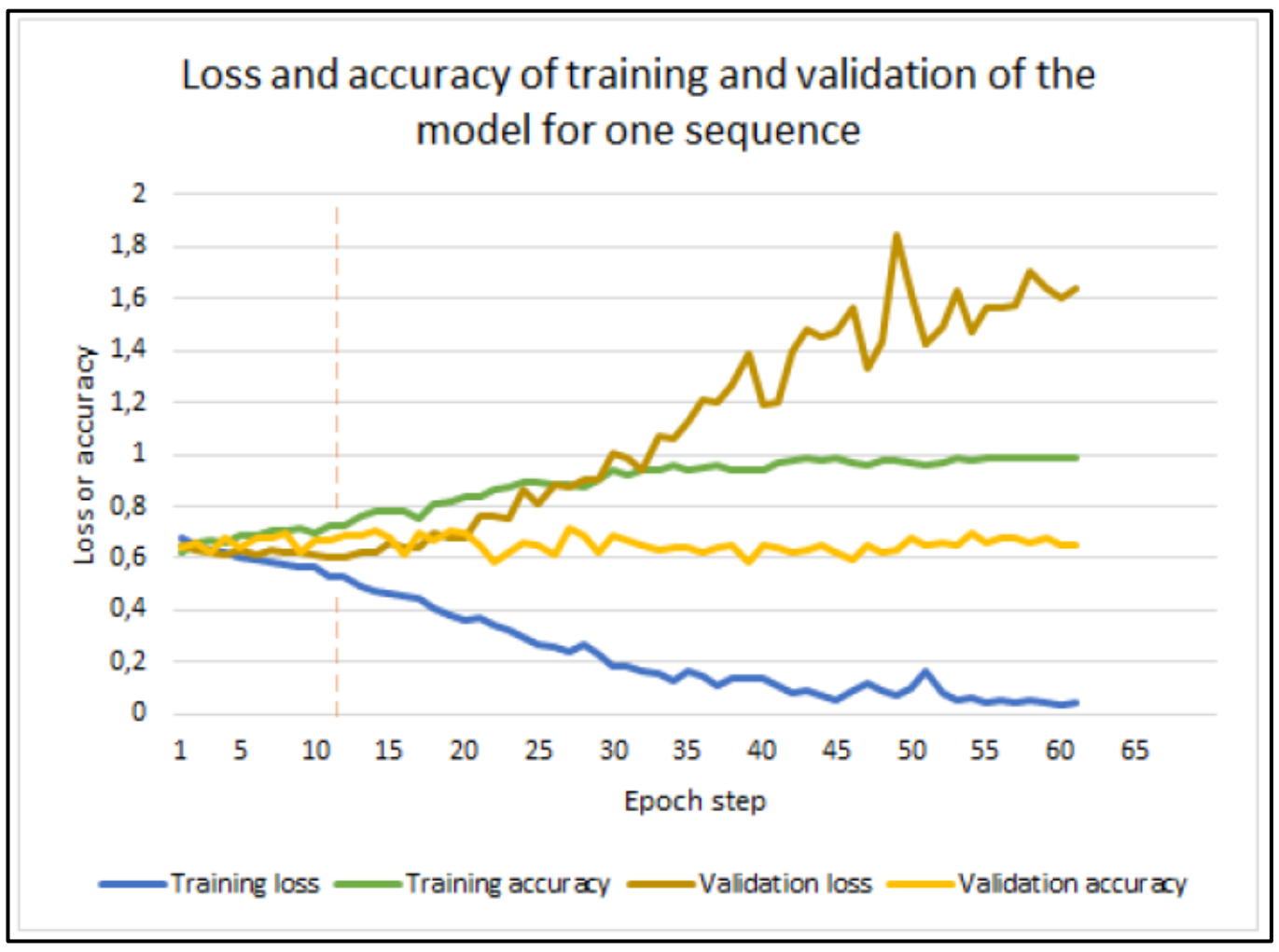

**Figure 5.** Loss and accuracy for training and validation of the convolutional neural network model for one sequence to learn a labeling that matches the grouping of two groups based on the answer values of the background question BQ1 (n=673).

In accordance with Table 1, we now continue to describe our proposed new methodology in respect to the step 4 that consists of comparing the validation accuracies of the machine learning model with the probabilities of pure chance of classifying the rating profiles correctly corresponding to groupings relying on the answer values of each background question (averaged from at least 100 separate training and validation sequences). Please see in Table 6 the two most right-sided columns. The probability of pure chance of classifying the rating profiles correctly is computed by dividing the size of the greatest group of the grouping ($n_1$, $n_2$ or $n_3$) by the number of all respondents (n=673). Then it is possible to compute the difference of the mean validation accuracy and the probability of pure chance of classifying the rating profiles correctly corresponding to each grouping. As Table 6 shows, the difference of the mean validation accuracy and the probability of pure chance of classifying the rating profiles correctly has the highest values for the groupings of two groups which are "BQ9, two groups" (0,17), "BQ1, two groups" (0,08) and "BQ6, two groups" (0,07). Furthermore, the difference has the highest values for the groupings of three groups which are "BQ9, three groups" (0,16), "BQ6, three groups" (0,03) and "BQ1, three groups" (0,03).

To describe our proposed new methodology in accordance with Table 1, the just mentioned step 4 is closely linked with the step 5 that we now continue to describe next. The step 5 consists of contrasting the validation accuracies of the machine learning model with the occurrence of statistically significant and non-significant rating differences for expression statements in respect to groupings based on the answer values of background questions (averaged from at least 100 separate training and validation sequences). We propose that this contrasting can be done intuitively by evaluating various properties of the rating differences concerning the expression statements for each grouping. These properties can include the frequencies, the strengths (levels) of statistical significance, rankings and distributions of the rating differences. We now illustrate this evaluation approach for the grouping "BQ1, two groups" as shown in Table 6.

For the grouping "BQ1, two groups" statistically significant rating differences emerge for eight expression statements which are ES6-ES10 and ES16-ES18. Among them ES6 has a statistical significance with the highest level that is $p<0.001$, ES8-ES10 have a statistical significance with the level $p<0.01$ and the remaining ES7 and ES16-ES18 have a statistical significance with the level $p<0.05$. Already these notions enable to identify rankings and distributions of the rating differences for expression statements in respect to the grouping "BQ1, two groups" based on the decreasing order of statistical significance (e.g., ES6 having the highest level) and the pattern of semantic topics of the expression statements that belong to the subset of eight expression statements that now reached statistical significance among all the 20 expression statements. Further rankings and distributions can be identified based on the values of rating differences for expression statements, for example ES6 having the highest positive rating difference value (0.07) and ES10 having the lowest negative rating difference value (-0.09), and for example the absolute values of each of the eight statistically significant rating differences being in the range of [0.05, 0.09] in a specific decreasing order.

## 4. Discussion

In accordance with the steps 1-6 of Table 1, motivated by the previous research and based on our gained findings we now discuss about implications for developing the methodology for interpretation of the patient's expressions to support his/her personalized care. The steps 1-2 of Table 1 are addressed by the main research question RQ1. In respect to our main research question RQ1, we have analyzed how different people rate the "need for help" for expression statements concerning imagined care situations related to the coronavirus COVID-19 epidemic and how this rating depends on the background information about the person.

When we computed Kendall rank-correlation measures for each comparable pair of parameter values of the "need for help" ratings, we found significant correlation (>=0.70 with the level $p<0.001$; see Akoglu, 2018) linking expression statements in five thematic subentities. These are: an infectious disease (suspecting to have an infectious disease, having it, or having it with a doctor's verification; ES16-ES18), a lack of coping independently (a lack of coping independently in everyday life or at home; ES14-ES15), the coronavirus (suspecting to have the coronavirus infection or having it; ES9-ES10), a fever (having a fever or a sudden rise of fever; ES7-ES8), and a flu/cough (having a flu or a cough; ES1-ES2). Furthermore, a significant correlation (>=0.70 with the level p<0.001; see Akoglu, 2018) linked background questions in a subentity about health (an estimated health condition or the satisfaction about health; BQ1&BQ6). The highest cosine similarity values emerging among the same value pairs seemed to support the clusters just identified by the correlation. In the same highest range of cosine similarity values, other linked expression statement pairs included having a sudden rise of fever and suspecting to have the coronavirus infection, having a shortness of breath and a weakening health condition, and having a sore throat and muscular ache.

We identified statistically significant rating differences for expression statements in respect to groupings based on the answer values of each background question, between two groups and between three groups (with Wilcoxon rank-sum test and Kruskal-Wallis test, respectively), as shown in Table 6. Supplementing tests of one-way analysis of variance (ANOVA) between groups also largely supported these findings, and indicated even some other statistically significant rating differences. To keep our analysis compact, we now discuss about the statistically significant rating differences especially in respect to Wilcoxon rank-sum test and Kruskal-Wallis test but similar notions apply well also in respect to ANOVA tests between groups.

In groupings of two groups, the highest number of statistically significant rating differences (p<0.05) emerged for the expression statements ES11 (7 groupings) and ES6 (6). The rating for ES11 differed statistically significantly for all the background questions, except BQ1, between two groups (lower answer values vs. higher answer values). The mean rating of ES11 was higher when getting lower answer values to BQ5-BQ7 ("group 1") than when getting higher answer values to BQ5-BQ7 ("group 2"). In contrast, the mean rating of ES11 was lower when getting lower answer values to BQ2, BQ4, BQ8 and BQ9 ("group 1") than when getting higher answer values to BQ2, BQ4, BQ8 and BQ9 ("group 2").

Since ES11 refers to an essential coronavirus-related situation (to be quarantined from meeting other people to prevent spreading an infectious disease), this emerging high differentiation of the "need for help" ratings can be considered as an important new finding that should be addressed when interpreting a person's need for help during an epidemic (such as the coronavirus COVID-19 epidemic). For example, the respondents who indicated a lower quality of life (BQ5, two groups) gave for ES11 a mean rating of 0.47, whereas the respondents who indicated a higher quality of life gave a mean rating of 0.41. On the other hand, the respondents who indicated a lower age (BQ9, two groups) gave for ES11 a mean rating of 0.41, whereas the respondents who indicated a higher age gave a mean rating of 0.46.

Besides groupings of two groups, ES11 and ES6 gained the highest number of statistically significant rating differences (p<0.05) also in respect to groupings of three groups (4 groupings for both ES11 and ES6). Other expression statements having a high number of statistically significant rating differences in groupings of two or three groups include ES8-ES10 (5 or 6 groupings). Since ES8-ES10 refer to an essential coronavirus-related situation (having a sudden rise of fever, suspecting to have the coronavirus infection or having it), also this emerging high differentiation of the "need for help" ratings can be considered as an important new finding that should be addressed when interpreting a person's need for help, for example to support personalized screening, diagnosis and care planning. These three expression statements ES8-ES10 gained lower mean ratings from respondent groups who indicated a lower estimated health condition (BQ1), a lower quality of life (BQ5) and being a man (BQ8), and higher mean ratings from the opposite groups, respectively.

Statistically significant rating differences (p<0.05) in groupings of two groups emerged the most for the background question BQ8 (13 expression statements), then followed by BQ9 (12), BQ1 (8), BQ5 (6), BQ2 (5), BQ7 (3), BQ4 (2), and BQ6 (2). Relatively similarly, in groupings of three groups, statistically significant rating differences (p<0.05) emerged the most for the background question BQ9 (13 expression statements), then followed by BQ5 (7), BQ1 (5), BQ6 (2), and BQ7 (2).

Figure 4 illustrates the emergence of two different ranking orders for the "need for help" ratings of expression statements ES1-ES20 in respect to the grouping based on the answer values of the background question BQ1 for two groups. Already these kinds of rankings can assist in addressing the needs of the patient depending on his/her background information, but the ratings can be exploited also in many other ways to create rankings that can support personalizing the care. Each background question is linked to a specific set of expression statements (if any) that show statistically significant rating differences for this background question. Based on the rating differences and their strengths (levels) of statistical significance, a ranking order can be identified for those expression statements that are linked to by the same background question. On the other hand, an expression statement can get different rating differences and strengths (levels) of statistical significance for different background questions (if any). This enables to identify for each expression statement a ranking order of background questions that link to it.

These various ranking orders offer an opportunity to find some distinctive link patterns between the person's "need for help" ratings for expression statements and his/her answer values to background questions, and vice versa. For example, in groupings of two groups, ES14 and ES15 show statistically significant rating differences for BQ2 (0.06 and 0.08, respectively) but not for BQ5, and on the other hand ES16 and ES17 show statistically significant rating differences for BQ5 (-0.06 and -0.07, respectively) but not for BQ2. This emerging differentiation may enable a conclusion that the "need for help" ratings about coping independently (ES14-ES15) are more closely linked to having a health problem that reduces ability (BQ2) than to the quality of life (BQ5). Similarly, it may be concluded that the "need for help" ratings about an infectious disease (ES16-ES17) are more closely linked to the quality of life (BQ5) than to having a health problem that reduces ability (BQ2).

After just discussing about the steps 1-2 of Table 1, we now continue to discuss about the steps 3-6. The steps 3-6 of Table 1 are addressed by the main research question RQ2. In respect to our main research question RQ2, we performed machine learning experiments with the answer value sets transformed to labeled raster images so that their labeling matched the groupings that we have just previously analyzed with Wilcoxon rank-sum test and Kruskal-Wallis test (as shown in Table 6). This was motivated by the assumption that machine learning enables more flexibility for modeling than for example logistic regression models of traditional statistics (Deo 2015). We trained and validated a convolutional neural network model to learn a labeling that matches the grouping. In groupings of two groups, the highest mean values of validation accuracy emerged for the background question BQ8 (0.79), then followed by BQ7 (0.72), BQ1 (0.69), BQ9 (0.68), BQ2 (0.66), BQ5 (0.60), BQ6 (0.60) and BQ4 (0.57). In groupings of three groups, the highest mean values of validation accuracy emerged for the background question BQ9 (0.50), then followed by BQ7 (0.47), BQ5 (0.42), BQ1 (0.40) and BQ6 (0.39).

As motivated in the chapters 2 and 3, due to the overall complexity of modeling semantics of a natural language and the limited size of the current data set our gained results are *not* meant to introduce a model that can actually learn the groupings very well. Instead, we aim now to propose and experimentally motivate a new methodology that can be used for analyzing how the machine learning models are influenced by the properties of the data so that these notions can be exploited to develop better machine learning models. We have chosen the specific openly available implementation of a convolutional neural network (adapted from TensorFlow image classification tutorial, 2020) as a baseline architecture to gain measures of the performance of machine learning that enable comparison between our parallel data subsets as well as offer our current results to be compared later with future experiments in a well-documented way.

Since our essential goal is to ensure generating and evaluating comparable measures concerning the machine learning experiments, we do not want to rely just on the value of validation accuracy but instead we preferably want to observe especially the difference of the mean validation accuracy and the probability of pure chance of classifying the rating profiles correctly corresponding to groupings relying on the answer values of each background question (as shown in Table 6). As described in the chapter 3, the difference of the mean validation accuracy and the probability of pure chance of classifying the rating profiles correctly has varied values for different groupings and has the highest values for the groupings of two or three groups in respect to background questions BQ9, BQ1 and BQ6 so that the difference values remain clearly above the value zero. Thus at least for the groups of these background questions BQ9, BQ1 and BQ6 the mean values of validation accuracy are clearly above the probabilities of pure chance. This in turn allows us to make a conclusion that in respect to these groupings the machine learning results may have been well-influenced by the properties of the data and possibly especially by such properties that are related to the statistically significant rating differences that we have identified with traditional statistical methods. Due to the limited size of our current data set, it is possible that various dependencies remain now unnoticed. Thus it may be possible that even those groupings that do not now reach such mean values of validation accuracy that are above the probabilities of pure chance can still in future experiments reach them when the size of the data set is increased sufficiently.

In accordance with Table 1, in our proposed new methodology a specific role is reserved for the step 6. The step 6 consists of drawing conclusions about the applicability of the current machine learning model in this knowledge context. Based on the conclusions further finetuning can be done for the model and then it is possible to iteratively repeat the steps 2-6. Since the distributional properties of the questionnaire answers can vary extensively in different cases of using the methodology, it is thus challenging to offer now any comprehensive description about the principles how the conclusions should be drawn preferably in a general case and how the finetuning and iterative evaluation could be addressed suitably. Therefore within the limited space constraints of this research article and relying on the previous research and our new experimental results, we now suggest that a general guideline for carrying out the step 6 is to emphasize parallel and complementing data analysis methods so that initial weaker findings could become gradually more verified with cumulative further analysis that cross-examines the identified dependencies and influences. Anyway, our results reported in Table 6 motivate an illustration of empirical application of the step 6 in the current case of using the methodology with our gathered experimental data that has a limited size.

Based on the just mentioned notions, we therefore suggest that although the mean values of validation accuracy remained relatively low and only partially above the values of pure chance for the groupings, our machine learning experiments however managed to show the applicability of a baseline convolutional neural network model to support detecting the need for help in the patient's expressions in respect to groupings relying on the answer values of each background question. Thus especially at least for the groupings relying on the background questions BQ9, BQ1 and BQ6 it appears that the machine learning results may be well-influenced by the statistically significant rating differences that we identified for certain specific expression statements, as shown in Table 6. These influences may be especially strong (reaching partially even the statistically significant rating differences of the level $p<0.001$) in respect to the "need for help" ratings for expression statements ES1-ES5 and ES14-ES15 (concerning BQ9), ES6 (concerning BQ1), and ES11 (concerning BQ6).

Therefore with our current data set in accordance with the step 6 of Table 1, some possible conclusions for further finetuning and iterative evaluation of the current baseline machine learning model can include for example adjusting the model's internal computational logic so that it can better address those certain specific expression statements that have been identified to influence the model's performance. The model's adjustments should preferably take into account the particular statistical and semantic properties of these expression statements. These adjustments can consist of among others modification of the model's layers, filters, pooling, optimizers, activation functions and loss functions. In addition, the adjustments can extend to cover comparing alternative machine learning architectures and their variants and hybrids as well as preprocessing options such as input data formulation and regularization and supplementing statistical or rule-based techniques.

Furthermore, with our current data set the finetuning of the machine learning model may benefit from emphasizing especially those expression statements which reached the highest statistically significant rating differences in the four thematic subentities that we identified among them, as we discussed above. Thus it may be beneficial to aim at finetuning the model to learn the groupings in respect to background questions so that for each grouping the adjustments can address especially those thematic subentities of expression statements that have the highest statistically significant rating differences for this grouping. Thus in the finetuning of the baseline model it can be possible to emphasize the following thematic subentities: having respiratory symptoms or a weakening health condition (ES1-ES5) concerning groupings in respect to BQ9; a lack of coping independently (ES14-ES15) concerning groupings in respect to BQ9; having muscular ache (ES6) concerning groupings in respect to BQ1; and being quarantined due to an infectious disease (ES11) concerning groupings in respect to BQ6.

These our notions are motivated by the previous research that has shown the applicability of an artificial neural network model in identifying the affectivity of online messages about the coronavirus (Jelodar et al., 2020) by reaching a testing accuracy of 81.15 percent in classification that relied on a training set of 338666 messages and a testing set of 112888 messages about the coronavirus extracted from the online messaging service Reddit between 20 January and 19 March 2020. Besides having a bigger data set than ours, Jelodar et al. (2020) used additional methods of Latent Dirichlet Allocation (LDA) and a pre-existing emotion vocabulary and rules (SentiStrength algorithm) to supplement a Long Short-Term Memory (LSTM) recurrent neural network (RNN) algorithm. In contrast, our results purposefully rely on using just a basic implementation of a convolutional neural network algorithm (TensorFlow image classification tutorial, 2020) that we feed with our gathered questionnaire answers (n=673).

Since we used a relatively small data set of answers and the distributions of some answer values were positioned in a relatively narrow or skewed subrange of the scale range, this may have limited the classification ability of our machine learning model. These partially narrow and skewed distributions have also caused that the probability of pure chance of classifying the rating profiles correctly has varied values for different groupings since that probability is defined based on the size of the greatest group of the grouping ($n_1$, $n_2$ or $n_3$) that reaches varying values for different groupings. This variability in turn has given the motivation that we want to observe especially the difference of the mean validation accuracy and the probability of pure chance of classifying the rating profiles correctly corresponding to groupings (as shown in Table 6).

Despite the challenges outlined above we have managed to identify some emerging link patterns between our results of machine learning and traditional statistical analysis. For two groups and three groups, the highest mean values of validation accuracy emerged for the background questions BQ8 and BQ9, respectively, which also reached the highest number of statistically significant rating differences ($p<0.05$) for expression statements in respect to the same groupings with Wilcoxon rank-sum test and Kruskal-Wallis test (see Table 6). However, the difference of the mean validation accuracy and the probability of pure chance of classifying the rating profiles correctly corresponding to groupings is now clearly above the zero only for the groupings of BQ9 and not for the grouping of BQ8. Thus we suggest making a conclusion that our machine learning results may be influenced by the statistically significant rating differences that we have identified for the groupings of BQ9 but possibly not by the statistically significant rating differences that we have identified for the grouping of BQ8.

We expect that by accumulating a larger data set of answers, it is possible to reach higher values for the difference of the mean validation accuracy and the probability of pure chance of classifying the rating profiles correctly corresponding to groupings. This in turn can enable achieving a more detailed understanding about how the machine learning results depend on and are influenced by the statistically significant rating differences concerning the groupings.

Accumulating knowledge from even sparse data points of diverse single-time interpretative measurements with machine learning gets fruitful support from the previous research that has found relatively good reliability even for single-item observations with increasing efficiency, avoiding confusion and enabling to accumulate answers from people who are hard to reach (Boateng et al., 2018; Kulikowski, 2018; Siegel & van Dolen, 2020; Tolvanen et al., 2019). We now present a new comparative analysis approach to identify and evaluate with traditional statistical methods the dependencies that can explain the machine learning results. Thus our analysis approach enables to develop better human-understandable machine learning and so helps to address the traditional challenges of interpreting reliably and intuitively machine learning results (Deo 2015). Therefore, our analysis approach can offer also support for developing reliable evaluation metrics for healthcare chatbots (Abd-Alrazaq et al., 2020) and their ability for semantic understanding (Laranjo et al., 2018).

Our results can be considered as a supplement to already existing machine learning approaches that have been applied in classification of medical literature, patient records, clinical narratives and patient phenotypes (Hughes et al., 2017; Zhao et al., 2017; Gehrmann et al., 2018; Rojas-Barahona et al., 2018; Yao et al., 2019; Qing et al., 2019; Shickel et al., 2019). However, a specific novelty in our approach is that besides gathering answers about the person's current real-life situation, we also gathered rating answers that measured the degree of the "need for help" that the person associated with the given imagined care situations. Thus with our "need for help" rating model (Lahti, 2017; Lahti 2018) we developed a new methodology that extracts the person's behavioral patterns (such as conceptualizations, attitudes and reasonings) associated with various possible future care situations depicted by expression statements. With machine learning these identified behavioral patterns are then linked to certain background information about the person thus enabling to create predictive models. For example, in the context of clinical decision support systems (CDSS), our results can assist in detecting the patient's need for help and thus enhance reasoning that addresses distinctive and differentiated needs of the patient to enable personalized screening, diagnosis and care planning. Also in self-care and rehabilitation, our results can assist to implement monitoring and recording of the emerging need for help in the person's everyday life so that necessary assistance can be alerted.

We decided to gather now ratings in respect to the "need for help" since this semantic dimension emerged strongly in the context of health-related online discussions in our previous analysis (Lahti et al., 2018). However, the selection of the "need for help" dimension can be motivated also by its intuitive relatedness to the dominance dimension (Bradley & Lang, 1999a; Warriner et al., 2013) that reflects the degree of ability to cope and to be in the control of one's own life situations, and also to the approach-avoidance dimension (Mauss & Robinson, 2009) that reflects the desire to reach some relieving assistance or to be reached by this assistance.

Our results indicated statistically significant rating differences depending on the person's sex and age that can be considered to get support from corresponding previous results (Warriner et al., 2013) in which female and older respondents gave on average smaller rating values of pleasure, arousal and dominance than male and younger respondents, respectively, for a diverse set of words. Furthermore, our results concerning statistically significant rating differences depending on the person's health and wellbeing get support from the previous findings of Warriner et al. (2013) in which the most feared medical conditions were also rated to be among the diseases that represent the lowest rating values of pleasure and dominance and the highest rating values of arousal.

To measure the "need for help" ratings the most reliably, the measurements should be done in real-life situations that involve negative experiences but since that is ethically challenging, we now measured the "need for help" with imagined situations. Anyway, experimental setups containing real-life exposure to pain and threats to pain (Sullivan et al., 1995) indicated that helplessness correlated highly with rumination and moderately with magnification. Since this previous result has resemblance with our significant correlation ($>=0.70$ with the level $p<0.001$; see Akoglu, 2018) between ratings of suspecting to have the coronavirus infection or having it (ES9-ES10) and between ratings of suspecting to have an infectious disease, having it, or having it with a doctor's verification (ES16-ES18), this offers support that our measurements of imagined situations can indeed be relatively reliably paralleled with real-life situations. In addition, Berna et al. (2011) have found links between self-identified most significant mental imagery describing the patient's pain and associated triggers, affects, meanings and avoidance patterns.

Our aim to generalize imaginary-based measurement results to corresponding real-life situations gets also support from the previous findings that the patterns of neural activation during imagery and actual perception have a strong overlap (Ganis et al., 2004; McNorgan, 2012). Neuroimaging experiments have indicated that self-report ratings of vividness of mental imagery can correlate with activation of the same sensory-specific cortices as activated in perception (Cui et al., 2007; Herholz et al., 2012; Belardinelli et al., 2009). Anyway, there is evidence that imagining a future event increases the person's perception concerning the probability that the imagined event will occur (Carroll, 1978, Sherman et al., 1985). It has been also shown that people perceive the likelihood of contracting a disease higher when the description of the disease is easier to imagine than when it is harder to imagine (Sherman et al., 1985), and for imagined symptoms people prioritized selecting a simple separate cause than a more complex combination of causes even if the likelihood value for the combination of all the causes was displayed to be higher than for simple separate causes (Lombrozo 2007). These previously found adjusting effects on probabilities and prioritization concerning imagining and reasoning may contribute also to the patterns of dependence and influence that we have now identified between our machine learning results and statistically significant rating differences.

## Conclusions

With our new methodology (see Table 1) statistically significant differences of self-rated "need for help" can be linked to machine learning results. We found statistically significant correlations and high cosine similarity values between various health-related expression statement pairs concerning the "need for help" ratings and a background question pair. We also identified statistically significant rating differences for several health-related expression statements in respect to groupings based on the answer values of background questions, such as the ratings of suspecting to have the coronavirus infection and having it depending on the estimated health condition, quality of life and sex. Our new methodology enabled us to identify how some of the statistically significant rating differences may be linked to machine learning results thus helping to develop better human-understandable machine learning models.

Resembling the previous research that has developed machine learning methods for extracting health-related knowledge (Hughes et al., 2017; Zhao et al., 2017; Gehrmann et al., 2018; Rojas-Barahona et al., 2018; Yao et al., 2019; Qing et al., 2019; Shickel et al., 2019) and evaluated the affectivity of online messages about the coronavirus (Jelodar et al., 2020), our results offer insight about the applicability of machine learning to extract useful knowledge from health-related expression statements to support healthcare services, such as to provide personalized screening and care. However, to our best knowledge our research is the first of its kind to develop and use the "need for help" rating model (Lahti, 2017; Lahti, 2018) to gather self-rated interpretations about health-related expression statements that are then analyzed to identify statistically significant rating differences in respect to groupings based on the answer values of background questions, and then also to show the applicability of machine learning to learn the groupings concerning the ratings. Furthermore, with our new methodology we propose and experimentally motivate how to enable comparable measurements between parallel data subsets as well as for future experiments in a well-documented way.Our research contribution gets some additional value also from the successful data acquisition process that involved respondents belonging to Finnish patient and disabled people's organizations, other health-related organizations and professionals, and educational institutions (n=673) and thus representing a diversity of health conditions, abilities and attitudes. In addition, our results enable to compare the statistically significant rating differences in groupings in respect to the person's background information and to further contrast them with the training and validation metrics gained in machine learning experiments based on the same groupings (see Table 6). We also publish an open access data set about our measurements and results in the supplementing Appendix A.

Future research should continue exploring and analyzing how different people interpret and evaluate health-related expression statements and how this possibly depends on the person's background information. A specific emphasis should be given for developing adaptive modular methods that can be flexibly applied for various purposes of health analytics and also enhance fertile standardized practices that ensure comparability. Furthermore, the emerging new methods and algorithms should be well human-understandable for everyone and provided with open access, accompanied with appropriately and sufficiently anonymized data sets. In this spirit, we suggest that also our current findings and results can be used as a part of a greater reasoning entity to develop computational methods to identify, interpret and address the needs of the patient in diverse knowledge processes of healthcare to support personalized care.

## Abbreviations

**ANOVA:** analysis of variance
**BQ:** background question
**CDSS:** clinical decision support system
**COVID-19:** coronavirus disease 2019
**DIHEML:** the research project "Development of method for interpretation of health expressions based on machine learning to support various care events and persons"
**ES:** expression statement
**LDA:** Latent Dirichlet Allocation
**LSTM:** Long Short-Term Memory
**M:** mean
**Mdn:** median
**RNN:** recurrent neural network
**RQ:** research question
**SD:** standard deviation
**THL:** Finnish Institute for Health and Welfare (Terveyden ja hyvinvoinnin laitos, THL)
**WHO:** World Health Organization

xxxxxxxxxxxxxxxxxxxxxxxxxxxxxxxxxxxxxxxxxxxxxxxxxxxxxxxxxxxxxxxx

## Appendix A for the research article "Detecting the patient's need for help with machine learning"

Lauri Lahti, Department of Computer Science, Aalto University, Finland, lauri.lahti@aalto.fi, version 22 March 2021

**Table A1. Kendall rank-correlation and cosine similarity measures for each comparable pair of parameter values of the "need for help" ratings of expression statements ES1-ES20 and the answers of the background questions BQ1 and BQ5-BQ7 (n=673).**

Kendall rank-correlation measures are shown on the upper-right region of the table and cosine similarity measures are shown on the lower-left region of the table.

Before computing cosine similarity measures the answer values of each parameter were normalized by the formula *(x - min(x))/(max(x)-min(x))* and then these new values were shifted so that the mean value was positioned to the zero by the formula *(x - mean(x))*. The statistical significance levels were defined as p<0,05, p<0,01 and p<0,001, denoted by symbols *, ** and ***, respectively.

| | *ES1* | *ES2* | *ES3* | *ES4* | *ES5* | *ES6* | *ES7* | *ES8* | *ES9* | *ES10* | *ES11* | *ES12* | *ES13* | *ES14* | *ES15* | *ES16* | *ES17* | *ES18* | *ES19* | *ES20* | *BQ1* | *BQ5* | *BQ6* | *BQ7* |
|---|---|---|---|---|---|---|---|---|---|---|---|---|---|---|---|---|---|---|---|---|---|---|---|---|
| *ES1* | | 0.73 *** | 0.40 *** | 0.32 *** | 0.55 *** | 0.40 *** | 0.40 *** | 0.33 *** | 0.26 *** | 0.26 *** | 0.16 *** | 0.22 *** | 0.24 *** | 0.23 *** | 0.22 *** | 0.24 *** | 0.23 *** | 0.28 *** | 0.36 *** | 0.16 *** | -0.01 | -0.01 | -0.03 | -0.03 |
| *ES2* | 0.80 | | 0.56 *** | 0.43 *** | 0.64 *** | 0.45 *** | 0.51 *** | 0.43 *** | 0.38 *** | 0.36 *** | 0.19 *** | 0.30 *** | 0.30 *** | 0.32 *** | 0.31 *** | 0.31 *** | 0.30 *** | 0.34 *** | 0.40 *** | 0.07 * | 0.00 | 0.00 | -0.01 | -0.03 |
| *ES3* | 0.44 | 0.65 | | 0.67 *** | 0.52 *** | 0.34 *** | 0.56 *** | 0.55 *** | 0.51 *** | 0.51 *** | 0.20 *** | 0.40 *** | 0.34 *** | 0.51 *** | 0.52 *** | 0.43 *** | 0.43 *** | 0.43 *** | 0.41 *** | -0.13 *** | 0.03 | 0.06 * | 0.01 | 0.01 |
| *ES4* | 0.35 | 0.52 | 0.83 | | 0.50 *** | 0.38 *** | 0.57 *** | 0.59 *** | 0.56 *** | 0.55 *** | 0.24 *** | 0.43 *** | 0.38 *** | 0.52 *** | 0.53 *** | 0.45 *** | 0.44 *** | 0.44 *** | 0.44 *** | -0.13 *** | 0.01 | 0.04 | -0.03 | -0.04 |
| *ES5* | 0.60 | 0.73 | 0.62 | 0.62 | | 0.57 *** | 0.65 *** | 0.57 *** | 0.51 *** | 0.49 *** | 0.26 *** | 0.38 *** | 0.38 *** | 0.38 *** | 0.38 *** | 0.45 *** | 0.44 *** | 0.47 *** | 0.47 *** | 0.05 | 0.02 | 0.02 | 0.01 | -0.02 |
| *ES6* | 0.45 | 0.54 | 0.43 | 0.49 | 0.66 | | 0.49 *** | 0.40 *** | 0.35 *** | 0.31 *** | 0.23 *** | 0.26 *** | 0.30 *** | 0.25 *** | 0.24 *** | 0.28 *** | 0.27 *** | 0.30 *** | 0.40 *** | 0.12 *** | -0.11 *** | -0.10 *** | -0.11 *** | -0.11 *** |
| *ES7* | 0.43 | 0.59 | 0.71 | 0.74 | 0.75 | 0.57 | | 0.75 *** | 0.62 *** | 0.61 *** | 0.27 *** | 0.45 *** | 0.41 *** | 0.46 *** | 0.47 *** | 0.53 *** | 0.53 *** | 0.54 *** | 0.49 *** | -0.06 * | 0.05 | 0.05 | 0.03 | -0.01 |
| *ES8* | 0.33 | 0.49 | 0.73 | 0.78 | 0.65 | 0.43 | 0.87 | | 0.67 *** | 0.64 *** | 0.28 *** | 0.47 *** | 0.45 *** | 0.49 *** | 0.51 *** | 0.55 *** | 0.55 *** | 0.54 *** | 0.46 *** | -0.13 *** | 0.08 ** | 0.07 * | 0.04 | 0.01 |
| *ES9* | 0.27 | 0.44 | 0.70 | 0.76 | 0.60 | 0.40 | 0.78 | 0.87 | | 0.79 *** | 0.28 *** | 0.47 *** | 0.43 *** | 0.49 *** | 0.51 *** | 0.61 *** | 0.60 *** | 0.63 *** | 0.46 *** | -0.18 *** | 0.07 * | 0.10 *** | 0.05 | 0.02 |
| *ES10* | 0.27 | 0.43 | 0.71 | 0.76 | 0.58 | 0.36 | 0.77 | 0.86 | 0.92 | | 0.28 *** | 0.47 *** | 0.42 *** | 0.48 *** | 0.52 *** | 0.64 *** | 0.65 *** | 0.60 *** | 0.43 *** | -0.21 *** | 0.07 * | 0.10 *** | 0.05 | 0.02 |
| *ES11* | 0.20 | 0.24 | 0.27 | 0.31 | 0.34 | 0.28 | 0.35 | 0.36 | 0.37 | 0.35 | | 0.56 *** | 0.49 *** | 0.32 *** | 0.31 *** | 0.35 *** | 0.33 *** | 0.35 *** | 0.33 *** | 0.06 * | -0.08 ** | -0.07 * | -0.10 *** | -0.10 ** |
| *ES12* | 0.25 | 0.37 | 0.55 | 0.58 | 0.48 | 0.33 | 0.59 | 0.64 | 0.64 | 0.63 | 0.66 | | 0.65 *** | 0.57 *** | 0.54 *** | 0.51 *** | 0.50 *** | 0.51 *** | 0.43 *** | -0.09 ** | 0.01 | 0.01 | -0.02 | -0.05 |
| *ES13* | 0.27 | 0.36 | 0.46 | 0.51 | 0.46 | 0.37 | 0.52 | 0.58 | 0.57 | 0.55 | 0.60 | 0.77 | | 0.51 *** | 0.49 *** | 0.49 *** | 0.47 *** | 0.50 *** | 0.47 *** | -0.02 | 0.00 | 0.01 | -0.02 | -0.04 |
| *ES14* | 0.27 | 0.41 | 0.69 | 0.70 | 0.49 | 0.34 | 0.62 | 0.69 | 0.69 | 0.68 | 0.38 | 0.69 | 0.62 | | 0.86 *** | 0.51 *** | 0.49 *** | 0.50 *** | 0.42 *** | -0.16 *** | 0.02 | 0.04 | 0.00 | -0.04 |
| *ES15* | 0.26 | 0.38 | 0.70 | 0.69 | 0.48 | 0.30 | 0.63 | 0.71 | 0.71 | 0.71 | 0.36 | 0.67 | 0.60 | 0.95 | | 0.54 *** | 0.52 *** | 0.52 *** | 0.40 *** | -0.18 *** | 0.04 | 0.05 | 0.01 | -0.02 |
| *ES16* | 0.27 | 0.39 | 0.60 | 0.64 | 0.55 | 0.35 | 0.68 | 0.75 | 0.79 | 0.80 | 0.44 | 0.66 | 0.62 | 0.70 | 0.72 | | 0.91 *** | 0.78 *** | 0.50 *** | -0.14 *** | 0.04 | 0.07 * | 0.02 | -0.01 |
| *ES17* | 0.26 | 0.39 | 0.62 | 0.65 | 0.55 | 0.35 | 0.69 | 0.77 | 0.81 | 0.82 | 0.42 | 0.66 | 0.61 | 0.70 | 0.71 | 0.97 | | 0.77 *** | 0.49 *** | -0.15 *** | 0.05 | 0.07 * | 0.02 | -0.01 |
| *ES18* | 0.31 | 0.41 | 0.58 | 0.61 | 0.57 | 0.37 | 0.67 | 0.72 | 0.78 | 0.75 | 0.44 | 0.65 | 0.63 | 0.67 | 0.68 | 0.90 | 0.89 | | 0.58 *** | -0.07 * | 0.04 | 0.06 | 0.02 | -0.02 |
| *ES19* | 0.43 | 0.48 | 0.53 | 0.57 | 0.57 | 0.50 | 0.59 | 0.58 | 0.58 | 0.55 | 0.42 | 0.55 | 0.58 | 0.55 | 0.52 | 0.62 | 0.62 | 0.70 | | 0.03 | -0.06 | -0.05 | -0.07 * | -0.08 ** |
| *ES20* | 0.13 | -0.01 | -0.31 | -0.35 | -0.06 | 0.04 | -0.23 | -0.36 | -0.40 | -0.42 | -0.02 | -0.24 | -0.17 | -0.36 | -0.37 | -0.36 | -0.36 | -0.28 | -0.15 | | -0.01 | -0.05 | 0.00 | -0.02 |
| *BQ1* | 0.00 | 0.01 | 0.03 | -0.04 | 0.03 | -0.15 | 0.08 | 0.11 | 0.10 | 0.09 | -0.09 | 0.01 | 0.01 | 0.00 | 0.04 | 0.05 | 0.07 | 0.06 | -0.08 | 0.03 | | 0.63 *** | 0.71 *** | 0.57 *** |
| *BQ5* | -0.03 | 0.00 | 0.08 | 0.02 | 0.02 | -0.11 | 0.08 | 0.11 | 0.13 | 0.12 | -0.09 | 0.02 | 0.02 | 0.03 | 0.05 | 0.09 | 0.10 | 0.09 | -0.05 | -0.01 | 0.75 | | 0.68 *** | 0.58 *** |
| *BQ6* | -0.04 | -0.02 | 0.00 | -0.08 | 0.00 | -0.15 | 0.04 | 0.07 | 0.06 | 0.05 | -0.11 | -0.02 | -0.02 | -0.02 | 0.01 | 0.03 | 0.03 | 0.03 | -0.11 | 0.04 | 0.82 | 0.80 | | 0.63 *** |
| *BQ7* | -0.03 | -0.03 | 0.01 | -0.09 | -0.01 | -0.15 | 0.00 | 0.03 | 0.03 | 0.03 | -0.10 | -0.05 | -0.04 | -0.07 | -0.02 | 0.00 | 0.00 | 0.00 | -0.11 | 0.02 | 0.70 | 0.71 | 0.77 | |

**Table A2. The mean values of the "need for help" ratings of each expression statement (ES1-ES20) in respect to groupings based on the answer values of the background question (BQ), for two groups or three groups.**

$M_1$, $M_2$ and $M_3$ show the mean values for each group and the number of persons is denoted by $n_1$, $n_2$ and $n_3$ (n=673). We computed Wilcoxon rank-sum test (i.e., Mann–Whitney U test) between two groups and Kruskal-Wallis test between three groups to identify statistically significant rating differences at significance levels p<0,05, p<0,01 and p<0,001, denoted by symbols *, ** and ***, respectively.

| *Grouping based on the answer value of the background question* | *ES1* | *ES2* | *ES3* | *ES4* | *ES5* | *ES6* | *ES7* | *ES8* | *ES9* | *ES10* | *ES11* | *ES12* | *ES13* | *ES14* | *ES15* | *ES16* | *ES17* | *ES18* | *ES19* | *ES20* |
|---|---|---|---|---|---|---|---|---|---|---|---|---|---|---|---|---|---|---|---|---|
| BQ1, two groups: x<7 ($n_1$=263), x>=7 ($n_2$=410) | $M_1$=0.247<br>$M_2$=0.243 | $M_1$=0.282<br>$M_2$=0.286 | $M_1$=0.523<br>$M_2$=0.553 | $M_1$=0.612<br>$M_2$=0.604 | $M_1$=0.267<br>$M_2$=0.292 | $M_1$=0.421<br>$M_2$=0.353 *** | $M_1$=0.370<br>$M_2$=0.422 * | $M_1$=0.497<br>$M_2$=0.577 ** | $M_1$=0.511<br>$M_2$=0.594 ** | $M_1$=0.552<br>$M_2$=0.640 ** | $M_1$=0.464<br>$M_2$=0.414 | $M_1$=0.465<br>$M_2$=0.495 | $M_1$=0.440<br>$M_2$=0.454 | $M_1$=0.605<br>$M_2$=0.625 | $M_1$=0.584<br>$M_2$=0.629 | $M_1$=0.460<br>$M_2$=0.521 * | $M_1$=0.484<br>$M_2$=0.560 * | $M_1$=0.385<br>$M_2$=0.439 * | $M_1$=0.396<br>$M_2$=0.370 | $M_1$=0.257<br>$M_2$=0.261 |
| BQ1, three groups: x<6 ($n_1$= 218), 6<=x<8 ($n_2$=207), x>=8 ($n_3$=248) | $M_1$=0.238<br>$M_2$=0.256<br>$M_3$=0.241 | $M_1$=0.271<br>$M_2$=0.307<br>$M_3$=0.278 | $M_1$=0.522<br>$M_2$=0.550<br>$M_3$=0.552 | $M_1$=0.622<br>$M_2$=0.600<br>$M_3$=0.601 | $M_1$=0.264<br>$M_2$=0.312<br>$M_3$=0.273 * | $M_1$=0.422<br>$M_2$=0.383<br>$M_3$=0.339 ** | $M_1$=0.370<br>$M_2$=0.414<br>$M_3$=0.418 | $M_1$=0.493<br>$M_2$=0.567<br>$M_3$=0.575 * | $M_1$=0.506<br>$M_2$=0.584<br>$M_3$=0.591 * | $M_1$=0.551<br>$M_2$=0.624<br>$M_3$=0.638 | $M_1$=0.465<br>$M_2$=0.457<br>$M_3$=0.386 * | $M_1$=0.473<br>$M_2$=0.492<br>$M_3$=0.485 | $M_1$=0.437<br>$M_2$=0.463<br>$M_3$=0.448 | $M_1$=0.616<br>$M_2$=0.604<br>$M_3$=0.628 | $M_1$=0.593<br>$M_2$=0.601<br>$M_3$=0.635 | $M_1$=0.469<br>$M_2$=0.514<br>$M_3$=0.509 | $M_1$=0.488<br>$M_2$=0.558<br>$M_3$=0.546 | $M_1$=0.385<br>$M_2$=0.442<br>$M_3$=0.426 | $M_1$=0.400<br>$M_2$=0.392<br>$M_3$=0.352 | $M_1$=0.246<br>$M_2$=0.279<br>$M_3$=0.255 |
| BQ2, two groups: x<2 ($n_1$=219), x>=2 ($n_2$=454) | $M_1$=0.253<br>$M_2$=0.241 | $M_1$=0.300<br>$M_2$=0.277 | $M_1$=0.583<br>$M_2$=0.521 * | $M_1$=0.622<br>$M_2$=0.600 | $M_1$=0.288<br>$M_2$=0.279 | $M_1$=0.337<br>$M_2$=0.400 ** | $M_1$=0.421<br>$M_2$=0.392 | $M_1$=0.580<br>$M_2$=0.529 | $M_1$=0.615<br>$M_2$=0.536 | $M_1$=0.660<br>$M_2$=0.579 | $M_1$=0.378<br>$M_2$=0.461 *** | $M_1$=0.477<br>$M_2$=0.487 | $M_1$=0.463<br>$M_2$=0.442 | $M_1$=0.655<br>$M_2$=0.599 * | $M_1$=0.663<br>$M_2$=0.587 * | $M_1$=0.534<br>$M_2$=0.480 | $M_1$=0.573<br>$M_2$=0.511 | $M_1$=0.448<br>$M_2$=0.403 | $M_1$=0.369<br>$M_2$=0.386 | $M_1$=0.233<br>$M_2$=0.272 |
| BQ4, two groups: x<2 ($n_1$=364), x>=2 ($n_2$=309) | $M_1$=0.249<br>$M_2$=0.239 | $M_1$=0.290<br>$M_2$=0.278 | $M_1$=0.549<br>$M_2$=0.533 | $M_1$=0.615<br>$M_2$=0.599 | $M_1$=0.285<br>$M_2$=0.279 | $M_1$=0.354<br>$M_2$=0.409 ** | $M_1$=0.413<br>$M_2$=0.388 | $M_1$=0.570<br>$M_2$=0.517 | $M_1$=0.580<br>$M_2$=0.540 | $M_1$=0.633<br>$M_2$=0.574 | $M_1$=0.408<br>$M_2$=0.464 * | $M_1$=0.481<br>$M_2$=0.486 | $M_1$=0.464<br>$M_2$=0.431 | $M_1$=0.630<br>$M_2$=0.601 | $M_1$=0.635<br>$M_2$=0.583 | $M_1$=0.512<br>$M_2$=0.480 | $M_1$=0.551<br>$M_2$=0.507 | $M_1$=0.428<br>$M_2$=0.405 | $M_1$=0.373<br>$M_2$=0.389 | $M_1$=0.252<br>$M_2$=0.268 |
| BQ5, two groups: x<7 ($n_1$=274), x>=7 ($n_2$=399) | $M_1$=0.252<br>$M_2$=0.239 | $M_1$=0.287<br>$M_2$=0.283 | $M_1$=0.521<br>$M_2$=0.556 | $M_1$=0.603<br>$M_2$=0.611 | $M_1$=0.274<br>$M_2$=0.288 | $M_1$=0.415<br>$M_2$=0.355 ** | $M_1$=0.378<br>$M_2$=0.417 | $M_1$=0.510<br>$M_2$=0.570 | $M_1$=0.515<br>$M_2$=0.593 ** | $M_1$=0.555<br>$M_2$=0.640 ** | $M_1$=0.470<br>$M_2$=0.409 * | $M_1$=0.477<br>$M_2$=0.487 | $M_1$=0.441<br>$M_2$=0.454 | $M_1$=0.598<br>$M_2$=0.630 | $M_1$=0.584<br>$M_2$=0.630 | $M_1$=0.459<br>$M_2$=0.523 * | $M_1$=0.491<br>$M_2$=0.558 * | $M_1$=0.391<br>$M_2$=0.436 | $M_1$=0.404<br>$M_2$=0.364 | $M_1$=0.255<br>$M_2$=0.262 |
| BQ5, three groups: x<6 ($n_1$=190), 6<=x<8 ($n_2$=271), x>=8 ($n_3$=212) | $M_1$=0.242<br>$M_2$=0.257<br>$M_3$=0.232 | $M_1$=0.273<br>$M_2$=0.296<br>$M_3$=0.280 | $M_1$=0.496<br>$M_2$=0.554<br>$M_3$=0.567 | $M_1$=0.594<br>$M_2$=0.606<br>$M_3$=0.621 | $M_1$=0.262<br>$M_2$=0.304<br>$M_3$=0.273 | $M_1$=0.412<br>$M_2$=0.389<br>$M_3$=0.338 ** | $M_1$=0.362<br>$M_2$=0.414<br>$M_3$=0.420 | $M_1$=0.476<br>$M_2$=0.565<br>$M_3$=0.583 * | $M_1$=0.478<br>$M_2$=0.571<br>$M_3$=0.624 *** | $M_1$=0.520<br>$M_2$=0.614<br>$M_3$=0.671 *** | $M_1$=0.461<br>$M_2$=0.441<br>$M_3$=0.400 | $M_1$=0.464<br>$M_2$=0.495<br>$M_3$=0.486 | $M_1$=0.419<br>$M_2$=0.472<br>$M_3$=0.446 | $M_1$=0.589<br>$M_2$=0.617<br>$M_3$=0.641 | $M_1$=0.562<br>$M_2$=0.619<br>$M_3$=0.646 | $M_1$=0.435<br>$M_2$=0.517<br>$M_3$=0.528 * | $M_1$=0.467<br>$M_2$=0.546<br>$M_3$=0.568 * | $M_1$=0.373<br>$M_2$=0.431<br>$M_3$=0.440 | $M_1$=0.402<br>$M_2$=0.381<br>$M_3$=0.360 | $M_1$=0.265<br>$M_2$=0.279<br>$M_3$=0.230 * |
| BQ6, two groups: x<7 ($n_1$=318), x>=7 ($n_2$=355) | $M_1$=0.247<br>$M_2$=0.243 | $M_1$=0.280<br>$M_2$=0.289 | $M_1$=0.536<br>$M_2$=0.546 | $M_1$=0.621<br>$M_2$=0.595 | $M_1$=0.280<br>$M_2$=0.284 | $M_1$=0.409<br>$M_2$=0.353 ** | $M_1$=0.387<br>$M_2$=0.414 | $M_1$=0.522<br>$M_2$=0.567 | $M_1$=0.541<br>$M_2$=0.580 | $M_1$=0.592<br>$M_2$=0.618 | $M_1$=0.478<br>$M_2$=0.394 *** | $M_1$=0.489<br>$M_2$=0.479 | $M_1$=0.448<br>$M_2$=0.450 | $M_1$=0.620<br>$M_2$=0.614 | $M_1$=0.609<br>$M_2$=0.614 | $M_1$=0.486<br>$M_2$=0.508 | $M_1$=0.519<br>$M_2$=0.542 | $M_1$=0.412<br>$M_2$=0.422 | $M_1$=0.403<br>$M_2$=0.360 | $M_1$=0.250<br>$M_2$=0.268 |
| BQ6, three groups: x<6 ($n_1$=240), 6<=x<8 ($n_2$=229), x>=8 ($n_3$=204) | $M_1$=0.256<br>$M_2$=0.241<br>$M_3$=0.235 | $M_1$=0.285<br>$M_2$=0.286<br>$M_3$=0.282 | $M_1$=0.531<br>$M_2$=0.549<br>$M_3$=0.545 | $M_1$=0.628<br>$M_2$=0.598<br>$M_3$=0.594 | $M_1$=0.270<br>$M_2$=0.293<br>$M_3$=0.284 | $M_1$=0.414<br>$M_2$=0.374<br>$M_3$=0.345 * | $M_1$=0.381<br>$M_2$=0.401<br>$M_3$=0.425 | $M_1$=0.506<br>$M_2$=0.565<br>$M_3$=0.571 | $M_1$=0.523<br>$M_2$=0.580<br>$M_3$=0.586 | $M_1$=0.573<br>$M_2$=0.616<br>$M_3$=0.632 | $M_1$=0.470<br>$M_2$=0.443<br>$M_3$=0.380 ** | $M_1$=0.483<br>$M_2$=0.491<br>$M_3$=0.475 | $M_1$=0.439<br>$M_2$=0.465<br>$M_3$=0.443 | $M_1$=0.616<br>$M_2$=0.611<br>$M_3$=0.625 | $M_1$=0.595<br>$M_2$=0.610<br>$M_3$=0.631 | $M_1$=0.480<br>$M_2$=0.503<br>$M_3$=0.511 | $M_1$=0.513<br>$M_2$=0.540<br>$M_3$=0.542 | $M_1$=0.403<br>$M_2$=0.420<br>$M_3$=0.432 | $M_1$=0.405<br>$M_2$=0.373<br>$M_3$=0.360 | $M_1$=0.258<br>$M_2$=0.251<br>$M_3$=0.270 |
| BQ7, two groups: x<7 ($n_1$=201), x>=7 ($n_2$=472) | $M_1$=0.262<br>$M_2$=0.237 | $M_1$=0.300<br>$M_2$=0.278 | $M_1$=0.550<br>$M_2$=0.538 | $M_1$=0.657<br>$M_2$=0.586 | $M_1$=0.287<br>$M_2$=0.280 | $M_1$=0.434<br>$M_2$=0.356 *** | $M_1$=0.405<br>$M_2$=0.400 | $M_1$=0.536<br>$M_2$=0.550 | $M_1$=0.541<br>$M_2$=0.570 | $M_1$=0.590<br>$M_2$=0.613 | $M_1$=0.484<br>$M_2$=0.412 ** | $M_1$=0.513<br>$M_2$=0.471 | $M_1$=0.466<br>$M_2$=0.442 | $M_1$=0.655<br>$M_2$=0.601 | $M_1$=0.622<br>$M_2$=0.607 | $M_1$=0.492<br>$M_2$=0.500 | $M_1$=0.533<br>$M_2$=0.530 | $M_1$=0.419<br>$M_2$=0.417 | $M_1$=0.427<br>$M_2$=0.360 ** | $M_1$=0.259<br>$M_2$=0.259 |
| BQ7, three groups: x<6 ($n_1$=143), 6<=x<8 ($n_2$=214), x>=8 ($n_3$=316) | $M_1$=0.243<br>$M_2$=0.250<br>$M_3$=0.241 | $M_1$=0.281<br>$M_2$=0.291<br>$M_3$=0.282 | $M_1$=0.520<br>$M_2$=0.549<br>$M_3$=0.547 | $M_1$=0.659<br>$M_2$=0.597<br>$M_3$=0.591 | $M_1$=0.278<br>$M_2$=0.292<br>$M_3$=0.278 | $M_1$=0.432<br>$M_2$=0.393<br>$M_3$=0.347 ** | $M_1$=0.400<br>$M_2$=0.400<br>$M_3$=0.403 | $M_1$=0.520<br>$M_2$=0.550<br>$M_3$=0.554 | $M_1$=0.528<br>$M_2$=0.567<br>$M_3$=0.573 | $M_1$=0.589<br>$M_2$=0.599<br>$M_3$=0.618 | $M_1$=0.487<br>$M_2$=0.460<br>$M_3$=0.392 ** | $M_1$=0.505<br>$M_2$=0.498<br>$M_3$=0.464 | $M_1$=0.457<br>$M_2$=0.459<br>$M_3$=0.438 | $M_1$=0.648<br>$M_2$=0.621<br>$M_3$=0.600 | $M_1$=0.609<br>$M_2$=0.627<br>$M_3$=0.602 | $M_1$=0.494<br>$M_2$=0.503<br>$M_3$=0.495 | $M_1$=0.536<br>$M_2$=0.529<br>$M_3$=0.530 | $M_1$=0.410<br>$M_2$=0.431<br>$M_3$=0.411 | $M_1$=0.420<br>$M_2$=0.382<br>$M_3$=0.361 | $M_1$=0.248<br>$M_2$=0.269<br>$M_3$=0.258 |
| BQ8, two groups: x<2 ($n_1$=123), x>=2 ($n_2$=550) | $M_1$=0.241<br>$M_2$=0.245 | $M_1$=0.269<br>$M_2$=0.288 | $M_1$=0.460<br>$M_2$=0.560 ** | $M_1$=0.516<br>$M_2$=0.628 *** | $M_1$=0.284<br>$M_2$=0.282 | $M_1$=0.390<br>$M_2$=0.377 | $M_1$=0.359<br>$M_2$=0.411 | $M_1$=0.474<br>$M_2$=0.562 * | $M_1$=0.477<br>$M_2$=0.580 * | $M_1$=0.509<br>$M_2$=0.627 ** | $M_1$=0.367<br>$M_2$=0.449 ** | $M_1$=0.376<br>$M_2$=0.507<br>*** | $M_1$=0.389<br>$M_2$=0.462 * | $M_1$=0.454<br>$M_2$=0.653 *** | $M_1$=0.447<br>$M_2$=0.648 *** | $M_1$=0.420<br>$M_2$=0.515 * | $M_1$=0.454<br>$M_2$=0.548 * | $M_1$=0.355<br>$M_2$=0.431 * | $M_1$=0.354<br>$M_2$=0.386 | $M_1$=0.257<br>$M_2$=0.260 |
| BQ9, two groups: x<51 ($n_1$=333), x>=51 ($n_2$=340) | $M_1$=0.288<br>$M_2$=0.202 *** | $M_1$=0.333<br>$M_2$=0.237 *** | $M_1$=0.606<br>$M_2$=0.479 *** | $M_1$=0.659<br>$M_2$=0.557 *** | $M_1$=0.320<br>$M_2$=0.245 *** | $M_1$=0.394<br>$M_2$=0.365 | $M_1$=0.430<br>$M_2$=0.373 * | $M_1$=0.592<br>$M_2$=0.501 * | $M_1$=0.598<br>$M_2$=0.526 | $M_1$=0.655<br>$M_2$=0.557 | $M_1$=0.410<br>$M_2$=0.457 * | $M_1$=0.491<br>$M_2$=0.476 | $M_1$=0.477<br>$M_2$=0.422 * | $M_1$=0.686<br>$M_2$=0.549 *** | $M_1$=0.684<br>$M_2$=0.540 *** | $M_1$=0.508<br>$M_2$=0.487 | $M_1$=0.547<br>$M_2$=0.515 | $M_1$=0.422<br>$M_2$=0.414 | $M_1$=0.402<br>$M_2$=0.359 * | $M_1$=0.241<br>$M_2$=0.277 |
| BQ9, three groups: x<40 ($n_1$=225), 40<=x<60 ($n_2$=231), x>=60 ($n_3$=217) | $M_1$=0.301<br>$M_2$=0.229<br>$M_3$=0.203 *** | $M_1$=0.347<br>$M_2$=0.276<br>$M_3$=0.229 *** | $M_1$=0.618<br>$M_2$=0.553<br>$M_3$=0.450 *** | $M_1$=0.655<br>$M_2$=0.639<br>$M_3$=0.525 *** | $M_1$=0.321<br>$M_2$=0.287<br>$M_3$=0.236 *** | $M_1$=0.391<br>$M_2$=0.388<br>$M_3$=0.359 | $M_1$=0.434<br>$M_2$=0.423<br>$M_3$=0.344 ** | $M_1$=0.597<br>$M_2$=0.571<br>$M_3$=0.465 * | $M_1$=0.606<br>$M_2$=0.568<br>$M_3$=0.507 | $M_1$=0.659<br>$M_2$=0.639<br>$M_3$=0.515 * | $M_1$=0.374<br>$M_2$=0.471<br>$M_3$=0.456 ** | $M_1$=0.467<br>$M_2$=0.511<br>$M_3$=0.471 | $M_1$=0.480<br>$M_2$=0.437<br>$M_3$=0.429 | $M_1$=0.691<br>$M_2$=0.639<br>$M_3$=0.517 *** | $M_1$=0.692<br>$M_2$=0.628<br>$M_3$=0.510 *** | $M_1$=0.502<br>$M_2$=0.506<br>$M_3$=0.484 | $M_1$=0.545<br>$M_2$=0.540<br>$M_3$=0.506 | $M_1$=0.419<br>$M_2$=0.409<br>$M_3$=0.425 | $M_1$=0.411<br>$M_2$=0.378<br>$M_3$=0.351 * | $M_1$=0.240<br>$M_2$=0.231<br>$M_3$=0.309 * |

**Table A3. The median values of the "need for help" ratings of each expression statement (ES1-ES20) in respect to groupings based on the answer values of the background question (BQ), for two groups or three groups.**

$Mdn_1$, $Mdn_2$ and $Mdn_3$ show the median values for each group and the number of persons is denoted by $n_1$, $n_2$ and $n_3$ (n=673).

| *Grouping based on the answer value of the background question* | *ES1* | *ES2* | *ES3* | *ES4* | *ES5* | *ES6* | *ES7* | *ES8* | *ES9* | *ES10* |
|---|---|---|---|---|---|---|---|---|---|---|
| BQ1, two groups: x<7 ($n_1$=263), x>=7 ($n_2$=410) | $Mdn_1$=0.2<br>$Mdn_2$=0.2 | $Mdn_1$=0.2<br>$Mdn_2$=0.3 | $Mdn_1$=0.6<br>$Mdn_2$=0.7 | $Mdn_1$=0.7<br>$Mdn_2$=0.7 | $Mdn_1$=0.2<br>$Mdn_2$=0.3 | $Mdn_1$=0.4<br>$Mdn_2$=0.3 | $Mdn_1$=0.3<br>$Mdn_2$=0.5 | $Mdn_1$=0.6<br>$Mdn_2$=0.7 | $Mdn_1$=0.6<br>$Mdn_2$=0.8 | $Mdn_1$=0.8<br>$Mdn_2$=0.8 |
| BQ1, three groups: x<6 ($n_1$=218), 6<=x<8 ($n_2$=207), x>=8 ($n_3$=248) | $Mdn_1$=0.2<br>$Mdn_2$=0.2<br>$Mdn_3$=0.2 | $Mdn_1$=0.2<br>$Mdn_2$=0.3<br>$Mdn_3$=0.25 | $Mdn_1$=0.6<br>$Mdn_2$=0.7<br>$Mdn_3$=0.65 | $Mdn_1$=0.7<br>$Mdn_2$=0.7<br>$Mdn_3$=0.7 | $Mdn_1$=0.2<br>$Mdn_2$=0.3<br>$Mdn_3$=0.2 | $Mdn_1$=0.4<br>$Mdn_2$=0.4<br>$Mdn_3$=0.3 | $Mdn_1$=0.3<br>$Mdn_2$=0.5<br>$Mdn_3$=0.45 | $Mdn_1$=0.6<br>$Mdn_2$=0.7<br>$Mdn_3$=0.7 | $Mdn_1$=0.6<br>$Mdn_2$=0.8<br>$Mdn_3$=0.75 | $Mdn_1$=0.7<br>$Mdn_2$=0.8<br>$Mdn_3$=0.8 |
| BQ2, two groups: x<2 ($n_1$=219), x>=2 ($n_2$=454) | $Mdn_1$=0.2<br>$Mdn_2$=0.2 | $Mdn_1$=0.3<br>$Mdn_2$=0.2 | $Mdn_1$=0.7<br>$Mdn_2$=0.6 | $Mdn_1$=0.7<br>$Mdn_2$=0.7 | $Mdn_1$=0.3<br>$Mdn_2$=0.2 | $Mdn_1$=0.3<br>$Mdn_2$=0.4 | $Mdn_1$=0.4<br>$Mdn_2$=0.4 | $Mdn_1$=0.7<br>$Mdn_2$=0.7 | $Mdn_1$=0.8<br>$Mdn_2$=0.7 | $Mdn_1$=0.8<br>$Mdn_2$=0.8 |
| BQ4, two groups: x<2 ($n_1$=364), x>=2 ($n_2$=309) | $Mdn_1$=0.2<br>$Mdn_2$=0.2 | $Mdn_1$=0.3<br>$Mdn_2$=0.2 | $Mdn_1$=0.65<br>$Mdn_2$=0.6 | $Mdn_1$=0.7<br>$Mdn_2$=0.7 | $Mdn_1$=0.3<br>$Mdn_2$=0.2 | $Mdn_1$=0.3<br>$Mdn_2$=0.4 | $Mdn_1$=0.4<br>$Mdn_2$=0.4 | $Mdn_1$=0.7<br>$Mdn_2$=0.7 | $Mdn_1$=0.7<br>$Mdn_2$=0.7 | $Mdn_1$=0.8<br>$Mdn_2$=0.8 |
| BQ5, two groups: x<7 ($n_1$=274), x>=7 ($n_2$=399) | $Mdn_1$=0.2<br>$Mdn_2$=0.2 | $Mdn_1$=0.2<br>$Mdn_2$=0.3 | $Mdn_1$=0.6<br>$Mdn_2$=0.7 | $Mdn_1$=0.7<br>$Mdn_2$=0.7 | $Mdn_1$=0.2<br>$Mdn_2$=0.3 | $Mdn_1$=0.4<br>$Mdn_2$=0.3 | $Mdn_1$=0.3<br>$Mdn_2$=0.4 | $Mdn_1$=0.6<br>$Mdn_2$=0.7 | $Mdn_1$=0.7<br>$Mdn_2$=0.8 | $Mdn_1$=0.8<br>$Mdn_2$=0.8 |
| BQ5, three groups: x<6 ($n_1$=190), 6<=x<8 ($n_2$=271), x>=8 ($n_3$=212) | $Mdn_1$=0.2<br>$Mdn_2$=0.2<br>$Mdn_3$=0.2 | $Mdn_1$=0.2<br>$Mdn_2$=0.3<br>$Mdn_3$=0.2 | $Mdn_1$=0.55<br>$Mdn_2$=0.7<br>$Mdn_3$=0.7 | $Mdn_1$=0.7<br>$Mdn_2$=0.7<br>$Mdn_3$=0.7 | $Mdn_1$=0.2<br>$Mdn_2$=0.3<br>$Mdn_3$=0.2 | $Mdn_1$=0.4<br>$Mdn_2$=0.4<br>$Mdn_3$=0.3 | $Mdn_1$=0.3<br>$Mdn_2$=0.5<br>$Mdn_3$=0.4 | $Mdn_1$=0.6<br>$Mdn_2$=0.8<br>$Mdn_3$=0.7 | $Mdn_1$=0.55<br>$Mdn_2$=0.8<br>$Mdn_3$=0.8 | $Mdn_1$=0.7<br>$Mdn_2$=0.8<br>$Mdn_3$=0.8 |
| BQ6, two groups: x<7 ($n_1$=318), x>=7 ($n_2$=355) | $Mdn_1$=0.2<br>$Mdn_2$=0.2 | $Mdn_1$=0.2<br>$Mdn_2$=0.3 | $Mdn_1$=0.6<br>$Mdn_2$=0.7 | $Mdn_1$=0.7<br>$Mdn_2$=0.7 | $Mdn_1$=0.2<br>$Mdn_2$=0.3 | $Mdn_1$=0.4<br>$Mdn_2$=0.3 | $Mdn_1$=0.4<br>$Mdn_2$=0.4 | $Mdn_1$=0.7<br>$Mdn_2$=0.7 | $Mdn_1$=0.7<br>$Mdn_2$=0.7 | $Mdn_1$=0.8<br>$Mdn_2$=0.8 |
| BQ6, three groups: x<6 ($n_1$=240), 6<=x<8 ($n_2$=229), x>=8 ($n_3$=204) | $Mdn_1$=0.2<br>$Mdn_2$=0.2<br>$Mdn_3$=0.2 | $Mdn_1$=0.2<br>$Mdn_2$=0.3<br>$Mdn_3$=0.2 | $Mdn_1$=0.6<br>$Mdn_2$=0.6<br>$Mdn_3$=0.65 | $Mdn_1$=0.7<br>$Mdn_2$=0.7<br>$Mdn_3$=0.7 | $Mdn_1$=0.2<br>$Mdn_2$=0.3<br>$Mdn_3$=0.3 | $Mdn_1$=0.4<br>$Mdn_2$=0.4<br>$Mdn_3$=0.3 | $Mdn_1$=0.4<br>$Mdn_2$=0.4<br>$Mdn_3$=0.5 | $Mdn_1$=0.6<br>$Mdn_2$=0.7<br>$Mdn_3$=0.7 | $Mdn_1$=0.6<br>$Mdn_2$=0.8<br>$Mdn_3$=0.7 | $Mdn_1$=0.8<br>$Mdn_2$=0.8<br>$Mdn_3$=0.8 |
| BQ7, two groups: x<7 ($n_1$=201), x>=7 ($n_2$=472) | $Mdn_1$=0.2<br>$Mdn_2$=0.2 | $Mdn_1$=0.3<br>$Mdn_2$=0.2 | $Mdn_1$=0.6<br>$Mdn_2$=0.6 | $Mdn_1$=0.7<br>$Mdn_2$=0.7 | $Mdn_1$=0.3<br>$Mdn_2$=0.3 | $Mdn_1$=0.4<br>$Mdn_2$=0.3 | $Mdn_1$=0.4<br>$Mdn_2$=0.4 | $Mdn_1$=0.7<br>$Mdn_2$=0.7 | $Mdn_1$=0.7<br>$Mdn_2$=0.7 | $Mdn_1$=0.8<br>$Mdn_2$=0.8 |
| BQ7, three groups: x<6 ($n_1$=143), 6<=x<8 ($n_2$=214), x>=8 ($n_3$=316) | $Mdn_1$=0.2<br>$Mdn_2$=0.2<br>$Mdn_3$=0.2 | $Mdn_1$=0.2<br>$Mdn_2$=0.3<br>$Mdn_3$=0.2 | $Mdn_1$=0.6<br>$Mdn_2$=0.65<br>$Mdn_3$=0.65 | $Mdn_1$=0.7<br>$Mdn_2$=0.7<br>$Mdn_3$=0.7 | $Mdn_1$=0.2<br>$Mdn_2$=0.3<br>$Mdn_3$=0.25 | $Mdn_1$=0.4<br>$Mdn_2$=0.4<br>$Mdn_3$=0.3 | $Mdn_1$=0.4<br>$Mdn_2$=0.5<br>$Mdn_3$=0.4 | $Mdn_1$=0.6<br>$Mdn_2$=0.7<br>$Mdn_3$=0.7 | $Mdn_1$=0.6<br>$Mdn_2$=0.8<br>$Mdn_3$=0.7 | $Mdn_1$=0.8<br>$Mdn_2$=0.8<br>$Mdn_3$=0.8 |
| BQ8, two groups: x<2 ($n_1$=123), x>=2 ($n_2$=550) | $Mdn_1$=0.2<br>$Mdn_2$=0.2 | $Mdn_1$=0.2<br>$Mdn_2$=0.25 | $Mdn_1$=0.5<br>$Mdn_2$=0.7 | $Mdn_1$=0.6<br>$Mdn_2$=0.7 | $Mdn_1$=0.2<br>$Mdn_2$=0.3 | $Mdn_1$=0.4<br>$Mdn_2$=0.4 | $Mdn_1$=0.3<br>$Mdn_2$=0.4 | $Mdn_1$=0.6<br>$Mdn_2$=0.7 | $Mdn_1$=0.5<br>$Mdn_2$=0.7 | $Mdn_1$=0.6<br>$Mdn_2$=0.8 |
| BQ9, two groups: x<51 ($n_1$=333), x>=51 ($n_2$=340) | $Mdn_1$=0.2<br>Mdn2=0.2 | $Mdn_1$=0.3<br>$Mdn_2$=0.2 | $Mdn_1$=0.7<br>$Mdn_2$=0.5 | $Mdn_1$=0.7<br>$Mdn_2$=0.7 | $Mdn_1$=0.3<br>$Mdn_2$=0.2 | $Mdn_1$=0.4<br>$Mdn_2$=0.3 | $Mdn_1$=0.5<br>$Mdn_2$=0.4 | $Mdn_1$=0.7<br>$Mdn_2$=0.6 | $Mdn_1$=0.7<br>$Mdn_2$=0.7 | $Mdn_1$=0.8<br>$Mdn_2$=0.8 |
| BQ9, three groups: x<40 ($n_1$=225), 40<=x<60 ($n_2$=231), x>=60 ($n_3$=217) | $Mdn_1$=0.3<br>$Mdn_2$=0.2<br>$Mdn_3$=0.1 | $Mdn_1$=0.3<br>$Mdn_2$=0.2<br>$Mdn_3$=0.2 | $Mdn_1$=0.7<br>$Mdn_2$=0.7<br>$Mdn_3$=0.5 | $Mdn_1$=0.7<br>$Mdn_2$=0.8<br>$Mdn_3$=0.6 | $Mdn_1$=0.3<br>$Mdn_2$=0.2<br>$Mdn_3$=0.2 | $Mdn_1$=0.4<br>$Mdn_2$=0.4<br>$Mdn_3$=0.3 | $Mdn_1$=0.5<br>$Mdn_2$=0.4<br>$Mdn_3$=0.3 | $Mdn_1$=0.7<br>$Mdn_2$=0.7<br>$Mdn_3$=0.6 | $Mdn_1$=0.7<br>$Mdn_2$=0.7<br>$Mdn_3$=0.7 | $Mdn_1$=0.8<br>$Mdn_2$=0.8<br>$Mdn_3$=0.7 |

| *Grouping based on the answer value of the background question* | *ES11* | *ES12* | *ES13* | *ES14* | *ES15* | *ES16* | *ES17* | *ES18* | *ES19* | *ES20* |
|---|---|---|---|---|---|---|---|---|---|---|
| BQ1, two groups: x<7 ($n_1$=263), x>=7 ($n_2$=410) | $Mdn_1$=0.5<br>$Mdn_2$=0.4 | $Mdn_1$=0.5<br>$Mdn_2$=0.6 | $Mdn_1$=0.5<br>$Mdn_2$=0.5 | $Mdn_1$=0.8<br>$Mdn_2$=0.8 | $Mdn_1$=0.8<br>$Mdn_2$=0.8 | $Mdn_1$=0.5<br>$Mdn_2$=0.6 | $Mdn_1$=0.5<br>$Mdn_2$=0.7 | $Mdn_1$=0.3<br>$Mdn_2$=0.5 | $Mdn_1$=0.4<br>$Mdn_2$=0.4 | $Mdn_1$=0.2<br>$Mdn_2$=0.1 |
| BQ1, three groups: x<6 ($n_1$=218), 6<=x<8 ($n_2$=207), x>=8 ($n_3$=248) | $Mdn_1$=0.5<br>$Mdn_2$=0.4<br>$Mdn_3$=0.3 | $Mdn_1$=0.5<br>$Mdn_2$=0.6<br>$Mdn_3$=0.5 | $Mdn_1$=0.4<br>$Mdn_2$=0.5<br>$Mdn_3$=0.5 | $Mdn_1$=0.8<br>$Mdn_2$=0.8<br>$Mdn_3$=0.8 | $Mdn_1$=0.8<br>$Mdn_2$=0.8<br>$Mdn_3$=0.8 | $Mdn_1$=0.5<br>$Mdn_2$=0.6<br>$Mdn_3$=0.55 | $Mdn_1$=0.5<br>$Mdn_2$=0.7<br>$Mdn_3$=0.6 | $Mdn_1$=0.3<br>$Mdn_2$=0.5<br>$Mdn_3$=0.5 | $Mdn_1$=0.4<br>$Mdn_2$=0.4<br>$Mdn_3$=0.3 | $Mdn_1$=0.1<br>$Mdn_2$=0.2<br>$Mdn_3$=0.1 |
| BQ2, two groups: x<2 ($n_1$=219), x>=2 ($n_2$=454) | $Mdn_1$=0.3<br>$Mdn_2$=0.5 | $Mdn_1$=0.5<br>$Mdn_2$=0.5 | $Mdn_1$=0.5<br>$Mdn_2$=0.5 | $Mdn_1$=0.8<br>$Mdn_2$=0.8 | $Mdn_1$=0.8<br>$Mdn_2$=0.8 | $Mdn_1$=0.6<br>$Mdn_2$=0.5 | $Mdn_1$=0.7<br>$Mdn_2$=0.6 | $Mdn_1$=0.5<br>$Mdn_2$=0.4 | $Mdn_1$=0.4<br>$Mdn_2$=0.4 | $Mdn_1$=0.1<br>$Mdn_2$=0.2 |
| BQ4, two groups: x<2 ($n_1$=364), x>=2 ($n_2$=309) | $Mdn_1$=0.35<br>$Mdn_2$=0.5 | $Mdn_1$=0.5<br>$Mdn_2$=0.6 | $Mdn_1$=0.5<br>$Mdn_2$=0.4 | $Mdn_1$=0.8<br>$Mdn_2$=0.8 | $Mdn_1$=0.8<br>$Mdn_2$=0.8 | $Mdn_1$=0.6<br>$Mdn_2$=0.5 | $Mdn_1$=0.6<br>$Mdn_2$=0.6 | $Mdn_1$=0.45<br>$Mdn_2$=0.4 | $Mdn_1$=0.4<br>$Mdn_2$=0.4 | $Mdn_1$=0.1<br>$Mdn_2$=0.2 |
| BQ5, two groups: x<7 ($n_1$=274), x>=7 ($n_2$=399) | $Mdn_1$=0.5<br>$Mdn_2$=0.4 | $Mdn_1$=0.5<br>$Mdn_2$=0.5 | $Mdn_1$=0.45<br>$Mdn_2$=0.5 | $Mdn_1$=0.8<br>$Mdn_2$=0.8 | $Mdn_1$=0.75<br>$Mdn_2$=0.8 | $Mdn_1$=0.5<br>$Mdn_2$=0.6 | $Mdn_1$=0.6<br>$Mdn_2$=0.7 | $Mdn_1$=0.4<br>$Mdn_2$=0.5 | $Mdn_1$=0.4<br>$Mdn_2$=0.3 | $Mdn_1$=0.2<br>$Mdn_2$=0.1 |
| BQ5, three groups: x<6 ($n_1$=190), 6<=x<8 ($n_2$=271), x>=8 ($n_3$=212) | $Mdn_1$=0.5<br>$Mdn_2$=0.4<br>$Mdn_3$=0.3 | $Mdn_1$=0.5<br>$Mdn_2$=0.5<br>$Mdn_3$=0.5 | $Mdn_1$=0.4<br>$Mdn_2$=0.5<br>$Mdn_3$=0.45 | $Mdn_1$=0.8<br>$Mdn_2$=0.8<br>$Mdn_3$=0.8 | $Mdn_1$=0.7<br>$Mdn_2$=0.8<br>$Mdn_3$=0.9 | $Mdn_1$=0.5<br>$Mdn_2$=0.6<br>$Mdn_3$=0.6 | $Mdn_1$=0.5<br>$Mdn_2$=0.6<br>$Mdn_3$=0.7 | $Mdn_1$=0.3<br>$Mdn_2$=0.5<br>$Mdn_3$=0.5 | $Mdn_1$=0.4<br>$Mdn_2$=0.4<br>$Mdn_3$=0.3 | $Mdn_1$=0.2<br>$Mdn_2$=0.2<br>$Mdn_3$=0.1 |
| BQ6, two groups: x<7 ($n_1$=318), x>=7 ($n_2$=355) | $Mdn_1$=0.5<br>$Mdn_2$=0.3 | $Mdn_1$=0.55<br>$Mdn_2$=0.5 | $Mdn_1$=0.5<br>$Mdn_2$=0.5 | $Mdn_1$=0.8<br>$Mdn_2$=0.8 | $Mdn_1$=0.8<br>$Mdn_2$=0.8 | $Mdn_1$=0.5<br>$Mdn_2$=0.6 | $Mdn_1$=0.6<br>$Mdn_2$=0.6 | $Mdn_1$=0.4<br>$Mdn_2$=0.5 | $Mdn_1$=0.4<br>$Mdn_2$=0.3 | $Mdn_1$=0.2<br>$Mdn_2$=0.2 |
| BQ6, three groups: x<6 ($n_1$=240), 6<=x<8 ($n_2$=229), x>=8 ($n_3$=204) | $Mdn_1$=0.5<br>$Mdn_2$=0.4<br>$Mdn_3$=0.3 | $Mdn_1$=0.5<br>$Mdn_2$=0.5<br>$Mdn_3$=0.5 | $Mdn_1$=0.4<br>$Mdn_2$=0.5<br>$Mdn_3$=0.5 | $Mdn_1$=0.8<br>$Mdn_2$=0.8<br>$Mdn_3$=0.8 | $Mdn_1$=0.8<br>$Mdn_2$=0.8<br>$Mdn_3$=0.8 | $Mdn_1$=0.5<br>$Mdn_2$=0.5<br>$Mdn_3$=0.6 | $Mdn_1$=0.6<br>$Mdn_2$=0.7<br>$Mdn_3$=0.6 | $Mdn_1$=0.4<br>$Mdn_2$=0.4<br>$Mdn_3$=0.5 | $Mdn_1$=0.4<br>$Mdn_2$=0.4<br>$Mdn_3$=0.3 | $Mdn_1$=0.2<br>$Mdn_2$=0.1<br>$Mdn_3$=0.1 |
| BQ7, two groups: x<7 ($n_1$=201), x>=7 ($n_2$=472) | $Mdn_1$=0.5<br>$Mdn_2$=0.4 | $Mdn_1$=0.6<br>$Mdn_2$=0.5 | $Mdn_1$=0.5<br>$Mdn_2$=0.5 | $Mdn_1$=0.8<br>$Mdn_2$=0.8 | $Mdn_1$=0.8<br>$Mdn_2$=0.8 | $Mdn_1$=0.6<br>$Mdn_2$=0.5 | $Mdn_1$=0.6<br>$Mdn_2$=0.6 | $Mdn_1$=0.4<br>$Mdn_2$=0.4 | $Mdn_1$=0.4<br>$Mdn_2$=0.35 | $Mdn_1$=0.2<br>$Mdn_2$=0.1 |
| BQ7, three groups: x<6 ($n_1$=143), 6<=x<8 ($n_2$=214), x>=8 ($n_3$=316) | $Mdn_1$=0.5<br>$Mdn_2$=0.5<br>$Mdn_3$=0.3 | $Mdn_1$=0.6<br>$Mdn_2$=0.6<br>$Mdn_3$=0.5 | $Mdn_1$=0.5<br>$Mdn_2$=0.5<br>$Mdn_3$=0.45 | $Mdn_1$=0.8<br>$Mdn_2$=0.8<br>$Mdn_3$=0.8 | $Mdn_1$=0.8<br>$Mdn_2$=0.8<br>$Mdn_3$=0.8 | $Mdn_1$=0.6<br>$Mdn_2$=0.6<br>$Mdn_3$=0.5 | $Mdn_1$=0.6<br>$Mdn_2$=0.6<br>$Mdn_3$=0.6 | $Mdn_1$=0.4<br>$Mdn_2$=0.5<br>$Mdn_3$=0.4 | $Mdn_1$=0.4<br>$Mdn_2$=0.4<br>$Mdn_3$=0.4 | $Mdn_1$=0.2<br>$Mdn_2$=0.2<br>$Mdn_3$=0.1 |
| BQ8, two groups: x<2 ($n_1$=123), x>=2 ($n_2$=550) | $Mdn_1$=0.3<br>$Mdn_2$=0.4 | $Mdn_1$=0.3<br>$Mdn_2$=0.6 | $Mdn_1$=0.4<br>$Mdn_2$=0.5 | $Mdn_1$=0.5<br>$Mdn_2$=0.8 | $Mdn_1$=0.5<br>$Mdn_2$=0.8 | $Mdn_1$=0.5<br>$Mdn_2$=0.6 | $Mdn_1$=0.5<br>$Mdn_2$=0.65 | $Mdn_1$=0.3<br>$Mdn_2$=0.5 | $Mdn_1$=0.3<br>$Mdn_2$=0.4 | $Mdn_1$=0.1<br>$Mdn_2$=0.2 |
| BQ9, two groups: x<51 ($n_1$=333), x>=51 ($n_2$=340) | $Mdn_1$=0.4<br>$Mdn_2$=0.5 | $Mdn_1$=0.5<br>$Mdn_2$=0.5 | $Mdn_1$=0.5<br>$Mdn_2$=0.4 | $Mdn_1$=0.8<br>$Mdn_2$=0.7 | $Mdn_1$=0.8<br>$Mdn_2$=0.7 | $Mdn_1$=0.6<br>$Mdn_2$=0.5 | $Mdn_1$=0.6<br>$Mdn_2$=0.6 | $Mdn_1$=0.4<br>$Mdn_2$=0.4 | $Mdn_1$=0.4<br>$Mdn_2$=0.3 | $Mdn_1$=0.2<br>$Mdn_2$=0.15 |
| BQ9, three groups: x<40 ($n_1$=225), 40<=x<60 ($n_2$=231), x>=60 ($n_3$=217) | $Mdn_1$=0.3<br>$Mdn_2$=0.5<br>$Mdn_3$=0.5 | $Mdn_1$=0.5<br>$Mdn_2$=0.6<br>$Mdn_3$=0.5 | $Mdn_1$=0.5<br>$Mdn_2$=0.4<br>$Mdn_3$=0.4 | $Mdn_1$=0.8<br>$Mdn_2$=0.8<br>$Mdn_3$=0.5 | $Mdn_1$=0.8<br>$Mdn_2$=0.8<br>$Mdn_3$=0.5 | $Mdn_1$=0.5<br>$Mdn_2$=0.6<br>$Mdn_3$=0.6 | $Mdn_1$=0.6<br>$Mdn_2$=0.6<br>$Mdn_3$=0.6 | $Mdn_1$=0.4<br>$Mdn_2$=0.4<br>$Mdn_3$=0.5 | $Mdn_1$=0.4<br>$Mdn_2$=0.4<br>$Mdn_3$=0.3 | $Mdn_1$=0.2<br>$Mdn_2$=0.1<br>$Mdn_3$=0.2 |

**Table A4. The standard deviation values of the "need for help" ratings of each expression statement (ES1-ES20) in respect to groupings based on the answer values of the background question (BQ), for two groups or three groups.**

$SD_1$, $SD_2$ and $SD_3$ show the standard deviation values for each group and the number of persons is denoted by $n_1$, $n_2$ and $n_3$ (n=673).

| *Grouping based on the answer value of the background question* | *ES1* | *ES2* | *ES3* | *ES4* | *ES5* | *ES6* | *ES7* | *ES8* | *ES9* | *ES10* |
|---|---|---|---|---|---|---|---|---|---|---|
| BQ1, two groups: x<7 ($n_1$=263), x>=7 ($n_2$=410) | $SD_1$=0.22<br>$SD_2$=0.21 | $SD_1$=0.23<br>$SD_2$=0.22 | $SD_1$=0.32<br>$SD_2$=0.34 | $SD_1$=0.29<br>$SD_2$=0.34 | $SD_1$=0.24<br>$SD_2$=0.24 | $SD_1$=0.26<br>$SD_2$=0.24 | $SD_1$=0.31<br>$SD_2$=0.3 | $SD_1$=0.37<br>$SD_2$=0.36 | $SD_1$=0.39<br>$SD_2$=0.38 | $SD_1$=0.42<br>$SD_2$=0.41 |
| BQ1, three groups: x<6 ($n_1$= 218), 6<=x<8 ($n_2$=207), x>=8 ($n_3$=248) | $SD_1$=0.22<br>$SD_2$=0.21<br>$SD_3$=0.21 | $SD_1$=0.23<br>$SD_2$=0.23<br>$SD_3$=0.22 | $SD_1$=0.32<br>$SD_2$=0.33<br>$SD_3$=0.34 | $SD_1$=0.29<br>$SD_2$=0.32<br>$SD_3$=0.36 | $SD_1$=0.25<br>$SD_2$=0.24<br>$SD_3$=0.23 | $SD_1$=0.26<br>$SD_2$=0.24<br>$SD_3$=0.23 | $SD_1$=0.31<br>$SD_2$=0.3<br>$SD_3$=0.3 | $SD_1$=0.37<br>$SD_2$=0.36<br>$SD_3$=0.36 | $SD_1$=0.39<br>$SD_2$=0.39<br>$SD_3$=0.38 | $SD_1$=0.42<br>$SD_2$=0.42<br>$SD_3$=0.4 |
| BQ2, two groups: x<2 ($n_1$=219), x>=2 ($n_2$=454) | $SD_1$=0.21<br>$SD_2$=0.22 | $SD_1$=0.22<br>$SD_2$=0.23 | $SD_1$=0.31<br>$SD_2$=0.34 | $SD_1$=0.32<br>$SD_2$=0.32 | $SD_1$=0.21<br>$SD_2$=0.25 | $SD_1$=0.22<br>$SD_2$=0.26 | $SD_1$=0.28<br>$SD_2$=0.32 | $SD_1$=0.34<br>$SD_2$=0.38 | $SD_1$=0.36<br>$SD_2$=0.4 | $SD_1$=0.38<br>$SD_2$=0.43 |
| BQ4, two groups: x<2 ($n_1$=364), x>=2 ($n_2$=309) | $SD_1$=0.21<br>$SD_2$=0.22 | $SD_1$=0.22<br>$SD_2$=0.23 | $SD_1$=0.33<br>$SD_2$=0.33 | $SD_1$=0.32<br>$SD_2$=0.33 | $SD_1$=0.23<br>$SD_2$=0.25 | $SD_1$=0.23<br>$SD_2$=0.26 | $SD_1$=0.3<br>$SD_2$=0.32 | $SD_1$=0.35<br>$SD_2$=0.38 | $SD_1$=0.38<br>$SD_2$=0.4 | $SD_1$=0.4<br>$SD_2$=0.43 |
| BQ5, two groups: x<7 ($n_1$=274), x>=7 ($n_2$=399) | $SD_1$=0.23<br>$SD_2$=0.21 | $SD_1$=0.23<br>$SD_2$=0.22 | $SD_1$=0.32<br>$SD_2$=0.34 | $SD_1$=0.3<br>$SD_2$=0.34 | $SD_1$=0.24<br>$SD_2$=0.24 | $SD_1$=0.26<br>$SD_2$=0.24 | $SD_1$=0.31<br>$SD_2$=0.3 | $SD_1$=0.37<br>$SD_2$=0.36 | $SD_1$=0.39<br>$SD_2$=0.38 | $SD_1$=0.42<br>$SD_2$=0.41 |
| BQ5, three groups: x<6 ($n_1$=190), 6<=x<8 ($n_2$=271), x>=8 ($n_3$=212) | $SD_1$=0.24<br>$SD_2$=0.21<br>$SD_3$=0.21 | $SD_1$=0.24<br>$SD_2$=0.22<br>$SD_3$=0.22 | $SD_1$=0.33<br>$SD_2$=0.32<br>$SD_3$=0.34 | $SD_1$=0.3<br>$SD_2$=0.32<br>$SD_3$=0.34 | $SD_1$=0.25<br>$SD_2$=0.24<br>$SD_3$=0.23 | $SD_1$=0.26<br>$SD_2$=0.24<br>$SD_3$=0.23 | $SD_1$=0.32<br>$SD_2$=0.3<br>$SD_3$=0.3 | $SD_1$=0.38<br>$SD_2$=0.37<br>$SD_3$=0.35 | $SD_1$=0.4<br>$SD_2$=0.39<br>$SD_3$=0.37 | $SD_1$=0.42<br>$SD_2$=0.42<br>$SD_3$=0.39 |
| BQ6, two groups: x<7 ($n_1$=318), x>=7 ($n_2$=355) | $SD_1$=0.22<br>$SD_2$=0.21 | $SD_1$=0.23<br>$SD_2$=0.22 | $SD_1$=0.32<br>$SD_2$=0.34 | $SD_1$=0.3<br>$SD_2$=0.34 | $SD_1$=0.25<br>$SD_2$=0.23 | $SD_1$=0.25<br>$SD_2$=0.24 | $SD_1$=0.31<br>$SD_2$=0.3 | $SD_1$=0.37<br>$SD_2$=0.36 | $SD_1$=0.39<br>$SD_2$=0.39 | $SD_1$=0.42<br>$SD_2$=0.41 |
| BQ6, three groups: x<6 ($n_1$=240), 6<=x<8 ($n_2$=229), x>=8 ($n_3$=204) | $SD_1$=0.23<br>$SD_2$=0.21<br>$SD_3$=0.21 | $SD_1$=0.23<br>$SD_2$=0.23<br>$SD_3$=0.22 | $SD_1$=0.32<br>$SD_2$=0.33<br>$SD_3$=0.34 | $SD_1$=0.29<br>$SD_2$=0.32<br>$SD_3$=0.36 | $SD_1$=0.24<br>$SD_2$=0.24<br>$SD_3$=0.23 | $SD_1$=0.26<br>$SD_2$=0.23<br>$SD_3$=0.24 | $SD_1$=0.31<br>$SD_2$=0.3<br>$SD_3$=0.31 | $SD_1$=0.37<br>$SD_2$=0.36<br>$SD_3$=0.36 | $SD_1$=0.39<br>$SD_2$=0.38<br>$SD_3$=0.39 | $SD_1$=0.42<br>$SD_2$=0.41<br>$SD_3$=0.41 |
| BQ7, two groups: x<7 ($n_1$=201), x>=7 ($n_2$=472) | $SD_1$=0.22<br>$SD_2$=0.21 | $SD_1$=0.23<br>$SD_2$=0.23 | $SD_1$=0.31<br>$SD_2$=0.34 | $SD_1$=0.28<br>$SD_2$=0.34 | $SD_1$=0.24<br>$SD_2$=0.24 | $SD_1$=0.26<br>$SD_2$=0.24 | $SD_1$=0.31<br>$SD_2$=0.31 | $SD_1$=0.36<br>$SD_2$=0.37 | $SD_1$=0.38<br>$SD_2$=0.39 | $SD_1$=0.41<br>$SD_2$=0.42 |
| BQ7, three groups: x<6 ($n_1$=143), 6<=x<8 ($n_2$=214), x>=8 ($n_3$=316) | $SD_1$=0.22<br>$SD_2$=0.22<br>$SD_3$=0.21 | $SD_1$=0.23<br>$SD_2$=0.22<br>$SD_3$=0.23 | $SD_1$=0.32<br>$SD_2$=0.31<br>$SD_3$=0.35 | $SD_1$=0.27<br>$SD_2$=0.32<br>$SD_3$=0.35 | $SD_1$=0.24<br>$SD_2$=0.25<br>$SD_3$=0.23 | $SD_1$=0.26<br>$SD_2$=0.25<br>$SD_3$=0.23 | $SD_1$=0.31<br>$SD_2$=0.31<br>$SD_3$=0.31 | $SD_1$=0.36<br>$SD_2$=0.37<br>$SD_3$=0.36 | $SD_1$=0.38<br>$SD_2$=0.39<br>$SD_3$=0.39 | $SD_1$=0.42<br>$SD_2$=0.41<br>$SD_3$=0.42 |
| BQ8, two groups: x<2 ($n_1$=123), x>=2 ($n_2$=550) | $SD_1$=0.23<br>$SD_2$=0.21 | $SD_1$=0.25<br>$SD_2$=0.22 | $SD_1$=0.33<br>$SD_2$=0.33 | $SD_1$=0.32<br>$SD_2$=0.32 | $SD_1$=0.26<br>$SD_2$=0.23 | $SD_1$=0.26<br>$SD_2$=0.24 | $SD_1$=0.31<br>$SD_2$=0.3 | $SD_1$=0.38<br>$SD_2$=0.36 | $SD_1$=0.4<br>$SD_2$=0.38 | $SD_1$=0.42<br>$SD_2$=0.41 |
| BQ9, two groups: x<51 ($n_1$=333), x>=51 ($n_2$=340) | $SD_1$=0.22<br>$SD_2$=0.2 | $SD_1$=0.22<br>$SD_2$=0.22 | $SD_1$=0.31<br>$SD_2$=0.35 | $SD_1$=0.29<br>$SD_2$=0.34 | $SD_1$=0.23<br>$SD_2$=0.24 | $SD_1$=0.24<br>$SD_2$=0.26 | $SD_1$=0.28<br>$SD_2$=0.33 | $SD_1$=0.33<br>$SD_2$=0.39 | $SD_1$=0.36<br>$SD_2$=0.41 | $SD_1$=0.38<br>$SD_2$=0.44 |
| BQ9, three groups: x<40 ($n_1$=225), 40<=x<60 ($n_2$=231), x>=60 ($n_3$=217) | $SD_1$=0.22<br>$SD_2$=0.21<br>$SD_3$=0.21 | $SD_1$=0.22<br>$SD_2$=0.22<br>$SD_3$=0.22 | $SD_1$=0.29<br>$SD_2$=0.34<br>$SD_3$=0.34 | $SD_1$=0.28<br>$SD_2$=0.33<br>$SD_3$=0.35 | $SD_1$=0.22<br>$SD_2$=0.24<br>$SD_3$=0.25 | $SD_1$=0.24<br>$SD_2$=0.25<br>$SD_3$=0.25 | $SD_1$=0.28<br>$SD_2$=0.31<br>$SD_3$=0.33 | $SD_1$=0.33<br>$SD_2$=0.36<br>$SD_3$=0.39 | $SD_1$=0.35<br>$SD_2$=0.38<br>$SD_3$=0.43 | $SD_1$=0.37<br>$SD_2$=0.4<br>$SD_3$=0.45 |

| *Grouping based on the answer value of the background question* | *ES11* | *ES12* | *ES13* | *ES14* | *ES15* | *ES16* | *ES17* | *ES18* | *ES19* | *ES20* |
|---|---|---|---|---|---|---|---|---|---|---|
| BQ1, two groups: x<7 ($n_1$=263), x>=7 ($n_2$=410) | $SD_1$=0.32<br>$SD_2$=0.3 | $SD_1$=0.35<br>$SD_2$=0.34 | $SD_1$=0.32<br>$SD_2$=0.32 | $SD_1$=0.37<br>$SD_2$=0.38 | $SD_1$=0.39<br>$SD_2$=0.4 | $SD_1$=0.37<br>$SD_2$=0.36 | $SD_1$=0.38<br>$SD_2$=0.37 | $SD_1$=0.32<br>$SD_2$=0.31 | $SD_1$=0.27<br>$SD_2$=0.27 | $SD_1$=0.27<br>$SD_2$=0.3 |
| BQ1, three groups: x<6 ($n_1$= 218), 6<=x<8 ($n_2$=207), x>=8 ($n_3$=248) | $SD_1$=0.32<br>$SD_2$=0.3<br>$SD_3$=0.3 | $SD_1$=0.35<br>$SD_2$=0.33<br>$SD_3$=0.34 | $SD_1$=0.33<br>$SD_2$=0.31<br>$SD_3$=0.32 | $SD_1$=0.37<br>$SD_2$=0.37<br>$SD_3$=0.39 | $SD_1$=0.39<br>$SD_2$=0.39<br>$SD_3$=0.4 | $SD_1$=0.38<br>$SD_2$=0.36<br>$SD_3$=0.36 | $SD_1$=0.39<br>$SD_2$=0.38<br>$SD_3$=0.37 | $SD_1$=0.32<br>$SD_2$=0.32<br>$SD_3$=0.31 | $SD_1$=0.28<br>$SD_2$=0.26<br>$SD_3$=0.27 | $SD_1$=0.27<br>$SD_2$=0.29<br>$SD_3$=0.3 |
| BQ2, two groups: x<2 ($n_1$=219), x>=2 ($n_2$=454) | $SD_1$=0.29<br>$SD_2$=0.31 | $SD_1$=0.33<br>$SD_2$=0.35 | $SD_1$=0.31<br>$SD_2$=0.33 | $SD_1$=0.38<br>$SD_2$=0.38 | $SD_1$=0.38<br>$SD_2$=0.4 | $SD_1$=0.34<br>$SD_2$=0.38 | $SD_1$=0.35<br>$SD_2$=0.39 | $SD_1$=0.3<br>$SD_2$=0.33 | $SD_1$=0.26<br>$SD_2$=0.28 | $SD_1$=0.27<br>$SD_2$=0.29 |
| BQ4, two groups: x<2 ($n_1$=364), x>=2 ($n_2$=309) | $SD_1$=0.31<br>$SD_2$=0.31 | $SD_1$=0.34<br>$SD_2$=0.35 | $SD_1$=0.32<br>$SD_2$=0.32 | $SD_1$=0.38<br>$SD_2$=0.38 | $SD_1$=0.39<br>$SD_2$=0.4 | $SD_1$=0.36<br>$SD_2$=0.38 | $SD_1$=0.37<br>$SD_2$=0.39 | $SD_1$=0.31<br>$SD_2$=0.33 | $SD_1$=0.26<br>$SD_2$=0.28 | $SD_1$=0.29<br>$SD_2$=0.29 |
| BQ5, two groups: x<7 ($n_1$=274), x>=7 ($n_2$=399) | $SD_1$=0.32<br>$SD_2$=0.3 | $SD_1$=0.35<br>$SD_2$=0.34 | $SD_1$=0.33<br>$SD_2$=0.32 | $SD_1$=0.37<br>$SD_2$=0.38 | $SD_1$=0.39<br>$SD_2$=0.4 | $SD_1$=0.37<br>$SD_2$=0.36 | $SD_1$=0.38<br>$SD_2$=0.37 | $SD_1$=0.32<br>$SD_2$=0.31 | $SD_1$=0.27<br>$SD_2$=0.27 | $SD_1$=0.26<br>$SD_2$=0.3 |
| BQ5, three groups: x<6 ($n_1$=190), 6<=x<8 ($n_2$=271), x>=8 ($n_3$=212) | $SD_1$=0.32<br>$SD_2$=0.31<br>$SD_3$=0.3 | $SD_1$=0.36<br>$SD_2$=0.33<br>$SD_3$=0.34 | $SD_1$=0.33<br>$SD_2$=0.32<br>$SD_3$=0.31 | $SD_1$=0.38<br>$SD_2$=0.37<br>$SD_3$=0.39 | $SD_1$=0.4<br>$SD_2$=0.39<br>$SD_3$=0.4 | $SD_1$=0.38<br>$SD_2$=0.36<br>$SD_3$=0.36 | $SD_1$=0.39<br>$SD_2$=0.37<br>$SD_3$=0.37 | $SD_1$=0.33<br>$SD_2$=0.31<br>$SD_3$=0.31 | $SD_1$=0.28<br>$SD_2$=0.27<br>$SD_3$=0.26 | $SD_1$=0.26<br>$SD_2$=0.29<br>$SD_3$=0.3 |
| BQ6, two groups: x<7 ($n_1$=318), x>=7 ($n_2$=355) | $SD_1$=0.32<br>$SD_2$=0.3 | $SD_1$=0.35<br>$SD_2$=0.33 | $SD_1$=0.33<br>$SD_2$=0.32 | $SD_1$=0.37<br>$SD_2$=0.39 | $SD_1$=0.39<br>$SD_2$=0.4 | $SD_1$=0.37<br>$SD_2$=0.37 | $SD_1$=0.38<br>$SD_2$=0.38 | $SD_1$=0.32<br>$SD_2$=0.32 | $SD_1$=0.27<br>$SD_2$=0.27 | $SD_1$=0.27<br>$SD_2$=0.3 |
| BQ6, three groups: x<6 ($n_1$=240), 6<=x<8 ($n_2$=229), x>=8 ($n_3$=204) | $SD_1$=0.31<br>$SD_2$=0.31<br>$SD_3$=0.3 | $SD_1$=0.36<br>$SD_2$=0.33<br>$SD_3$=0.34 | $SD_1$=0.33<br>$SD_2$=0.32<br>$SD_3$=0.32 | $SD_1$=0.37<br>$SD_2$=0.38<br>$SD_3$=0.39 | $SD_1$=0.39<br>$SD_2$=0.39<br>$SD_3$=0.4 | $SD_1$=0.38<br>$SD_2$=0.36<br>$SD_3$=0.37 | $SD_1$=0.39<br>$SD_2$=0.36<br>$SD_3$=0.38 | $SD_1$=0.33<br>$SD_2$=0.31<br>$SD_3$=0.32 | $SD_1$=0.27<br>$SD_2$=0.26<br>$SD_3$=0.28 | $SD_1$=0.27<br>$SD_2$=0.29<br>$SD_3$=0.31 |
| BQ7, two groups: x<7 ($n_1$=201), x>=7 ($n_2$=472) | $SD_1$=0.31<br>$SD_2$=0.31 | $SD_1$=0.35<br>$SD_2$=0.34 | $SD_1$=0.32<br>$SD_2$=0.32 | $SD_1$=0.35<br>$SD_2$=0.39 | $SD_1$=0.38<br>$SD_2$=0.4 | $SD_1$=0.37<br>$SD_2$=0.37 | $SD_1$=0.38<br>$SD_2$=0.38 | $SD_1$=0.32<br>$SD_2$=0.32 | $SD_1$=0.27<br>$SD_2$=0.27 | $SD_1$=0.27<br>$SD_2$=0.3 |
| BQ7, three groups: x<6 ($n_1$=143), 6<=x<8 ($n_2$=214), x>=8 ($n_3$=316) | $SD_1$=0.32<br>$SD_2$=0.31<br>$SD_3$=0.3 | $SD_1$=0.35<br>$SD_2$=0.33<br>$SD_3$=0.34 | $SD_1$=0.32<br>$SD_2$=0.32<br>$SD_3$=0.32 | $SD_1$=0.35<br>$SD_2$=0.37<br>$SD_3$=0.4 | $SD_1$=0.38<br>$SD_2$=0.38<br>$SD_3$=0.41 | $SD_1$=0.37<br>$SD_2$=0.38<br>$SD_3$=0.36 | $SD_1$=0.38<br>$SD_2$=0.38<br>$SD_3$=0.37 | $SD_1$=0.31<br>$SD_2$=0.33<br>$SD_3$=0.32 | $SD_1$=0.26<br>$SD_2$=0.27<br>$SD_3$=0.27 | $SD_1$=0.26<br>$SD_2$=0.29<br>$SD_3$=0.3 |
| BQ8, two groups: x<2 ($n_1$=123), x>=2 ($n_2$=550) | $SD_1$=0.32<br>$SD_2$=0.3 | $SD_1$=0.34<br>$SD_2$=0.34 | $SD_1$=0.32<br>$SD_2$=0.32 | $SD_1$=0.38<br>$SD_2$=0.37 | $SD_1$=0.39<br>$SD_2$=0.39 | $SD_1$=0.38<br>$SD_2$=0.36 | $SD_1$=0.4<br>$SD_2$=0.37 | $SD_1$=0.33<br>$SD_2$=0.31 | $SD_1$=0.28<br>$SD_2$=0.27 | $SD_1$=0.29<br>$SD_2$=0.29 |
| BQ9, two groups: x<51 ($n_1$=333), x>=51 ($n_2$=340) | $SD_1$=0.3<br>$SD_2$=0.31 | $SD_1$=0.32<br>$SD_2$=0.36 | $SD_1$=0.31<br>$SD_2$=0.33 | $SD_1$=0.35<br>$SD_2$=0.4 | $SD_1$=0.36<br>$SD_2$=0.41 | $SD_1$=0.33<br>$SD_2$=0.4 | $SD_1$=0.34<br>$SD_2$=0.41 | $SD_1$=0.29<br>$SD_2$=0.35 | $SD_1$=0.25<br>$SD_2$=0.29 | $SD_1$=0.27<br>$SD_2$=0.3 |
| BQ9, three groups: x<40 ($n_1$=225), 40<=x<60 ($n_2$=231), x>=60 ($n_3$=217) | $SD_1$=0.29<br>$SD_2$=0.32<br>$SD_3$=0.31 | $SD_1$=0.31<br>$SD_2$=0.35<br>$SD_3$=0.37 | $SD_1$=0.3<br>$SD_2$=0.32<br>$SD_3$=0.34 | $SD_1$=0.33<br>$SD_2$=0.38<br>$SD_3$=0.4 | $SD_1$=0.35<br>$SD_2$=0.4<br>$SD_3$=0.42 | $SD_1$=0.32<br>$SD_2$=0.36<br>$SD_3$=0.42 | $SD_1$=0.33<br>$SD_2$=0.37<br>$SD_3$=0.43 | $SD_1$=0.28<br>$SD_2$=0.31<br>$SD_3$=0.37 | $SD_1$=0.24<br>$SD_2$=0.26<br>$SD_3$=0.3 | $SD_1$=0.26<br>$SD_2$=0.28<br>$SD_3$=0.32 |

**Table A5. Results of Wilcoxon rank-sum test (i.e., Mann–Whitney U test) between two groups and Kruskal-Wallis test between three groups to identify statistically significant rating differences for expression statements ES1-ES20 in respect to groupings based on the answer values of each background question (BQ).**
This table shows the p-values concerning the rating differences between groups when the statistical significance levels were defined as p<0.05, p<0.01 and p<0.001, denoted by symbols *, ** and *** respectively (n=673).

| *Grouping based on the answer value of the background question* | *ES1* | *ES2* | *ES3* | *ES4* | *ES5* | *ES6* | *ES7* | *ES8* | *ES9* | *ES10* | *ES11* | *ES12* | *ES13* | *ES14* | *ES15* | *ES16* | *ES17* | *ES18* | *ES19* | *ES20* |
|---|---|---|---|---|---|---|---|---|---|---|---|---|---|---|---|---|---|---|---|---|
| BQ1, two groups: x<7 ($n_1$=263), x>=7 ($n_2$=410) | 0.9197 | 0.6643 | 0.1491 | 0.4769 | 0.1094 | 0.001105 ** | 0.03835 * | 0.007304 ** | 0.006771 ** | 0.004866 ** | 0.05642 | 0.3291 | 0.6438 | 0.3987 | 0.1316 | 0.04026 * | 0.01429 * | 0.0358 * | 0.3167 | 0.5732 |
| BQ1, three groups: x<6 ($n_1$= 218), 6<=x<8 ($n_2$=207), x>=8 ($n_3$=248) | 0.331 | 0.1774 | 0.4204 | 0.815 | 0.04487 * | 0.002876 ** | 0.2161 | 0.04889 * | 0.03546 * | 0.05269 | 0.01077 * | 0.9052 | 0.7382 | 0.4413 | 0.398 | 0.4829 | 0.2057 | 0.168 | 0.1383 | 0.3959 |
| BQ2, two groups: x<2 ($n_1$=219), x>=2 ($n_2$=454) | 0.2048 | 0.1175 | 0.04755 * | 0.4151 | 0.2517 | 0.003857 ** | 0.269 | 0.2825 | 0.05099 | 0.0657 | 0.001373 ** | 0.6126 | 0.4365 | 0.04 * | 0.01885 * | 0.122 | 0.1097 | 0.09556 | 0.5401 | 0.1131 |
| BQ4, two groups: x<2 ($n_1$=364), x>=2 ($n_2$=309) | 0.3973 | 0.3349 | 0.5797 | 0.5483 | 0.3886 | 0.006364 ** | 0.2846 | 0.1325 | 0.2279 | 0.09887 | 0.01654 * | 0.8479 | 0.1744 | 0.361 | 0.1451 | 0.2824 | 0.1716 | 0.3776 | 0.5248 | 0.3334 |
| BQ5, two groups: x<7 ($n_1$=274), x>=7 ($n_2$=399) | 0.7408 | 0.9772 | 0.1047 | 0.292 | 0.4035 | 0.002381 ** | 0.1343 | 0.05747 | 0.00434 ** | 0.003607 ** | 0.01683 * | 0.8051 | 0.6267 | 0.1929 | 0.1205 | 0.0271 * | 0.03034 * | 0.08549 | 0.06951 | 0.5023 |
| BQ5, three groups: x<6 ($n_1$=190), 6<=x<8 ($n_2$=271), x>=8 ($n_3$=212) | 0.1532 | 0.2886 | 0.07562 | 0.3356 | 0.06069 | 0.00926 ** | 0.1414 | 0.01166 * | 0.0006896 *** | 0.0005286 *** | 0.168 | 0.6902 | 0.2367 | 0.2365 | 0.09009 | 0.03014 * | 0.03286 * | 0.09426 | 0.3523 | 0.01388 * |
| BQ6, two groups: x<7 ($n_1$=318), x>=7 ($n_2$=355) | 0.9496 | 0.4908 | 0.5658 | 0.7009 | 0.5836 | 0.005581 ** | 0.3036 | 0.1441 | 0.1953 | 0.289 | 0.0005744 *** | 0.5883 | 0.9916 | 0.9809 | 0.9041 | 0.4146 | 0.4478 | 0.6883 | 0.05302 | 0.9842 |
| BQ6, three groups: x<6 ($n_1$=240), 6<=x<8 ($n_2$=229), x>=8 ($n_3$=204) | 0.6204 | 0.9745 | 0.7312 | 0.8557 | 0.44 | 0.01988 * | 0.3726 | 0.1359 | 0.1467 | 0.1851 | 0.009845 ** | 0.9204 | 0.6588 | 0.6927 | 0.602 | 0.6682 | 0.7609 | 0.625 | 0.262 | 0.6617 |
| BQ7, two groups: x<7 ($n_1$=201), x>=7 ($n_2$=472) | 0.1691 | 0.2554 | 0.8877 | 0.07533 | 0.6983 | 0.000459 *** | 0.7582 | 0.6412 | 0.3008 | 0.3424 | 0.00778 ** | 0.1283 | 0.3437 | 0.1559 | 0.5785 | 0.7799 | 0.9402 | 0.8928 | 0.004751 ** | 0.5095 |
| BQ7, three groups: x<6 ($n_1$=143), 6<=x<8 ($n_2$=214), x>=8 ($n_3$=316) | 0.8834 | 0.7608 | 0.5168 | 0.3388 | 0.8597 | 0.004156 ** | 1 | 0.6272 | 0.4356 | 0.6223 | 0.003545 ** | 0.349 | 0.6863 | 0.687 | 0.948 | 0.9213 | 0.9675 | 0.7298 | 0.1105 | 0.684 |
| BQ8, two groups: x<2 ($n_1$=123), x>=2 ($n_2$=550) | 0.5048 | 0.2059 | 0.003137 ** | 0.0001024 *** | 0.8198 | 0.5797 | 0.09938 | 0.02251 * | 0.01417 * | 0.00504 ** | 0.005765 ** | 0.0001656 *** | 0.02226 * | 1.453e-08 *** | 1.59e-08 *** | 0.01594 * | 0.03187 * | 0.02422 * | 0.2095 | 0.7081 |
| BQ9, two groups: x<51 ($n_1$=333), x>=51 ($n_2$=340) | 6.116e-09 *** | 3.613e-09 *** | 4.392e-06 *** | 0.000614 *** | 1.18e-06 *** | 0.07347 | 0.01332 * | 0.04664 * | 0.1505 | 0.0585 | 0.04851 * | 0.742 | 0.02965 * | 9.661e-05 *** | 9.863e-05 *** | 0.9912 | 0.978 | 0.8095 | 0.01925 * | 0.3238 |
| BQ9, three groups: x<40 ($n_1$=225), 40<=x<60 ($n_2$=231), x>=60 ($n_3$=217) | 1.808e-07 *** | 2.82e-08 *** | 6.349e-06 *** | 0.0004377 *** | 3.191e-05 *** | 0.2526 | 0.004295 ** | 0.01155 * | 0.2856 | 0.0245 * | 0.001738 ** | 0.2289 | 0.222 | 0.0002202 *** | 0.000565 *** | 0.8223 | 0.8839 | 0.8217 | 0.03513 * | 0.02525 * |

**Table A6. Results of one-way analysis of variance (ANOVA) between two groups and between three groups to identify statistically significant rating differences for expression statements ES1-ES20 in respect to groupings based on the answer values of each background question (BQ).**

This table shows the p-values concerning the rating differences between groups when the statistical significance levels were defined as p<0.05, p<0.01 and p<0.001, denoted by symbols *, ** and *** respectively (n=673).

| *Grouping based on the answer value of the background question* | *ES1* | *ES2* | *ES3* | *ES4* | *ES5* | *ES6* | *ES7* | *ES8* | *ES9* | *ES10* | *ES11* | *ES12* | *ES13* | *ES14* | *ES15* | *ES16* | *ES17* | *ES18* | *ES19* | *ES20* |
|---|---|---|---|---|---|---|---|---|---|---|---|---|---|---|---|---|---|---|---|---|
| BQ1, two groups: x<7 ($n_1$=263), x>=7 ($n_2$=410) | 0.833 | 0.814 | 0.253 | 0.761 | 0.184 | 0.000523 *** | 0.031 * | 0.00591 ** | 0.00703 ** | 0.00759 ** | 0.0427 * | 0.257 | 0.579 | 0.512 | 0.158 | 0.0348 * | 0.0101 * | 0.0322 * | 0.231 | 0.87 |
| BQ1, three groups: x<6 ($n_1$= 218), 6<=x<8 ($n_2$=207), x>=8 ($n_3$=248) | 0.646 | 0.23 | 0.561 | 0.705 | 0.0945 | 0.00115 ** | 0.186 | 0.0325 * | 0.0368 * | 0.0561 | 0.00935 ** | 0.845 | 0.711 | 0.8 | 0.469 | 0.375 | 0.115 | 0.16 | 0.118 | 0.474 |
| BQ2, two groups: x<2 ($n_1$=219), x>=2 ($n_2$=454) | 0.483 | 0.211 | 0.0237 * | 0.405 | 0.653 | 0.00207 ** | 0.247 | 0.0905 | 0.0135 * | 0.0174 * | 0.00104 ** | 0.727 | 0.414 | 0.0724 | 0.0194 * | 0.0733 | 0.0461 * | 0.0801 | 0.452 | 0.0966 |
| BQ4, two groups: x<2 ($n_1$=364), x>=2 ($n_2$=309) | 0.548 | 0.502 | 0.543 | 0.535 | 0.748 | 0.0037 ** | 0.299 | 0.0653 | 0.185 | 0.0661 | 0.0176 * | 0.849 | 0.177 | 0.315 | 0.093 | 0.249 | 0.135 | 0.341 | 0.456 | 0.453 |
| BQ5, two groups: x<7 ($n_1$=274), x>=7 ($n_2$=399) | 0.448 | 0.811 | 0.181 | 0.765 | 0.467 | 0.00174 ** | 0.108 | 0.037 * | 0.0108 * | 0.00916 ** | 0.0123 * | 0.707 | 0.592 | 0.279 | 0.137 | 0.0265 * | 0.0232 * | 0.0675 | 0.0545 | 0.775 |
| BQ5, three groups: x<6 ($n_1$=190), 6<=x<8 ($n_2$=271), x>=8 ($n_3$=212) | 0.432 | 0.526 | 0.0788 | 0.702 | 0.133 | 0.00783 ** | 0.105 | 0.00669 ** | 0.00069 *** | 0.00109 ** | 0.125 | 0.624 | 0.217 | 0.397 | 0.0944 | 0.0211 * | 0.0183 * | 0.0726 | 0.303 | 0.169 |
| BQ6, two groups: x<7 ($n_1$=318), x>=7 ($n_2$=355) | 0.823 | 0.626 | 0.713 | 0.301 | 0.814 | 0.00346 ** | 0.254 | 0.112 | 0.189 | 0.424 | 0.000383 *** | 0.703 | 0.953 | 0.821 | 0.877 | 0.443 | 0.428 | 0.684 | 0.0371 * | 0.426 |
| BQ6, three groups: x<6 ($n_1$=240), 6<=x<8 ($n_2$=229), x>=8 ($n_3$=204) | 0.572 | 0.976 | 0.819 | 0.455 | 0.588 | 0.0119 * | 0.316 | 0.108 | 0.159 | 0.288 | 0.00789 ** | 0.891 | 0.653 | 0.936 | 0.642 | 0.658 | 0.647 | 0.627 | 0.193 | 0.786 |
| BQ7, two groups: x<7 ($n_1$=201), x>=7 ($n_2$=472) | 0.179 | 0.239 | 0.654 | 0.00977 ** | 0.729 | 0.000157 *** | 0.841 | 0.667 | 0.382 | 0.511 | 0.00576 ** | 0.137 | 0.363 | 0.092 | 0.65 | 0.79 | 0.908 | 0.921 | 0.00306 ** | 0.996 |
| BQ7, three groups: x<6 ($n_1$=143), 6<=x<8 ($n_2$=214), x>=8 ($n_3$=316) | 0.902 | 0.896 | 0.673 | 0.0935 | 0.783 | 0.00171 ** | 0.993 | 0.641 | 0.505 | 0.757 | 0.00286 ** | 0.377 | 0.709 | 0.444 | 0.777 | 0.959 | 0.984 | 0.745 | 0.0905 | 0.788 |
| BQ8, two groups: x<2 ($n_1$=123), x>=2 ($n_2$=550) | 0.86 | 0.4 | 0.0026 ** | 0.000519 *** | 0.936 | 0.593 | 0.0924 | 0.0159 * | 0.00782 ** | 0.00413 ** | 0.00849 ** | 0.00012 *** | 0.0213 * | 1e-07 *** | 2.82e-07 *** | 0.00918 ** | 0.0123 * | 0.0162 * | 0.242 | 0.919 |
| BQ9, two groups: x<51 ($n_1$=333), x>=51 ($n_2$=340) | 1.49e-07 *** | 2.65e-08 *** | 5.44e-07 *** | 4.5e-05 *** | 3.52e-05 *** | 0.124 | 0.0149 * | 0.00118 ** | 0.0157 * | 0.00207 ** | 0.0479 * | 0.568 | 0.0268 * | 2.66e-06 *** | 2.14e-06 *** | 0.451 | 0.275 | 0.741 | 0.0375 * | 0.11 |
| BQ9, three groups: x<40 ($n_1$=225), 40<=x<60 ($n_2$=231), x>=60 ($n_3$=217) | 3.51e-06 *** | 2.3e-07 *** | 4.82e-07 *** | 2.2e-05 *** | 0.000736 *** | 0.313 | 0.00341 ** | 0.000308 *** | 0.0263 * | 0.000401 *** | 0.00145 ** | 0.309 | 0.19 | 4.26e-06 *** | 5.23e-06 *** | 0.802 | 0.511 | 0.868 | 0.0672 | 0.00704 ** |

**Table A7. Guidance texts of the online questionnaire.**

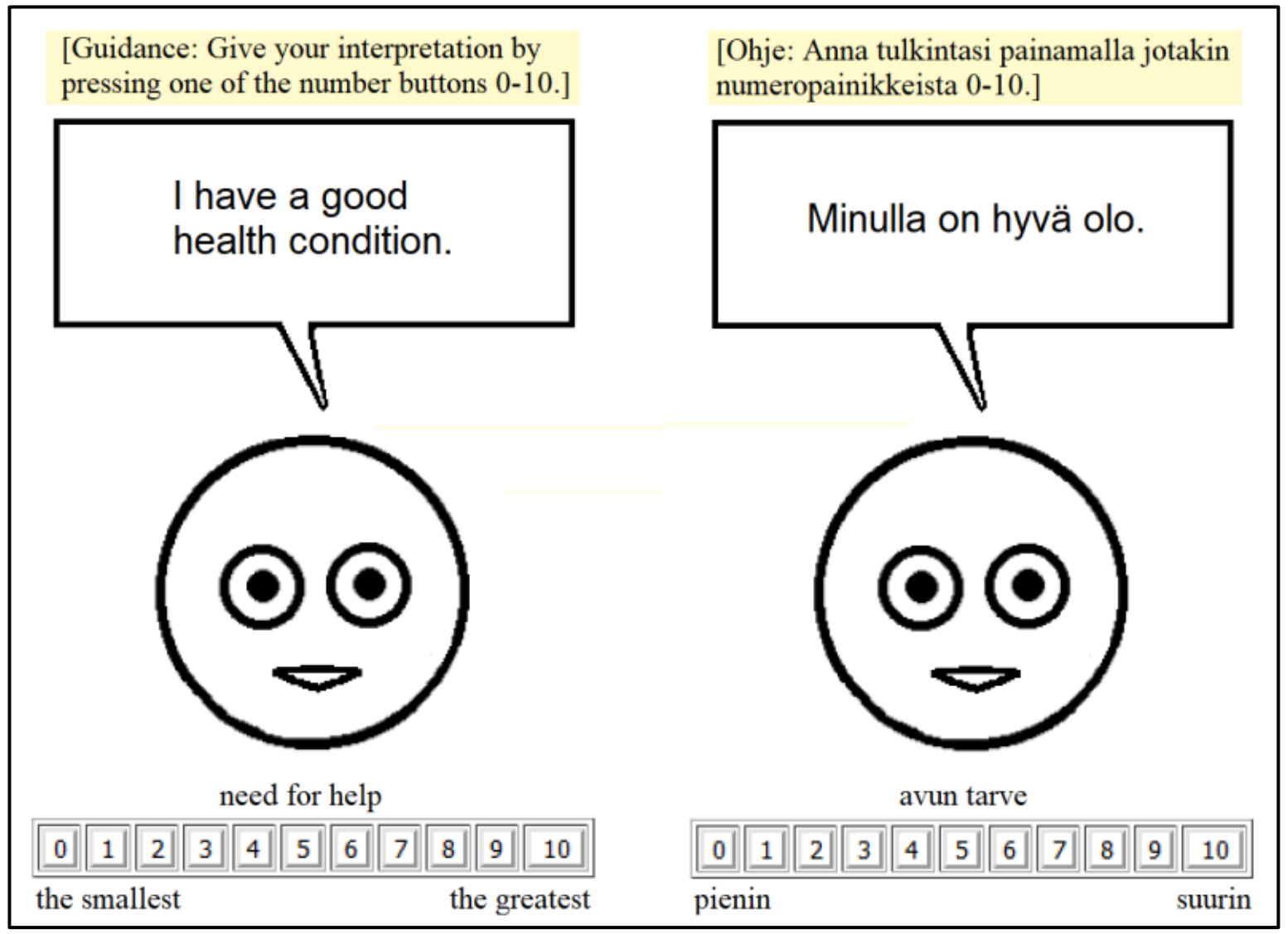

Figure 1. Gathering the “need for help” rating for an expression statement on an 11-point Likert scale with an online questionnaire.

Before the online questionnaire started to collect actual answers, the person was provided with the following guidance texts about how he/she should perform the interpretation tasks: "We ask you to evaluate different expressions, for example the expression 'I am happy'. Interpret how much each expression tells about the need for help. Give your interpretation about the expression on a numeric scale 0-10. 0 indicates the smallest possible need for help and 10 indicates the greatest possible need for help."

In Finnish:
"Pyydämme sinua arvioimaan erilaisia ilmaisuja, esimerkiksi ilmaisua 'olen iloinen'. Tulkitse, kuinka paljon kukin ilmaisu kertoo avun tarpeesta. Anna tulkintasi ilmaisusta numeroasteikolla 0-10. 0 tarkoittaa mahdollisimman pientä avun tarvetta ja 10 tarkoittaa mahdollisimman suurta avun tarvetta."

Then a small training phase allowed the person to get accustomed to give the "need for help" ratings by rating three expression statements: “I have a good health condition.”, “I have a bad health condition.” and “I have an ordinary health condition.” The answers that the person gave during the training phase were not included in our major analysis data set.

In Finnish:
"Minulla on hyvä olo." / "Minulla on huono olo." / "Minulla on tavallinen olo."

After the training phase, the person was provided with the following guidance texts to still further clarify how he/she should perform the interpretation tasks: “Do not interpret how much the expression tells about just your own situation. Instead, interpret what kind of impression this expression induces in you. Thus give your interpretation about the expression's meaning in respect to the mentioned property.” After showing those guidance texts, the person was allowed to start giving actual answers, i.e. to perform the actual interpretation tasks.

In Finnish:
"Älä tulkitse, kuinka paljon ilmaisu kertoo juuri sinun omasta tilanteestasi. Sen sijaan tulkitse, minkälaisen vaikutelman tämä ilmaisu herättää sinussa. Siis anna tulkintasi ilmaisun merkityksestä suhteessa mainittuun ominaisuuteen."

**Table A8. Expression statements (ES) concerning the coronavirus epidemic that were rated by the person in respect to the impression about the "need for help".**

| *Compact notation* | *Expression statement* | *Expression statement in Finnish* | *Range of values for the person's answer (indicating the "need for help" rating)* |
|---|---|---|---|
| ES1 | "I have a flu." | "Minulla on nuhaa." | 0-10 |
| ES2 | "I have a cough." | "Minulla on yskää." | 0-10 |
| ES3 | "I have a shortness of breath." | "Minulla on hengenahdistusta." | 0-10 |
| ES4 | "My health condition is weakening." | "Yleistilani heikkenee." | 0-10 |
| ES5 | "I have a sore throat." | "Minulla on kurkkukipua. " | 0-10 |
| ES6 | "I have muscular ache." | "Minulla on lihassärkyä." | 0-10 |
| ES7 | "I have a fever." | "Minulla on kuumetta." | 0-10 |
| ES8 | "A sudden fever rises for me with 38 degrees of Celsius or more." | "Minulle nousee äkillinen kuume, joka on 38 astetta Celsiusta tai enemmän." | 0-10 |
| ES9 | "I suspect that I have now become infected by the coronavirus." | "Epäilen, että olen nyt sairastunut koronavirukseen." | 0-10 |
| ES10 | "I have now become infected by the coronavirus." | "Olen nyt sairastunut koronavirukseen." | 0-10 |
| ES11 | "I am quarantined from meeting other people ordinarily so that the spreading of an infectious disease could be prevented." | "Olen eristettynä ihmisten tavanomaiselta tapaamiselta, jotta tartuntatautin leviäminen estyisi." | 0-10 |
| ES12 | "I must be inside a house without getting out." | "Joudun olemaan talon sisällä ilman ulospääsyä." | 0-10 |
| ES13 | "I must be without a human companion." | "Joudun olemaan ilman ihmisseuraa." | 0-10 |
| ES14 | "I do not cope in everyday life independently without getting help from other persons." | "En pärjää arkielämässä itsenäisesti ilman avun saamista muilta henkilöiltä." | 0-10 |
| ES15 | "I do not cope at home independently without getting help from persons who originate outside of my home." | "En pärjää kotona itsenäisesti ilman avun saamista kotini ulkopuolisilta henkilöiltä." | 0-10 |
| ES16 | "I have an infectious disease." | "Minulla on tartuntatauti." | 0-10 |
| ES17 | "I have an infectious disease that has been verified by a doctor." | "Minulla on tartuntatauti, jonka lääkäri on varmistanut." | 0-10 |
| ES18 | "I suspect that I have an infectious disease." | "Epäilen, että minulla on tartuntatauti." | 0-10 |
| ES19 | "I have a bad health condition." | "Minulla on huono olo." | 0-10 |
| ES20 | "I have an ordinary health condition." | "Minulla on tavallinen olo." | 0-10 |

**Table A9. Background questions (BQ) presented to the person.**

| *Compact notation* | *Question about the person's background information* | *Question about the person's background information in Finnish* | *Range of values for the person's answer* | *Range of values for the person's answer in Finnish* |
|---|---|---|---|---|
| BQ1: an estimated health condition | "What kind of health condition you have currently according to your opinion?" (de Bruin et al., 1996; Koskinen et al., 2012) | "Minkälainen terveydentilasi on mielestäsi nykyisin?" (de Bruin et al., 1996; Koskinen et al., 2012) | A 9-point Likert scale supplied with the following partial labeling: "9 Good", "8 –", "7 Rather good", "6 –", "5 Medium", "4 –", "3 Rather bad", "2 –", "1 Bad". | A 9-point Likert scale supplied with the following partial labeling: "9 Hyvä", "8 –", "7 Melko hyvä", "6 –", "5 Keskitasoinen", "4 –", "3 Melko huono", "2 –", "1 Huono". |
| BQ2: a health problem reduces ability | "Do you have a permanent or long-lasting disease or such deficit, ailment or disability that reduces your ability to work or to perform your daily living activities? Here the question refers to all long-lasting diseases identified by a doctor, and also to such ailments not identified by a doctor which have lasted at least three months but which affect your ability to perform your daily living activities." (Koskinen et al., 2012) | "Onko sinulla jokin pysyvä tai pitkäaikainen sairaus tai jokin sellainen vika, vaiva tai vamma, joka vähentää työ- tai toimintakykyäsi? Tässä tarkoitetaan kaikkia lääkärin toteamia pitkäaikaisia sairauksia sekä myös vähintään kolme kuukautta kestäneitä vaivoja, joita lääkäri ei ole todennut, mutta jotka vaikuttavat toimintakykyysi." (Koskinen et al., 2012) | No or yes | Ei or kyllä. |
| BQ3: one or more diseases identified by a doctor | "Has there been a situation that a doctor has identified in you one or several of the following diseases?" (Koskinen et al., 2012) | "Onko lääkäri joskus todennut sinulla jonkin/joitakin seuraavista sairauksista?" (Koskinen et al., 2012) | The person answers by selecting one or more options from a list of diseases (Koskinen et al., 2012), see Appendix A. For some options there is a question "other, what?" and an adjacent text input box so that the person can write some additional information concerning that option. | Henkilö vastaa valitsemalla yhden tai useampia vaihtoehtoja sairauksia sisältävästä luettelosta (Koskinen et al., 2012), katso Appendix A. Joidenkin vaihtoehtojen kohdalla on kysymys "muu, mikä?" ja sen vieressä tekstin syöttämislaatikko, johon henkilö voi kirjoittaa täydentävää tietoa koskien kyseistä vaihtoehtoa. |
| BQ4: a continuous or repeated need for a doctor's care | "Do you need continuously or repeatedly care given by a doctor for a long-lasting disease, deficit or disability that you have just mentioned?" (Koskinen et al., 2012) | "Tarvitsetko jatkuvasti tai toistuvasti lääkärinhoitoa jonkin äsken mainitsemasi pitkäaikaisen sairauden, vian tai vamman takia?" (Koskinen et al., 2012) | No or yes | En or kyllä. |
| BQ5: the quality of life | "How would you rate your quality of life? Give your estimate based on the latest two weeks." (Nosikov & Gudex 2003; Aalto et al., 2013) | "Minkälaiseksi arvioit elämänlaatusi? Anna arviosi viimeisimpien kahden viikon ajalta." (Nosikov & Gudex 2003; Aalto et al., 2013) | A 9-point Likert scale supplied with the following partial labeling: "9 Very good", "8 –", "7 Good", "6 –", "5 Neither good nor bad", "4 –", "3 Bad", "2 –", "1 Very bad". | A 9-point Likert scale supplied with the following partial labeling: "9 Erittäin hyväksi", "8 –", "7 Hyväksi", "6 –", "5 Ei hyväksi eikä huonoksi", "4 –", "3 Huonoksi", "2 –", "1 Erittäin huonoksi". |
| BQ6: the satisfaction about health | "How satisfied are you with your health? Give your estimate based on the latest two weeks." (Nosikov & Gudex 2003; Aalto et al., 2013) | "Kuinka tyytyväinen olet terveyteesi? Anna arviosi viimeisimpien kahden viikon ajalta." (Nosikov & Gudex 2003; Aalto et al., 2013) | A 9-point Likert scale supplied with the following partial labeling: "9 Very satisfied", "8 –", "7 Satisfied", "6 –", "5 Neither satisfied nor unsatisfied", "4 –", "3 Unsatisfied", "2 –", "1 Very unsatisfied". | A 9-point Likert scale supplied with the following partial labeling: "9 Erittäin tyytyväinen", "8 –", "7 Tyytyväinen", "6 –", "5 Ei tyytyväinen eikä tyytymätön", "4 –", "3 Tyytymätön", "2 –", "1 Erittäin tyytymätön". |
| BQ7: the satisfaction about ability | "How satisfied are you with your ability to perform your daily living activities? Give your estimate based on the latest two weeks." (Nosikov & Gudex 2003; Aalto et al., 2013) | "Kuinka tyytyväinen olet kykyysi selviytyä päivittäisistä toiminnoistasi? Anna arviosi viimeisimpien kahden viikon ajalta." (Nosikov & Gudex 2003; Aalto et al., 2013) | A 9-point Likert scale supplied with the following partial labeling: "9 Very satisfied", "8 –", "7 Satisfied", "6 –", "5 Neither satisfied nor unsatisfied", "4 –", "3 Unsatisfied", "2 –", "1 Very unsatisfied". | A 9-point Likert scale supplied with the following partial labeling: "9 Erittäin tyytyväinen", "8 –", "7 Tyytyväinen", "6 –", "5 Ei tyytyväinen eikä tyytymätön", "4 –", "3 Tyytymätön", "2 –", "1 Erittäin tyytymätön". |
| BQ8: the sex | "Tell what is your sex. The answer alternatives are similar as in the earlier health surveys of Finnish Institute for Health and Welfare in Finland (THL) to maintain comparability with the earlier results." (Koskinen et al., 2012) | "Kerro sukupuolesi. Vastausvaihtoehdot ovat samankaltaiset kuin aiemmissa THL:n terveystutkimuksissa, jotta säilyisi vertailtavuus aiempiin tuloksiin." (Koskinen et al., 2012) | Man or woman | Mies or nainen. |
| BQ9: the age | "Tell what is your age." (Koskinen et al., 2012) | "Kerro ikäsi." (Koskinen et al., 2012) | 16 years, 17 years, ..., 99 years, 100 years or more | 16 vuotta, 17 vuotta, ..., 99 vuotta, 100 vuotta tai enemmän. |

**Table A10. Background question BQ3.**

| *Compact notation* | *Question about the person's background information* | *Question about the person's background information in Finnish* | *Range of values for the person's answer* | *Range of values for the person's answer in Finnish* |
|---|---|---|---|---|
| BQ3: one or more diseases identified by a doctor | "Has there been a situation that a doctor has identified in you one or several of the following diseases?" (Koskinen et al. 2012) | "Onko lääkäri joskus todennut sinulla jonkin/joitakin seuraavista sairauksista?" (Koskinen et al. 2012) | The person answers by selecting one or more options from a list of diseases (Koskinen et al., 2012), see Appendix A. For some options there is a question "other, what?" and an adjacent text input box so that the person can write some additional information concerning that option. | Henkilö vastaa valitsemalla yhden tai useampia vaihtoehtoja sairauksia sisältävästä luettelosta (Koskinen et al., 2012), katso Appendix A. Joidenkin vaihtoehtojen kohdalla on kysymys "muu, mikä?" ja sen vieressä tekstin syöttämislaatikko, johon henkilö voi kirjoittaa täydentävää tietoa koskien kyseistä vaihtoehtoa. |

The person was asked to indicate if a doctor had identified one or more diseases in him/her and to describe them (BQ3) (in a form adapted from Koskinen et al., 2012).

| *Questions about one or more diseases identified by a doctor (in a form adapted from Koskinen et al., 2012)* | *Questions about one or more diseases identified by a doctor in Finnish (in a form adapted from Koskinen et al., 2012)* |
|---|---|
| Has there been a situation that a doctor has identified in you one or several of the following diseases?<br><br>Select all diseases that belong to your response in the following way:<br>Click the square that is on the left side of the name of the disease and then a check mark emerges in it. You can remove the selection with a new click.<br>If needed, answer to the questions "other, what?" in the following way: Click the box that is besides the question and write your answer into it.<br><br>Finally, press the button "I save my response and continue forward". | Onko lääkäri joskus todennut sinulla jonkin/joitakin seuraavista sairauksista?<br><br>Valitse kaikki vastaukseesi kuuluvat sairaudet seuraavasti:<br>Napauta sairauden nimen vasemmalla puolella olevaa ruutua, jolloin siihen ilmestyy valintamerkki. Voit poistaa valinnan uudella napautuksella.<br>Tarvittaessa vastaa kysymyksiin "muu, mikä?" seuraavasti: Napauta kysymyksen vieressä näkyvää laatikkoa ja kirjoita siihen vastauksesi.<br><br>Lopuksi paina painiketta "Tallennan vastaukseni ja jatkan eteenpäin". |
| LUNG DISEASES<br>[_] 1. asthma<br>[_] 2. chronic obstructive pulmonary disease (COPD)<br>[_] 3. inflammation of the bronchi (chronic bronchitis, lung catarrh) | KEUHKOSAIRAUDET<br>[_] 1. astma<br>[_] 2. keuhkoputkien ahtauma (COPD)<br>[_] 3. krooninen keuhkoputkentulehdus (krooninen bronkiitti, keuhkokatarri) |
| HEART AND CIRCULATORY DISEASES<br>[_] 4. heart attack, i.e., myocardial infarction<br>[_] 5. coronary artery disease (atherosclerosis, angina pectoris)<br>[_] 6. congestive heart failure<br>[_] 7. high blood pressure, hypertension<br>[_] 8. stroke (intracranial hemorrhage, cerebral infarction) | SYDÄN- JA VERISUONISAIRAUDET<br>[_] 4. sydänveritulppa eli sydäninfarkti<br>[_] 5. sepelvaltimotauti (sepelvaltimoiden ahtauma, angina pectoris)<br>[_] 6. sydämen vajaatoiminta<br>[_] 7. kohonnut verenpaine, verenpainetauti<br>[_] 8. aivohalvaus (aivoverenvuoto, aivoveritulppa) |
| JOINT AND BACK DISEASES<br>[_] 9. rheumatoid arthritis<br>[_] 10. arthrosis (osteoarthritis)<br>10.a. If you answered to the question 10 yes, in which joints it has been identified?<br>You can select several answer alternatives.<br>[_] 10.1. knee<br>[_] 10.2. pelvis<br>[_] 10.3. hand<br>[_] 10.4. spine<br>[_] 10.5. other, what? [____________________]<br>[_] 11. back disease or other back deficit<br>[_] 12. neck disease or other neck deficit | NIVEL- JA SELKÄSAIRAUDET<br>[_] 9. nivelreuma<br>[_] 10. nivelkuluma (nivelrikko)<br>10a. Jos vastasit kysymykseen 10 kyllä, missä nivelissä se on todettu?<br>Voit valita useamman vastausvaihtoehdon.<br>[_] polvi<br>[_] lonkka<br>[_] käsi<br>[_] ranka<br>[_] muu, mikä? [____________________]<br>[_] 11. selkäsairaus tai muu selkävika<br>[_] 12. niskasairaus tai muu niskavika |
| INJURIES<br>[_] 13. permanent disability caused by an injury<br>13a. If you answered to the question 13 yes, a what kind of permanent disability is it? | TAPATURMAT<br>[_] 13. tapaturman aiheuttama pysyvä vamma<br>13a. Jos vastasit kysymykseen 13 kyllä, minkälainen pysyvä vamma on kyseessä? |
| You can select several answer alternatives.<br>[ ] 13.1. face or jaw injury<br>[ ] 13.2. some other head or brain injury<br>[ ] 13.3. visual impairment<br>[ ] 13.4. hearing impairment<br>[ ] 13.5. trauma in an upper limb or limbs<br>[ ] 13.6. pelvis fracture or its consequence<br>[ ] 13.7. some other trauma in a lower limb or limbs<br>[ ] 13.8. trauma in the body or back<br>[ ] 13.9. lung injury<br>[ ] 13.10. some other injury, what? [____________________] | Voit valita useamman vastausvaihtoehdon.<br>[ ] kasvo- tai leukavamma<br>[ ] jokin muu pää- tai aivovamma<br>[ ] näkövamma<br>[ ] kuulovamma<br>[ ] vamma yläraajassa/-raajoissa<br>[ ] lonkkamurtuma tai sen jälkitila<br>[ ] jokin muu vamma alaraajassa/-raajoissa<br>[ ] vamma vartalossa tai selässä<br>[ ] keuhkovamma<br>[ ] jokin muu vamma, mikä? [____________________] |
| MENTAL HEALTH PROBLEMS<br>[ ] 14. psychic or mental health-related disease<br>14a. If you answered to the question 14 yes, a what kind of disease is it?<br>You can select several answer alternatives.<br>[ ] 14.1. psychosis<br>[ ] 14.2. depression<br>[ ] 14.3. anxiety<br>[ ] 14.4. substance abuse problem<br>[ ] 14.5. other, what? [____________________] | MIELENTERVEYDEN ONGELMAT<br>[ ] 14. psyykkinen tai mielenterveyteen liittyvä sairaus<br>14a. Jos vastasit kysymykseen 14 kyllä, minkälainen sairaus on kyseessä?<br>Voit valita useamman vastausvaihtoehdon.<br>[ ] psykoosi<br>[ ] masennus<br>[ ] ahdistus<br>[ ] päihdeongelma<br>[ ] muu, mikä? [____________________] |
| VISION AND HEARING DEFICITS<br>[ ] 15. cataract<br>15a. If you answered to the question 15 yes, have you been in an eye surgery due to it?<br>[ ] Yes I have.<br>[ ] 16. glaucoma (ocular hypertension, glaucoma disease)<br>[ ] 17. macular degeneration<br>[ ] 18. hearing deficit, hearing disability or disease that weakens hearing | NÄKÖ- JA KUULOVIAT<br>[ ] 15. silmien harmaakaihi<br>15a. Jos vastasit kysymykseen 15 kyllä, oletko ollut sen takia silmäleikkauksessa?<br>[ ] Kyllä olen.<br>[ ] 16. silmien glaukooma (silmänpainetauti, viherkaihi)<br>[ ] 17. silmänpohjan rappeuma<br>[ ] 18. kuulovika, kuulovamma tai kuuloa heikentävä sairaus |
| OTHER DISEASES<br>[ ] 19. diabetes (diabetes mellitus)<br>[ ] 20. cancer disease (malignant tumor)<br>[ ] 21. Parkinson's disease<br>[ ] 22. involuntary urination, leakage of urine or urinary incontinence<br>23. Do you have still some other long-lasting disease, deficit, ailment or disability that a doctor has identified in you?<br>[ ] Yes I have.<br>[ ] 23a. If you answered to the question 23 yes, this other thing is what?<br>[____________________] | MUUT SAIRAUDET<br>[ ] 19. diabetes (sokeritauti)<br>[ ] 20. syöpätauti (pahanlaatuinen kasvain)<br>[ ] 21. Parkinsonintauti<br>[ ] 22. virtsan pidättämisen vaikeuksia, virtsan karkailua tai inkontinenssi<br>23. Onko sinulla vielä jokin muu lääkärin toteama pitkäaikainen sairaus, vika, vaiva tai vamma?<br>[ ] Kyllä on.<br>[ ] 23a. Jos vastasit kysymykseen 23 kyllä, mikä muu? [____________________] |
| Button "I save my response and continue forward" | Painike "Tallennan vastaukseni ja jatkan eteenpäin" |

**Table A11. Animation of the face figure that is shown together with the expression statements in the online questionnaire.**

The online questionnaire shows each expression statement as a speech bubble and below that an animation of a face figure is shown. The animation is based on 18 image frames that are illustrated in the following table. The animation is defined to present the frames 1-18 with the speed of 10 frames per second. In the end, the last frame 18 remains permanently displayed, thus there is no looping of the animation.

| | | | |
|---|---|---|---|
| Frames 1-5: 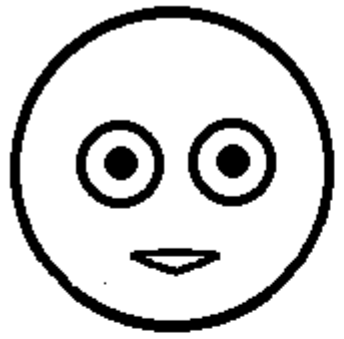 | Frame 6: 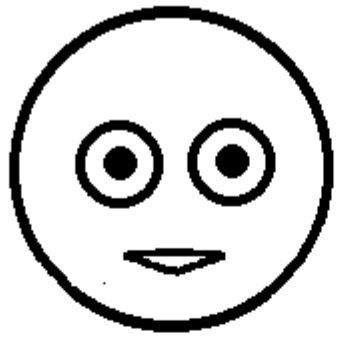 | Frame 7: 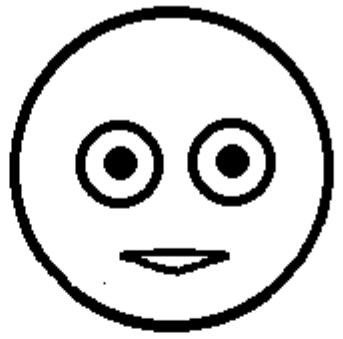 | Frames 8-10: 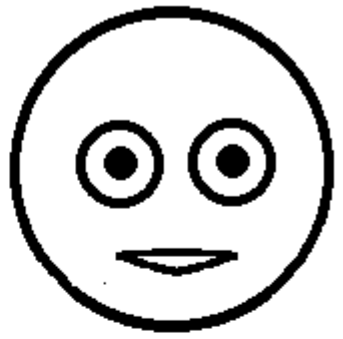 |
| Frame 11: 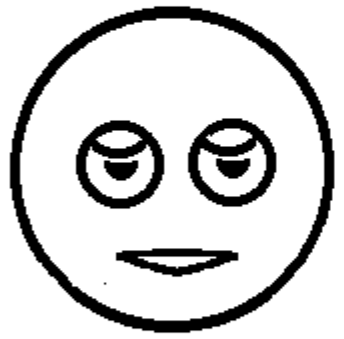 | Frames 12-13: 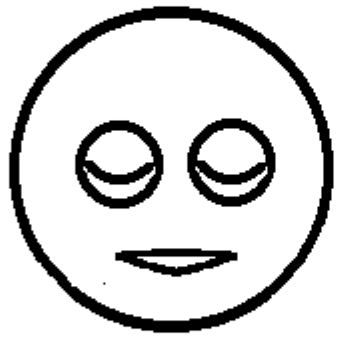 | Frame 14: 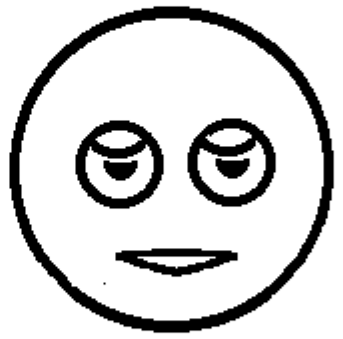 | Frame 15: 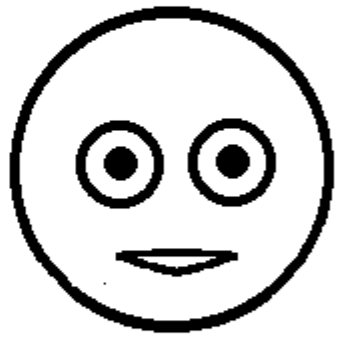 |
| Frame 16: 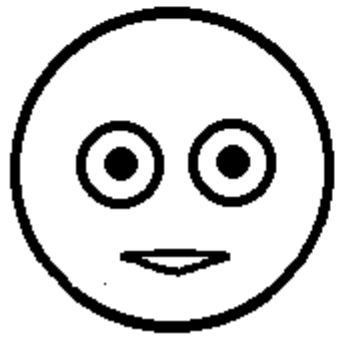 | Frame 17: 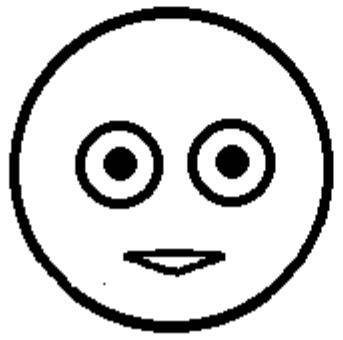 | Frame 18: 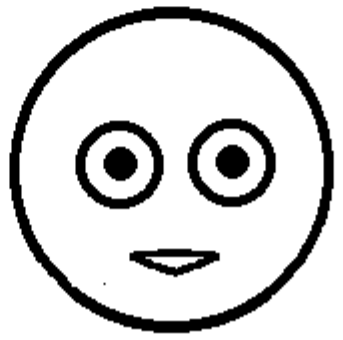 | (In the end, the last frame 18 remains permanently displayed.) |

**Table A12. Open data availability.**

While taking appropriate and sufficient anonymization actions in respect to addressing the General Data Protection Regulation of the European Union in handling the research data, DIHEML research project publishes an anonymized version of the current research data (an open access data set “Need for help related to the coronavirus COVID-19 epidemic”) in the supplementing Appendix A, to be used by anyone for non-commercial purposes (while citing this publication).

An open access data set “Need for help related to the coronavirus COVID-19 epidemic” to be used by anyone for non-commercial purposes (while citing this publication).

The data set will be added to this table at a later moment.